\documentclass[11pt,a4paper]{article}

\usepackage[utf8]{inputenc}
\usepackage[T1]{fontenc}
\usepackage{lmodern}
\usepackage[english]{babel}
\usepackage[a4paper,margin=2.4cm]{geometry}
\usepackage{microtype}
\usepackage{amsmath,amssymb,amsthm}
\usepackage{mathtools}
\usepackage{siunitx}
\usepackage{graphicx}
\usepackage{float}
\usepackage{booktabs}
\usepackage{longtable}
\usepackage{array}
\usepackage{multirow}
\usepackage{tabularx}
\usepackage{caption}
\usepackage{subcaption}
\usepackage{xcolor}
\usepackage{enumitem}
\usepackage{tikz}
\usetikzlibrary{positioning,calc,arrows.meta,decorations.pathreplacing,backgrounds,shapes.geometric,fit,patterns}
\usepackage[hidelinks,colorlinks=true,linkcolor=blue!60!black,citecolor=blue!60!black,urlcolor=blue!60!black]{hyperref}
\usepackage[nameinlink,capitalise]{cleveref}
\usepackage[acronym,nonumberlist,nopostdot,toc]{glossaries}

\graphicspath{{figures/}}

\DeclareSIUnit{\decare}{decare}
\DeclareSIUnit{\hectare}{ha}
\DeclareSIUnit{\litre}{L}
\DeclareSIUnit{\year}{yr}
\DeclareSIUnit{\Wh}{Wh}
\DeclareSIUnit{\kWh}{kWh}
\DeclareSIUnit{\auger}{augers}
\DeclareSIUnit{\EUR}{EUR}
\DeclareSIUnit{\USD}{USD}
\DeclareSIUnit{\hp}{hp}

\newcommand{\cabletractrepo}{\url{https://github.com/yilmazozgur/cabletract}}

\usepackage{titlesec}
\titleformat{\section}{\normalfont\Large\bfseries}{\thesection}{0.6em}{}
\titleformat{\subsection}{\normalfont\large\bfseries}{\thesubsection}{0.6em}{}
\titleformat{\subsubsection}{\normalfont\normalsize\bfseries}{\thesubsubsection}{0.6em}{}
\titlespacing*{\section}{0pt}{1.4ex}{0.7ex}
\titlespacing*{\subsection}{0pt}{1.1ex}{0.5ex}
\titlespacing*{\subsubsection}{0pt}{0.9ex}{0.4ex}

\makeglossaries

\newacronym{asabe}{ASABE}{American Society of Agricultural and Biological Engineers}
\newacronym{iwrc}{IWRC}{Independent Wire Rope Core}
\newacronym{ice}{ICE}{Internal Combustion Engine}
\newacronym{ice3}{ICE\,v3}{Inventory of Carbon and Energy, version 3 (Hammond \& Jones)}
\newacronym{ev}{EV}{Electric Vehicle}
\newacronym{wd}{4WD}{Four-Wheel Drive}
\newacronym{lca}{LCA}{Life-Cycle Assessment}
\newacronym{lcoe}{LCOE}{Levelised Cost of Energy / Operation}
\newacronym{npv}{NPV}{Net Present Value}
\newacronym{dcf}{DCF}{Discounted Cash Flow}
\newacronym{capex}{CAPEX}{Capital Expenditure}
\newacronym{opex}{OPEX}{Operating Expenditure}
\newacronym{bom}{BoM}{Bill of Materials}
\newacronym{bos}{BoS}{Balance of System (PV cabling, mounting, inverter)}
\newacronym{eu}{EU}{European Union}
\newacronym{latam}{LATAM}{Latin America}

\newacronym{pmsm}{PMSM}{Permanent-Magnet Synchronous Motor}
\newacronym{bldc}{BLDC}{Brushless DC motor}
\newacronym{pv}{PV}{Photovoltaic}
\newacronym{vawt}{VAWT}{Vertical-Axis Wind Turbine}
\newacronym{soc}{SoC}{State of Charge}
\newacronym{bms}{BMS}{Battery Management System}
\newacronym{gnss}{GNSS}{Global Navigation Satellite System}
\newacronym{gps}{GPS}{Global Positioning System}

\newacronym{tmy}{TMY}{Typical Meteorological Year}
\newacronym{ghi}{GHI}{Global Horizontal Irradiance}
\newacronym{doy}{DOY}{Day Of Year}
\newacronym{nrel}{NREL}{(US) National Renewable Energy Laboratory}
\newacronym{nsrdb}{NSRDB}{(NREL) National Solar Radiation Database}
\newacronym{pvgis}{PVGIS}{(EU JRC) Photovoltaic Geographical Information System}
\newacronym{sarah}{SARAH2}{Surface Solar Radiation Data Set --- Heliosat, version 2}
\newacronym{niwe}{NIWE}{(India) National Institute of Wind Energy}
\newacronym{inmet}{INMET}{(Brazil) National Institute of Meteorology}
\newacronym{swera}{SWERA}{Solar and Wind Energy Resource Assessment}

\newacronym{p10}{P10}{10th percentile of a sampled distribution}
\newacronym{p50}{P50}{50th (median) percentile of a sampled distribution}
\newacronym{p90}{P90}{90th percentile of a sampled distribution}
\newacronym{gbt}{GBT}{Gradient-Boosted Trees (regression)}
\newacronym{gbr}{GBR}{Gradient Boosting Regressor (scikit-learn)}
\newacronym{nsga}{NSGA-II}{Non-dominated Sorting Genetic Algorithm, version II}
\newacronym{salib}{SALib}{Sensitivity Analysis Library (Python)}
\newacronym{sobol1}{S1}{Sobol first-order sensitivity index}
\newacronym{sobolt}{ST}{Sobol total-order sensitivity index}
\newacronym{rmse}{RMSE}{Root Mean Squared Error}
\newacronym{r2}{R\textsuperscript{2}}{Coefficient of determination}

\newacronym{fem}{FEM}{Finite Element Method}
\newacronym{usda}{USDA NASS}{(US) Department of Agriculture, National Agricultural Statistics Service}
\newacronym{defra}{DEFRA}{(UK) Department for Environment, Food \& Rural Affairs}
\newacronym{eea}{EEA}{European Environment Agency}
\newacronym{ivl}{IVL}{IVL Swedish Environmental Research Institute}
\newacronym{inrae}{INRAE}{French National Research Institute for Agriculture, Food and Environment}
\newacronym{dlg}{DLG}{Deutsche Landwirtschafts-Gesellschaft}
\newacronym{bnef}{BloombergNEF}{Bloomberg New Energy Finance}
\newacronym{api}{API}{Application Programming Interface}

\newacronym{mu}{MU}{Main Unit (winch + motor + battery + harvester module)}
\newacronym{rq}{RQ}{Research Question}
\newacronym{c1}{C1}{compaction-reduction claim}
\newacronym{c2}{C2}{energy-reduction claim}
\newacronym{c3}{C3}{off-grid-feasibility claim}
\newacronym{c4}{C4}{economics and life-cycle CO\textsubscript{2} claim}

\title{\textbf{CableTract: A Co-Designed Cable-Driven Field Robot for\\
Low-Compaction, Off-Grid Capable Agriculture}\\[0.4em]
\large A Prototype-Free Feasibility Framework and Analytical Envelope%
\thanks{A preliminary version of this work is available as a preprint: \texttt{arXiv:2604.09938}.}}
\author{Özgür Yılmaz\\[0.3em]
\normalsize Adana Science and Technology University, Adana, Türkiye\\[0.15em]
\normalsize \texttt{ozguryilmaz@atu.edu.tr} \quad \texttt{yilmazozgur.kaan@gmail.com}}
\date{April 2026}

\begin{document}
\maketitle

\begin{abstract}
\noindent
Conventional field operations spend most of their energy moving the tractor, not the implement, and pay a soil-compaction penalty for it. We present \textbf{CableTract}, a prototype-free feasibility study of a two-module cable-driven field robot: a stationary Main Unit and a lighter helical-pile Anchor tension a cable across a strip while a lightweight ${\approx}\SI{250}{\kilogram}$ carriage does the work, keeping the heavy bodies on the headland. The central design lever is a 10-implement library \emph{co-designed} for the architecture --- narrower, slower, shallower and lighter than tractor implements --- which closes the energy, anchor and compaction budgets.

\medskip
An open, reproducible pipeline couples catenary and anchor mechanics, \gls{asabe} D497.7 draft estimates, an hourly solar/wind/battery simulator on six climate sites, a coverage planner, a contact-pressure compaction model, discounted-cash-flow economics with life-cycle CO\textsubscript{2}, and global sensitivity analysis. Five higher-fidelity studies --- a nonlinear $p$--$y$ anchor model, a closed-loop depth-control co-simulation, a cable-safety model, a soil--tool discrete-element model, and a co-design robustness sweep --- stress-test the load-bearing assumptions.

\medskip
Under the reference assumptions the architecture cuts compacted area by ${\sim}98\%$ and useful energy ${\sim}4\times$ versus an \SI{80}{hp} diesel tractor, and is favourable in a tractor-\emph{replacement} frame near capex parity; off-grid operation and additive-purchase economics are climate- and scale-dependent. The dominant unresolved risks are \emph{mechanical} --- anchor loading, depth control and cable safety --- which the studies bound but a prototype must ultimately test: a consistent analytical envelope for a future prototype, not a validated machine.

\end{abstract}

\tableofcontents

\glsresetall

\section{Introduction}\label{sec:intro}

A modern \SI{80}{hp} \gls{wd} utility tractor weighs ${\approx}\SI{4}{\tonne}$, of which the implement is typically \SIrange{200}{500}{\kilogram}. Each pass therefore drags ten to twenty times more body mass than tool mass across the field, paying a soil-compaction penalty (mean contact pressure \SIrange{100}{250}{\kilo\pascal}, well above the \SI{50}{\kilo\pascal} threshold for clay soils) and burning ${\approx}\SI{12}{\litre\per\hectare}$ of diesel even on light operations. Over a 15-year lifetime on a \SI{25}{\hectare} farm this is \SI{4500}{\litre} of diesel and ${\approx}\SI{12}{\tonne}$ CO\textsubscript{2}eq.

CableTract attacks the dead-weight problem at the \emph{architectural} level rather than the powertrain level. Two modules sit on opposite headlands of a strip: a \textbf{Main Unit} carrying the winch, motor, controller, battery, \gls{pv} array, and small wind turbine; and an \textbf{Anchor} resisting the cable reaction with helical screw piles. A lightweight implement carriage rolls along the cable between them. Only the carriage enters the field. The heavy bodies move only at the end of each strip, by one strip width, along the headland. Replacing the heavy traction body with a tensioned cable changes three things at once: (i) only the carriage's mass compacts the soil; (ii) the energy budget no longer carries a \SI{4}{\tonne} body; (iii) the Main Unit, being nearly stationary, can carry deployable \gls{pv} and a small wind turbine more easily than a fully mobile electric tractor.

The natural objection is that this only works if the implements are also redesigned. A conventional 8-row planter or a \SI{3}{\meter} disk harrow is built to be towed by a \SI{4}{\tonne} tractor; it expects high draft and high speed. Towing those \emph{unmodified} implements with a cable robot is the wrong problem. The right problem is to \textbf{co-design the carriage and a small library of implements together}: narrow strip widths matched to the cable swath, slower operating speeds that exploit the $v^2$ drop in soil draft, shallower depths that match what a carriage-mounted tool can resist, and lighter implement frames freed from the hitch loads of a real tractor. This co-design is the main analytical lever of the paper and we return to it explicitly in \cref{sec:codesign,sec:soil}.

The contribution is not an experimental validation. It is a \textbf{prototype-free, fully reproducible feasibility envelope} that asks one question:

\begin{quote}
\itshape Under what (climate $\times$ farm size $\times$ operation type) conditions does a co-designed cable-driven field robot beat a conventional diesel tractor on energy, compaction, \gls{npv}, and life-cycle CO\textsubscript{2} --- and where doesn't it?
\end{quote}

We answer it with an open analytical pipeline\footnote{\cabletractrepo} that runs end-to-end in under ten minutes from a clean checkout, ships with all input data bundled (no live \gls{api} calls), and regenerates every figure and table in this paper from the seven phase scripts (\texttt{run\_phase1}--\texttt{run\_phase7}). \Cref{sec:methods} walks through the model in the order in which the figures are generated; \cref{sec:results} reports the results against four explicit claims; and \cref{sec:discussion} closes the envelope with the binding constraints.

\paragraph{Key results (codesigned reference).} Each is conditional on the workload and purchase frame stated in the body:
\begin{itemize}[leftmargin=1.4em,itemsep=0.12em]
    \item \textbf{Energy:} \SI{889}{\Wh\per\decare}\footnote{One decare $=0.1$\,ha $=1000\,\si{\meter\squared}$; \SI{1}{\Wh\per\decare} $=\SI{10}{\Wh\per\hectare}$.} (\SI{8.89}{\kWh\per\hectare}) delivered electrical (flat-field bound \SI{921}{\Wh\per\decare} without regen) --- about $4\times$ less \emph{useful} (drawbar) work and ${\sim}13\times$ less \emph{primary fuel} energy than the ${\approx}\SI{120}{\kWh\per\hectare}$ ($12\,\si{\litre\per\hectare}$) an \SI{80}{hp} diesel tractor burns; most of the larger factor is generic ICE-to-electric conversion, not the architecture.
    \item \textbf{Compaction:} ${\sim}98\%$ less compacted \emph{area} and a ${\sim}73\times$ lower per-vehicle contact-pressure index, because only the \SI{250}{\kilogram} carriage enters the field.
    \item \textbf{Off-grid:} \SIrange{10}{14}{decares\per day} median on harvested energy across the six sites; \emph{climate-conditional} (achievable in solar-rich climates, grid-backed in northern temperate ones), and annual energy-positivity is a weaker claim than hourly autonomy.
    \item \textbf{Economics (replacement frame, near capex parity):} \gls{npv} vs diesel \SI{+3575}{\EUR} at \SI{25}{\hectare\per\year}/8\,\%, positive across \SIrange{1}{100}{\hectare\per\year}; discounted payback \SI{1.3}{\year} at \SI{25}{\hectare}, below one year by \SI{50}{\hectare}. In the \emph{additive} frame the \SI{25}{\hectare} \gls{npv} is negative (${\approx}\SI{-31}{\kilo\EUR}$), positive only above ${\sim}\SI{240}{\hectare\per\year}$.
    \item \textbf{Life-cycle CO\textsubscript{2}:} \SI{14.6}{\kilogram\per\hectare\per\year} versus 32.5 (diesel) and 22.9 (electric tractor) --- a $2.2\times$ improvement independent of grid decarbonisation.
\end{itemize}

\section{Related Work}\label{sec:related}

\subsection{Prior cable-driven agriculture}\label{sec:related-cable}

Cable-driven field machines are not new. The closest published prior art is \textbf{US 8,763,714 B2}~\cite{orlando2014patent} (Orlando \& Zoffoli), which describes a cable traction system using two technically \emph{identical} electric machines, each with a hoist and stabilisation, with the implement moving between them as cable is wound and unwound. The patent motivates the design through reduced soil degradation and electric propulsion, and includes a telescopic arm for cable placement during repositioning.

CableTract overlaps with that family but differs in three structural ways:

\begin{enumerate}[leftmargin=1.6em,itemsep=0.2em]
    \item \textbf{Asymmetric architecture.} CableTract uses a heavy \gls{mu} (\gls{pmsm}, controller, battery, harvester) and a much lighter Anchor (helical pile drive, sensors). The Anchor needs no winch and no high-power electronics, which roughly halves its \gls{bom}.
    \item \textbf{Helical-pile anchoring.} Cable reaction is resisted by screw augers driven into the soil rather than by the machine's mass alone. This is what makes the lighter Anchor module geotechnically defensible (\cref{sec:anchor}).
    \item \textbf{Co-designed implements.} CableTract is paired with an implement library sized for the cable architecture, not borrowed from tractor inventories (\cref{sec:codesign}).
\end{enumerate}

We do \emph{not} claim CableTract is the first cable-driven agricultural machine. The contribution is an analytical evaluation of the asymmetric \gls{mu}\,+\,Anchor architecture under co-designed implements, and the identification of the operating envelope in which it is genuinely competitive.

\subsection{Adjacent vehicle classes}\label{sec:related-adjacent}

CableTract sits in a design space already populated by five distinct vehicle classes, and a fair comparison must place the architecture against each of them rather than against diesel alone. We bundle a competitor table\footnote{See \texttt{tables/competitor\_comparison.csv} in the repository.} with public-domain reference numbers for each class (\cref{sec:competitor}). The five classes are:

\begin{itemize}[leftmargin=1.6em,itemsep=0.15em]
    \item \textbf{Conventional \SI{80}{hp} \gls{wd} diesel utility tractor} --- cost-floor reference; the thing CableTract has to beat on energy, compaction, and \gls{npv}.
    \item \textbf{Battery-electric tractors} (Monarch MK-V, Kubota LXe-261, Solectrac eUtility) --- same form factor as a tractor but with a Li-ion pack. Solves the fuel cost but not the compaction problem; usually grid-charged.
    \item \textbf{Autonomous PV-powered field robots} (Farmdroid FD-20) --- small, light, on-board solar; designed for sugar beet / row crops. Cannot do tillage. Off-grid by construction.
    \item \textbf{Autonomous wheeled spot sprayers} (EcoRobotix ARA) --- narrow, light, on-board PV; spraying only.
    \item \textbf{Autonomous weeders} (Naio Oz / Orio) --- vegetable / row-crop only; no heavy draft.
\end{itemize}

Across this landscape CableTract is the only architecture that simultaneously (a) is \emph{off-grid capable}, (b) supports \emph{primary tillage} (within its co-designed envelope), and (c) does so with a \SI{250}{\kilogram} in-field footprint rather than a \SIrange{1}{4}{\tonne} tractor body. The numerical comparison is in \cref{sec:competitor}.

\section{System Concept and Co-Design}\label{sec:concept}

\begin{figure}[H]
\centering
\includegraphics[width=0.92\linewidth]{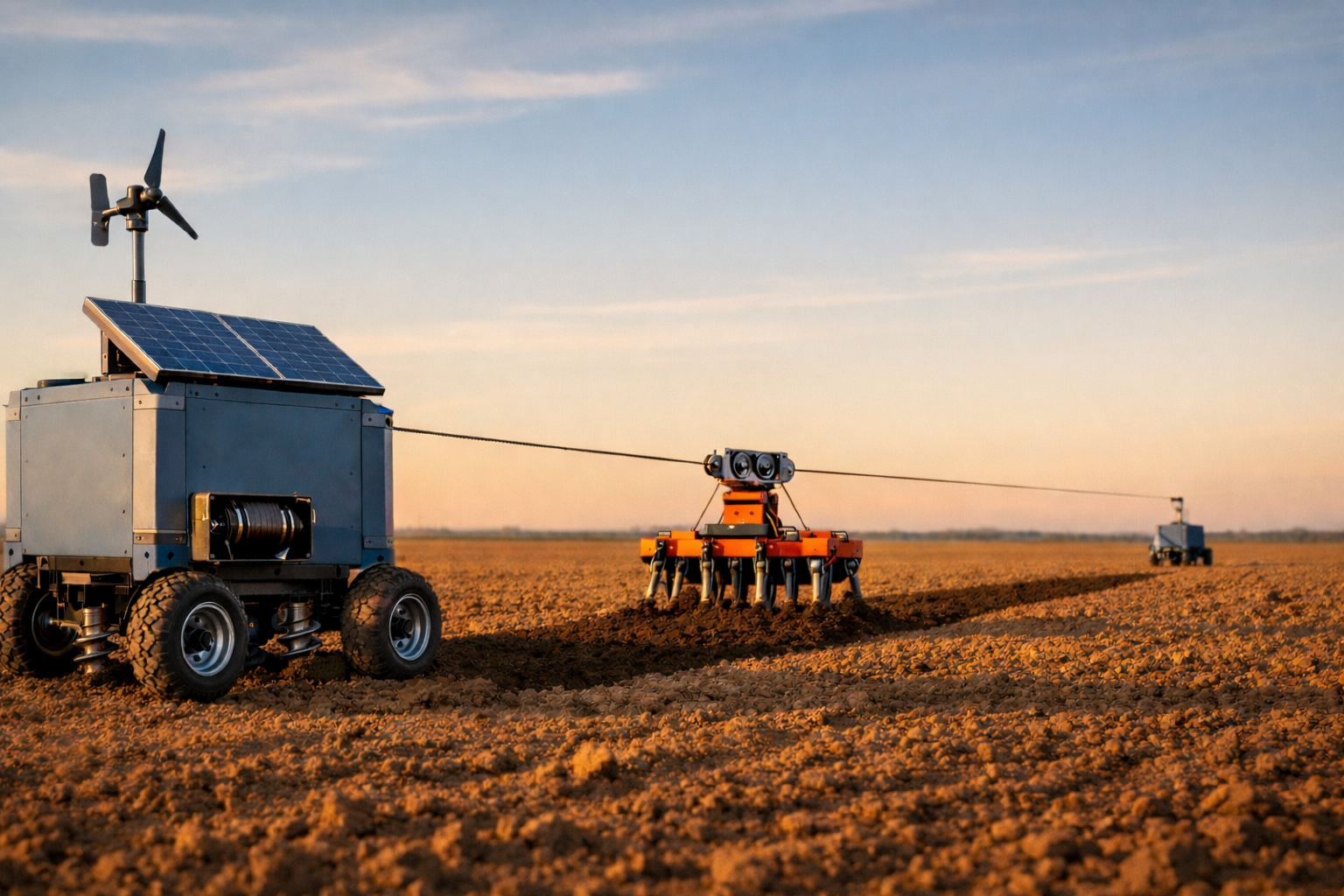}
\caption{CableTract operating in a field. The Main Unit (left) houses the winch, drivetrain, battery, PV panel and small wind turbine; the Anchor (right) is a passive helical-pile module that resists the cable reaction. A lightweight implement carriage runs along the tensioned cable across a typical \SI{50}{\meter} span. Only the carriage and the cable enter the field --- the two heavy modules stay on opposite headlands for the entire operation.}
\label{fig:f0b}
\end{figure}

\Cref{fig:f0b} shows the CableTract concept in operation: the two heavy modules rest on opposite headlands while only a tensioned cable and a lightweight implement carriage enter the field.

\subsection{Two-module architecture}\label{sec:two-module}

\begin{figure}[H]
\centering
\resizebox{\linewidth}{!}{%
\begin{tikzpicture}[
    every node/.style={font=\footnotesize},
    subsys/.style={draw=black, very thick, rounded corners=2pt, fill=blue!8,
                   minimum width=14mm, minimum height=6mm, align=center,
                   inner sep=2pt, font=\scriptsize},
    callout/.style={->, >=Stealth, thick, gray!70!black, shorten >=1pt},
    body/.style={draw=blue!50!black, very thick, fill=blue!18, rounded corners=4pt},
    ground/.style={pattern=north east lines, pattern color=brown!60!black, draw=brown!60!black},
    wheel/.style={draw=black, very thick, fill=gray!25, rounded corners=0.5pt},
]

\begin{scope}[xshift=0cm]

\draw[body] (0,0) rectangle (5.6,3.2);
\node[font=\bfseries\small] at (2.8,3.55) {Main Unit};

\draw[fill=blue!70!black, draw=black, thick]
    (0.4,3.2) -- ++(2.8,0) -- ++(-0.5,0.9) -- ++(-2.8,0) -- cycle;
\foreach \x in {0.0,0.4,0.8,1.2,1.6,2.0,2.4} {
    \draw[blue!90!white, very thin] (0.4+\x,3.25) -- (0.4+\x-0.5,4.05);
}
\node[font=\scriptsize, white] at (1.65,3.65) {PV};

\draw[thick] (4.6,3.2) -- (4.6,4.2);
\fill[gray!60] (4.6,4.2) circle (2pt);
\draw[thick] (4.6,4.2) -- ++(0.35,0.25);
\draw[thick] (4.6,4.2) -- ++(-0.35,0.25);
\draw[thick] (4.6,4.2) -- ++(0,0.4);

\node[subsys, fill=orange!18] (winch) at (4.0,1.9) {Winch\\+ drum};
\node[subsys, fill=yellow!25] (motor) at (4.0,0.7) {PMSM\\motor};
\node[subsys, fill=green!12] (batt)  at (1.6,1.9) {Battery\\\SI{9}{\kWh}};
\node[subsys, fill=red!12]   (ctrl)  at (1.6,0.7) {Controller\\GNSS / BMS};

\fill[gray!60] (5.6,1.9) circle (3pt);
\draw[thick] (5.6,1.9) circle (3pt);
\draw[very thick, black] (5.6,1.9) -- (7.0,2.0);
\node[anchor=west, font=\scriptsize] at (7.0,2.05) {cable};

\draw[wheel] (0.4,-0.05) rectangle (1.05,-0.7);
\fill[black] (0.725,-0.375) circle (0.1);
\draw[wheel] (4.55,-0.05) rectangle (5.2,-0.7);
\fill[black] (4.875,-0.375) circle (0.1);
\foreach \x in {1.6,2.5,3.4,4.3} {
    \draw[thick, gray!50!black] (\x,-0.05) -- (\x,-0.55);
    \draw[thick, gray!50!black] (\x-0.1,-0.55) -- (\x+0.1,-0.55)
                                                -- (\x,-0.75) -- cycle;
}

\draw[ground] (-0.3,-0.75) rectangle (5.9,-1.0);

\node[anchor=east, font=\scriptsize, align=right] (lblPV) at (-0.4,3.7) {PV panel\\\SI{15}{\meter\squared}};
\draw[callout] (lblPV.east) -- (1.6,3.7);

\node[anchor=east, font=\scriptsize, align=right] (lblWind) at (-0.4,4.6) {VAWT\\\SI{600}{\watt}};
\draw[callout] (lblWind.east) -- (4.55,4.65);

\node[anchor=west, font=\scriptsize, align=left] (lblWinch) at (6.2,1.4) {Winch +\\drum};
\draw[callout] (lblWinch.west) -- (winch.east);

\node[anchor=west, font=\scriptsize, align=left] (lblMot) at (6.2,0.4) {PMSM motor +\\inverter +\\gearbox};
\draw[callout] (lblMot.west) -- (motor.east);

\node[anchor=east, font=\scriptsize, align=right] (lblBatt) at (-0.4,2.3) {\SI{9}{\kWh}\\Li-ion};
\draw[callout] (lblBatt.east) -- (batt.west);

\node[anchor=east, font=\scriptsize, align=right] (lblCtrl) at (-0.4,0.7) {Controller,\\GNSS, BMS};
\draw[callout] (lblCtrl.east) -- (ctrl.west);

\node[anchor=north, font=\scriptsize, align=center] (lblAug) at (3.2,-1.2) {4 parallel auger drives\\(self-anchor)};
\draw[callout] (lblAug.north) -- (3.4,-0.75);

\node[anchor=east, font=\scriptsize, align=right] (lblWheel) at (-0.4,-0.4) {wheels\\(transport\\\& step)};
\draw[callout] (lblWheel.east) -- (0.4,-0.375);

\node[font=\scriptsize, gray!40!black] at (2.8,-2.0) {${\approx}\SI{1.4}{\tonne}$};

\end{scope}

\begin{scope}[xshift=11cm]

\draw[body] (0,0) rectangle (3.4,2.4);
\node[font=\bfseries\small] at (1.7,3.55) {Anchor};

\draw[thick] (0.4,2.4) rectangle (3.0,2.95);
\fill[gray!60] (1.7,2.95) circle (3.5pt);
\draw[thick] (1.7,2.95) circle (3.5pt);
\node[font=\scriptsize] at (1.7,2.65) {sheave};

\draw[very thick, black] (-1.4,1.85) -- (1.7,2.95);

\node[subsys, fill=red!12] (actrl) at (1.0,1.4) {Controller\\GNSS};
\node[subsys, fill=gray!12, minimum width=12mm] (aux) at (1.0,0.6) {Aux pack\\(9$\times$ aug drv)};

\draw[wheel] (0.25,-0.05) rectangle (0.75,-0.6);
\fill[black] (0.5,-0.325) circle (0.08);
\draw[wheel] (2.65,-0.05) rectangle (3.15,-0.6);
\fill[black] (2.9,-0.325) circle (0.08);

\foreach \x in {1.0,1.4,1.8,2.2,2.6} {
    \draw[thick, gray!50!black] (\x,-0.05) -- (\x,-0.55);
    \draw[thick, gray!50!black] (\x-0.08,-0.55) -- (\x+0.08,-0.55) -- (\x,-0.72) -- cycle;
}
\foreach \x in {0.85,1.2,1.6,2.0,2.4} {
    \draw[thick, gray!50!black] (\x,-0.05) -- (\x,-0.4);
    \draw[thick, gray!50!black] (\x-0.07,-0.4) -- (\x+0.07,-0.4) -- (\x,-0.55) -- cycle;
}

\draw[ground] (-0.3,-0.8) rectangle (3.7,-1.05);

\node[anchor=west, font=\scriptsize, align=left] (lblSheave) at (3.55,2.95) {Redirect\\sheave};
\draw[callout] (lblSheave.west) -- (2.5,2.95);

\node[anchor=west, font=\scriptsize, align=left] (lblCtrlA) at (3.55,1.4) {Controller,\\GNSS};
\draw[callout] (lblCtrlA.west) -- (actrl.east);

\node[anchor=west, font=\scriptsize, align=left] (lblAux) at (3.55,0.5) {Aux pack for\\9 parallel\\auger drives};
\draw[callout] (lblAux.west) -- (aux.east);

\node[anchor=north, font=\scriptsize, align=center] (lblAugA) at (1.8,-1.25) {9 parallel auger drives (3$\times$3 cluster, resist cable draft)};
\draw[callout] (lblAugA.north) -- (1.8,-0.8);

\node[anchor=east, font=\scriptsize, align=right] (lblWhA) at (-0.4,-0.35) {wheels};
\draw[callout] (lblWhA.east) -- (0.25,-0.325);

\node[font=\scriptsize, gray!40!black] at (1.7,-1.85) {${\approx}\SI{600}{\kilogram}$, no winch, no PV, no main battery};

\draw[->, very thick, red!70!black] (-0.6,1.85) -- ++(-0.7,0)
    node[anchor=east, font=\scriptsize, red!60!black] {$F_{\text{cable}}$};

\end{scope}

\end{tikzpicture}%
}
\caption{The two physically separable modules of CableTract. \textbf{Left:} the \textbf{Main Unit} carries the entire active drivetrain (PMSM + winch + drum), the energy stack (\SI{9}{\kWh} battery, \SI{15}{\meter\squared} PV, \SI{600}{\watt} VAWT) and the controller, and self-anchors with four parallel auger drives during operation (one BLDC drive motor per auger, so all four insert and retract concurrently). \textbf{Right:} the \textbf{Anchor} is a passive sheave block on top of nine parallel auger drives in a 3$\times$3 cluster that together resist the full horizontal cable reaction $F_{\text{cable}}$ at the opposite headland; it carries only an auxiliary pack powering the nine drives and a controller/GNSS for headland repositioning, no winch, no PV, no main battery. The parallel-drive choice on both modules is what keeps the per-round insert/retract cycle short enough to hit the \SI{60}{\second} setup target of \cref{tab:codesigned-params}; a single-drive-head design that visited each auger sequentially would not. The asymmetric split halves the BoM and mass of the second module relative to a symmetric two-winch design.}
\label{fig:f0a}
\end{figure}

CableTract has two physically separable modules (\cref{fig:f0a}; the deployed system is sketched in \cref{fig:f0b}):

\begin{itemize}[leftmargin=1.6em,itemsep=0.15em]
    \item \textbf{Main Unit} (${\approx}\SI{1.4}{\tonne}$ including PV deployable): \gls{pmsm} motor + 3-phase inverter + 2-stage gearbox + \SI{8}{\milli\meter} Dyneema-on-drum winch, \SI{9}{\kWh} Li-ion pack, \SI{15}{\meter\squared} flat-plate mono-Si \gls{pv}, \SI{600}{\watt} small \gls{vawt}, controller, \gls{gnss}, \gls{bms}, and 4 parallel auger drives for self-anchoring of the Main Unit itself (one small \gls{bldc} per auger, all four cycling concurrently).
    \item \textbf{Anchor} (${\approx}\SI{600}{\kilogram}$): 9 parallel auger drives in a 3$\times$3 cluster (one \gls{bldc} per auger, with a shared auxiliary battery pack and a single controller multiplexing across the nine drivers, so all nine insert and retract concurrently in a single per-round cycle), redirect sheave block, simple electric drive for headland repositioning on its own wheels, \gls{gnss}, no main battery beyond the auxiliary auger-drive pack.
\end{itemize}

In \emph{transport mode} the Anchor docks against the \gls{mu} and the assembly behaves as a compact \SI{1.5}{\meter} wide, \SI{1}{\meter} tall electric utility vehicle that can drive between fields on farm tracks. In \emph{work mode} the Anchor walks across the field on its own electric drive and plants itself at the far headland; the \gls{mu} runs the cable across the strip; the implement carriage rolls along the cable.

\subsection{Operating cycle}\label{sec:cycle}

\begin{figure}[H]
\centering
\begin{tikzpicture}[
    every node/.style={font=\scriptsize},
    mu/.style={draw=blue!50!black, very thick, fill=blue!22, rounded corners=2pt,
               minimum width=7mm, minimum height=6mm, align=center, font=\tiny\bfseries},
    anc/.style={draw=blue!50!black, very thick, fill=blue!22, rounded corners=2pt,
                minimum width=7mm, minimum height=6mm, align=center, font=\tiny\bfseries},
    impl/.style={draw=magenta!60!black, very thick, fill=magenta!60, rounded corners=1pt,
                 minimum width=2.6mm, minimum height=6mm, align=center,
                 font=\tiny\bfseries, text=white},
    field/.style={fill=brown!55!black},
    cable/.style={white, line width=1.2pt},
    dim/.style={<->, >=Stealth, gray!50!white, thick},
]

\begin{scope}[xshift=0cm]
\node[font=\scriptsize\bfseries, anchor=south] at (3.75,3.55) {(a) Forward leg --- loaded pull};

\node[font=\tiny\itshape, gray!40!black, anchor=south west] at (0,3.05) {Bird's-eye view};
\node[font=\tiny, gray!40!black, anchor=south east] at (7.5,3.05) {strip $w = \SI{1.5}{\meter}$};

\fill[field] (0,0) rectangle (7.5,3);

\node[mu] (mu1) at (0.7,1.5) {MU};
\node[anc] (an1) at (6.8,1.5) {An};

\draw[cable] ($(mu1.east) + (0,0.15)$) -- ($(an1.west) + (0,0.15)$);
\draw[cable] ($(mu1.east) + (0,-0.15)$) -- ($(an1.west) + (0,-0.15)$);

\node[impl] (im1) at (3.6,1.5) {};

\draw[->, very thick, white] (3.4,2.15) -- (1.6,2.15);
\node[white, font=\tiny, anchor=south] at (2.5,2.2) {loaded pull (\SIrange{1}{2.5}{\km\per\hour})};

\node[red!80!white, font=\tiny, anchor=north] at (3.6,1.1) {cable tension $T$};

\draw[dim] (mu1.south |- 0,0.45) -- (an1.south |- 0,0.45);
\node[white, font=\tiny, fill=brown!55!black, inner sep=1pt]
    at (3.75,0.45) {span $L = \SI{50}{\meter}$};
\end{scope}

\begin{scope}[xshift=8.6cm]
\node[font=\scriptsize\bfseries, anchor=south] at (3.75,3.55) {(b) Return leg + lateral step};

\node[font=\tiny\itshape, gray!40!black, anchor=south west] at (0,3.05) {Bird's-eye view};
\node[font=\tiny, gray!40!black, anchor=south east] at (7.5,3.05) {step laterally by $w$};

\fill[field] (0,0) rectangle (7.5,3);

\node[mu] (mu2) at (0.7,1.5) {MU};
\node[anc] (an2) at (6.8,1.5) {An};

\draw[cable] ($(mu2.east) + (0,0.15)$) -- ($(an2.west) + (0,0.15)$);
\draw[cable] ($(mu2.east) + (0,-0.15)$) -- ($(an2.west) + (0,-0.15)$);

\node[impl] (im2) at (4.5,1.5) {};

\draw[->, very thick, green!75!black] (4.7,2.15) -- (6.4,2.15);
\node[green!50!black, font=\tiny, anchor=south] at (5.55,2.2) {regen return};

\node[white, font=\tiny\itshape, anchor=north] at (4.5,1.1) {tool retracted};

\draw[->, very thick, yellow!85!black] (mu2.south) ++(0,-0.05) -- ++(0,-0.45);
\node[yellow!60!black, font=\tiny, anchor=north] at (0.7,1.0) {step};
\draw[->, very thick, yellow!85!black] (an2.south) ++(0,-0.05) -- ++(0,-0.45);
\node[yellow!60!black, font=\tiny, anchor=north] at (6.8,1.0) {step};

\draw[dim] (mu2.south |- 0,0.45) -- (an2.south |- 0,0.45);
\node[white, font=\tiny, fill=brown!55!black, inner sep=1pt]
    at (3.75,0.45) {span $L = \SI{50}{\meter}$};
\end{scope}

\end{tikzpicture}
\caption{One round of the CableTract operating cycle, bird's-eye view. \textbf{(a) Forward leg:} the Main Unit's winch reels in cable, the implement carriage moves loaded toward the MU at \SIrange{1}{2.5}{\kilo\meter\per\hour}, the working tool engages the soil, and the full draft $F_{\text{draft}}$ is reacted at the Anchor as $F_{\text{reaction}}$. \textbf{(b) Return leg:} the tool retracts, the cable goes slack, and the carriage rolls back to the Anchor end (downhill assist or active pull-back). The unloaded return leg recovers kinetic / potential energy via four-quadrant motor operation by default (\cref{sec:regen,sec:variant-norering}). After the return, the MU and the Anchor each step laterally by one strip width (\SI{1.5}{\meter}) along the headland and the cycle repeats on the next strip.}
\label{fig:f0c}
\end{figure}

One round of work proceeds as (\cref{fig:f0c}):

\begin{enumerate}[leftmargin=1.6em,itemsep=0.1em]
    \item The implement carriage starts at the Anchor end with the working tool engaged.
    \item The \gls{mu}'s winch reels in cable. The carriage moves loaded toward the \gls{mu} at the operating speed (typically \SIrange{1}{2.5}{\kilo\meter\per\hour}).
    \item At the \gls{mu} end, the carriage tool retracts, the cable goes slack, and the carriage either gravity-rolls back (downhill) or is towed back by a thin return line. The unloaded return leg can recover kinetic / potential energy via four-quadrant motor operation (regenerative braking, \cref{sec:regen}).
    \item The \gls{mu} and Anchor each step laterally by one strip width along the headland.
    \item Setup repeats on the next strip.
\end{enumerate}

Typical numbers for the codesigned reference: span \SI{50}{\meter}, strip width \SI{1.5}{\meter}, carriage mass ${\approx}\SI{250}{\kilogram}$, operating speed \SI{1.5}{\kilo\meter\per\hour}, setup time per round \SI{60}{\second}.

\subsection{Why an asymmetric architecture}\label{sec:asymmetry}

\begin{figure}[H]
\centering
\begin{tikzpicture}[
    every node/.style={font=\tiny},
    field/.style={fill=brown!55!black, draw=brown!30!black, very thick},
    tractor/.style={red!85!white, very thick, dashed, line cap=round},
    cable/.style={green!75!black, very thick},
    ctcarriage/.style={green!85!black, line width=2.2pt},
    legend/.style={font=\tiny, anchor=west},
]

\begin{scope}[xshift=0cm]
\node[font=\scriptsize\bfseries, anchor=south] at (3.5,4.35) {(a) Conventional tractor};
\fill[field] (0,0) rectangle (7,4.2);
\node[white, font=\tiny, anchor=north west] at (0.1,4.1) {Bird's-eye view};

\foreach \i [evaluate=\i as \y using {0.4+0.45*\i}] in {0,1,...,7} {
    \ifodd\i
        \draw[tractor] (6.5,\y) -- (0.5,\y);
    \else
        \draw[tractor] (0.5,\y) -- (6.5,\y);
    \fi
}
\foreach \i in {0,2,4,6} {
    \pgfmathsetmacro{\ya}{0.4+0.45*\i}
    \draw[tractor] (6.5,\ya) arc (-90:90:0.225);
}
\foreach \i in {1,3,5} {
    \pgfmathsetmacro{\ya}{0.4+0.45*\i}
    \draw[tractor] (0.5,\ya) arc (270:90:0.225);
}

\draw[fill=red!75!black, draw=black, thick]
    (6.15,3.75) rectangle (6.55,4.05);
\node[font=\tiny, white] at (6.35,3.9) {tr};

\node[white, font=\tiny\bfseries, align=center] at (3.5,0.15)
    {${\sim}\,100\,\%$ of field area is rolled over};
\end{scope}

\begin{scope}[xshift=8.4cm]
\node[font=\scriptsize\bfseries, anchor=south] at (3.5,4.35) {(b) CableTract};
\fill[field] (0,0) rectangle (7,4.2);
\node[white, font=\tiny, anchor=north west] at (0.1,4.1) {Bird's-eye view};

\draw[ctcarriage] (0.4,0.4) -- (0.4,3.8);
\draw[ctcarriage] (6.6,0.4) -- (6.6,3.8);

\foreach \y in {0.6,1.05,1.5,1.95,2.4,2.85,3.3,3.75} {
    \draw[cable, opacity=0.55] (0.4,\y) -- (6.6,\y);
}

\draw[fill=blue!40, draw=blue!50!black, very thick] (0.2,2.0) rectangle (0.6,2.35);
\node[white, font=\tiny] at (0.4,2.175) {MU};
\draw[fill=blue!40, draw=blue!50!black, very thick] (6.4,2.0) rectangle (6.8,2.35);
\node[white, font=\tiny] at (6.6,2.175) {An};

\draw[fill=magenta!60!white, draw=magenta!50!black, very thick]
    (3.3,2.07) rectangle (3.55,2.28);

\node[white, font=\tiny\bfseries, align=center] at (3.5,0.15)
    {${\sim}\,2\text{--}3\,\%$ of field area is rolled over};
\node[white, font=\tiny, align=center] at (3.5,3.9)
    {only the carriage enters the field};
\end{scope}

\node[legend, anchor=west] at (0,-0.45) {%
    \tikz{\draw[tractor] (0,0)--(0.5,0);} \, tractor body path
    \quad
    \tikz{\draw[ctcarriage] (0,0)--(0.5,0);} \, MU/Anchor walk on headland
    \quad
    \tikz{\draw[cable] (0,0)--(0.5,0);} \, cable / carriage strip};

\end{tikzpicture}
\caption{Why an asymmetric, headland-only architecture wins on compaction. \textbf{(a)} A conventional tractor must drive a boustrophedon (snake) path that covers essentially the entire field area at least once per pass --- every square metre of the soil is rolled over by a \SIrange{1}{4}{\tonne} body. \textbf{(b)} CableTract keeps the Main Unit and the Anchor on opposite headlands and runs only a \SI{250}{\kilogram} carriage along the cable across the field; over a full season the rolled-over fraction collapses to ${\sim}2\text{--}3\,\%$ of the field area (essentially the carriage strips). The contact-pressure $\times$ pass-count integrand drops by a much larger factor still --- see \cref{fig:f12} and \cref{sec:r-c1} for the quantitative compaction model.}
\label{fig:f0d}
\end{figure}

\Cref{fig:f0d} contrasts CableTract's headland-only coverage with a tractor's full-field boustrophedon path. Symmetric cable-traction systems (like the Orlando--Zoffoli patent~\cite{orlando2014patent}) need two winches, two batteries, two motor controllers, and two sets of high-power electronics. The asymmetric CableTract design moves all the power electronics into one module and lets the second module be a simple anchored drive. This roughly halves the \gls{bom} of the second module, halves its mass, and removes one of the two failure-prone winches.

The price of the asymmetry is that the cable reaction is one-sided. The Anchor must hold the entire horizontal draft against the soil. We address this with helical screw piles (\cref{sec:anchor}) and quantify the per-auger requirement against \emph{two} published lateral-capacity references --- a realistic medium-dense fixed-head nominal and a deliberately pessimistic loose-sand bound --- so that every auger count is reported with its soil assumption made explicit.

\subsection{Co-designed implement library}\label{sec:codesign}

The single most impactful design choice in this paper is \textbf{building implements for CableTract instead of borrowing them from a tractor}. A conventional 8-row planter is built to survive the hitch loads of a \SI{4}{\tonne} tractor at \SI{9}{\kilo\meter\per\hour} on a \SI{1.5}{\meter} depth-control wheel; it weighs \SI{800}{\kilogram} and pulls \SI{3.9}{\kilo\newton} of draft. CableTract neither needs nor can usefully exploit any of those numbers. The codesigned 4-row planter\footnote{Tabulated in \texttt{data/asabe\_d497\_cabletract.csv} of the repository.} is sized for a \SI{1.5}{\meter} strip, \SI{2}{\kilo\meter\per\hour} operating speed, $0.85\times$ the conventional working depth, and a carriage frame freed from tractor hitch loads. Its \gls{p50} draft drops from \SI{3.88}{\kilo\newton} to \SI{1.94}{\kilo\newton} --- a \textbf{factor of 2.0 reduction} (an 8-row tool becomes a 4-row tool on the narrow strip, run slower). We are careful below to separate which part of this is genuine implement re-engineering and which is simply a lighter operating point.

The 10-implement codesigned library and the achieved draft reduction (\cref{tab:codesigned-library}):

\begin{table}[H]
\centering\small
\caption{Codesigned implement library, \gls{p50} draft reduction relative to the conventional \gls{asabe} D497.7~\cite{asabe-d497} entry it derives from. Medium-texture soil, samples from \cref{sec:soil-monte-carlo}.}\label{tab:codesigned-library}
\resizebox{\textwidth}{!}{%
\begin{tabular}{lllrlrr}
\toprule
Operation & Conventional implement & & Conv. P50 (N) & Codesigned implement & Codes. P50 (N) & Reduction \\
\midrule
primary tillage   & subsoiler 1-shank      &  & 10\,981 & narrow ripper 1-shank (\SI{15}{\centi\meter}) & 3\,022 & $0.28\times$ \\
primary tillage   & chisel plow 7-tool     &  & 14\,174 & narrow chisel 4-tool (\SI{12}{\centi\meter}) & 4\,219 & $0.30\times$ \\
secondary tillage & disk harrow \SI{2.5}{\meter} tandem & & 8\,428  & narrow disk \SI{1.5}{\meter} & 3\,231 & $0.38\times$ \\
secondary tillage & field cultivator 11 S-tine & & 3\,175 & narrow cultivator 6-tool & 1\,347 & $0.42\times$ \\
primary tillage   & sweep plow \SI{3}{\meter} & & 15\,412 & narrow sweep \SI{1.5}{\meter} & 5\,174 & $0.34\times$ \\
seeding           & row planter 8-row      &  & 3\,876  & codesigned planter 4-row & 1\,935 & $0.50\times$ \\
seeding           & grain drill 24-row     &  & 9\,302  & codesigned drill 12-row & 4\,650 & $0.50\times$ \\
weeding           & rotary hoe \SI{4}{\meter} &  & 2\,251 & codesigned rotary hoe \SI{2}{\meter} & 788 & $0.35\times$ \\
spraying          & boom sprayer           &  & 204     & codesigned sprayer \SI{1}{\meter}   & 143 & $0.70\times$ \\
mowing            & rotary mower \SI{3}{\meter} & & 1\,452 & codesigned mower \SI{1.5}{\meter} & 523 & $0.36\times$ \\
\midrule
\multicolumn{6}{r}{\textbf{Median reduction}} & $\boldsymbol{0.37\times}$ \\
\bottomrule
\end{tabular}%
}
\end{table}

\textbf{Median reduction across the 10 operations: $0.37\times$ (${\approx}\,63\%$ lower draft).} The reduction comes from three independent levers, each defensible from the \gls{asabe} D497.7 equation
\begin{equation}\label{eq:d497}
D \;=\; F_{\!i} \,(A + B v + C v^{2})\, W \, T,
\end{equation}
where $D$ is draft (N), $v$ is field speed (\si{\kilo\meter\per\hour}), $W$ is the implement's natural width unit, $T$ is working depth, $F_{\!i}$ is a soil-texture multiplier, and $(A,B,C)$ are machine-specific coefficients. The three levers are:

\begin{enumerate}[leftmargin=1.6em,itemsep=0.15em]
    \item \textbf{Width.} Halving $W$ halves draft directly. The cable architecture wants narrow strips because the \SI{50}{\meter} cable span and the \SI{1.5}{\meter} strip width together produce a \SI{75}{\meter\squared} working envelope per round, which matches the carriage's mechanical envelope.
    \item \textbf{Speed.} The $B v + C v^{2}$ terms grow rapidly with speed. Operating at \SIrange{1.5}{2.5}{\kilo\meter\per\hour} instead of \SIrange{5}{9}{\kilo\meter\per\hour} drops the speed-dependent draft contribution by roughly 60\,\%. This is a \emph{physical} gain, not a marketing claim, and it is the same lever any slow-moving robot exploits. \Cref{fig:f5} quantifies the saving.
    \item \textbf{Depth.} Working at $0.85\times$ the conventional depth drops $T$ by 15\,\% directly. CableTract takes one extra pass per season for primary tillage to compensate, which is bookkept in the energy and economics models.
\end{enumerate}

\textbf{Accounting of the levers.} Two of these three levers are \emph{operating-point} choices rather than implement re-engineering, and we state this plainly. For the depth-dependent tillage tools (narrow ripper, chisel, disk, sweep, cultivator) the codesigned entries in \texttt{data/asabe\_d497\_cabletract.csv} carry the \emph{same} D497.7 $(A,B,C)$ coefficients as the conventional implement they derive from; the draft reduction is obtained by running the same tool geometry shallower, slower, and on a narrower strip. This is physically real --- \cref{eq:d497} is linear in $W$ and $T$ and rises with $v$ --- but it carries an agronomic obligation: a shallow narrow pass is not interchangeable with a deep wide one for every tillage objective (a \SI{15}{\centi\meter} ripper pass does not fracture a deep plough pan the way a \SI{40}{\centi\meter} subsoiler pass does), which is why we bookkeep an extra primary-tillage pass and restrict the heaviest operations to favourable soil in \cref{sec:anchor}. The row-count halving for the planter and drill (8$\to$4 rows, 24$\to$12 rows) halves per-pass draft but also halves the area covered per pass, so it is throughput- and energy-per-area-neutral; its benefit is the lighter frame and slower speed, not the row count itself. Only the lighter tools (rotary hoe, mower, sprayer) gain a genuine reduction in the constant term $A$ from a frame freed of tractor hitch loads. We therefore present the $0.37\times$ median as the achieved draft ratio at the codesigned \emph{operating point}, not as a claim that the tools are intrinsically half as draft-hungry. \Cref{tab:agronomic} states, per operation, where the reduction comes from and whether the codesigned pass is agronomically equivalent --- a design judgement to be confirmed by a prototype, not a measured result.

\begin{table}[H]
\centering\small
\caption{Agronomic-equivalence status of each co-designed operation. ``Source'' separates genuine implement re-engineering (a lighter frame lowers the constant term $A$) from operating-point choices (narrower $W$, slower $v$, shallower $T$, with D497 coefficients unchanged). Equivalence is a design judgement for a prototype to test.}\label{tab:agronomic}
\resizebox{\textwidth}{!}{%
\begin{tabular}{lll}
\toprule
Operation (conventional $\to$ codesigned) & Draft-reduction source & Agronomic-equivalence status \\
\midrule
subsoiler $\to$ narrow ripper (\SI{15}{\centi\meter})       & depth + speed (coeffs unchanged) & \emph{Not} equivalent for deep pan; needs 2 passes / favourable soil \\
chisel 7-tool $\to$ narrow chisel 4-tool                    & depth + speed + tool count       & Approx.\ equivalent with an extra pass \\
sweep \SI{3}{\meter} $\to$ narrow sweep \SI{1.5}{\meter}     & width + depth                    & Approx.\ equivalent (more strips) \\
disk \SI{2.5}{\meter} $\to$ narrow disk \SI{1.5}{\meter}     & width + lighter frame            & Likely equivalent (secondary tillage) \\
cultivator 11-tine $\to$ 6-tool                             & width + tool count               & Likely equivalent \\
planter 8-row $\to$ 4-row                                   & row count (area-neutral)         & Equivalent (twice the strips) \\
grain drill 24-row $\to$ 12-row                             & row count (area-neutral)         & Equivalent (twice the strips) \\
rotary hoe \SI{4}{\meter} $\to$ \SI{2}{\meter}              & width                            & Equivalent \\
boom sprayer $\to$ \SI{1}{\meter} sprayer                   & width + lighter frame ($A$)      & Equivalent \\
rotary mower \SI{3}{\meter} $\to$ \SI{1.5}{\meter}          & width + lighter frame ($A$)      & Equivalent \\
\bottomrule
\end{tabular}%
}
\end{table}

The price of the codesigned library is that the farmer cannot take a CableTract to a neighbour's field and hook up the neighbour's plough. CableTract is sold as a \emph{system} --- Main Unit + Anchor + carriage + implements --- not as a generic cable robot that accepts off-the-shelf tools. This is the same business model as Farmdroid, EcoRobotix, and Naio.

\section{Research Questions}\label{sec:rqs}

We organise the analysis around four claims and seven research questions. The four claims are the headline assertions the paper defends with code-traceable numbers; the seven research questions decompose each claim into the operational regime, the physical mechanism, and the binding constraint.

\paragraph{C1 --- Compaction.} CableTract reduces in-field compacted area by at least 90\,\% and the contact-energy index by at least $20\times$ compared with an \SI{80}{hp} \gls{wd} diesel tractor on the same field, because only the lightweight carriage rolls inside the field while the \gls{mu} and Anchor stay on the headland.

\paragraph{C2 --- Energy.} With a co-designed implement library and regenerative braking on the return leg (the default design), CableTract operates at ${\approx}\SI{0.89}{\kWh\per\decare}$ (${\approx}\SI{8.9}{\kWh\per\hectare}$) of delivered electrical energy --- about $4\times$ less \emph{useful} (drawbar) work, and ${\sim}13\times$ less \emph{primary fuel} energy, than a diesel tractor on the same operation, driven by (i) the absence of dead-weight transport, (ii) the $v^2$ speed reduction, (iii) the narrow-strip width, (iv) a small return-leg regenerative recovery, and (v) electric rather than thermal conversion.

\paragraph{C3 --- Off-grid.} Off-grid operation is achievable in solar-rich climates (Konya, Ludhiana, São Paulo) with \SI{15}{\meter\squared} of \gls{pv} and a \SI{9}{\kWh} battery, conditional in Mediterranean ones (Palencia, Des Moines), and not achievable at any practical hardware scale in northern temperate climates (Beauce) under a \SI{6}{\hour\per day} duty cycle. The off-grid claim is \emph{climate-conditional}, not categorical.

\paragraph{C4 --- Economics and life-cycle CO\textsubscript{2}.} The codesigned CableTract is \gls{npv}-positive (replacement frame) across the full \SIrange{1}{100}{\hectare} farm-size range at all three discount rates we test (5/8/12\,\%), with discounted payback of \SI{1.30}{\year} at \SI{25}{\hectare} annual operating area and below one year by \SI{50}{\hectare}, and lifecycle CO\textsubscript{2} of \SI{14.6}{\kilogram\per\hectare\per\year} versus 32.5 for diesel --- a $2.2\times$ improvement that does not depend on grid decarbonisation.

\medskip
The research questions:

\begin{description}[leftmargin=2.0em,style=nextline,itemsep=0.25em]
    \item[\gls{rq}1] What is the compacted-area reduction and contact-energy reduction across rectangle, L-shape, and irregular-concave field classes?
    \item[\gls{rq}2] What energy per decare does the codesigned reference achieve, and how does it decompose between mechanical work, drivetrain losses, and dead-weight motion?
    \item[\gls{rq}3] Under which combinations of annual \gls{ghi}, panel area, and battery capacity is the system off-grid capable?
    \item[\gls{rq}4] For which operation classes is the architecture most favourable, and where does the helical-pile anchor envelope bind?
    \item[\gls{rq}5] How do polygon shape and size erode the throughput on real fields?
    \item[\gls{rq}6] Is the codesigned reference Pareto-optimal on (CAPEX $\times$ throughput $\times$ payback $\times$ energy intensity), or does it leave headroom?
    \item[\gls{rq}7] Across the (annual \gls{ghi} $\times$ farm size) plane, where is CableTract \gls{npv}-positive against diesel, and where does the off-grid feasibility envelope bind first?
\end{description}

\section{Methods}\label{sec:methods}

\subsection{Modelling philosophy}\label{sec:philosophy}

The analysis is parametric and designed to expose its assumptions. Every parameter has units, every default value has a citation in a bundled data file, and every claim in \cref{sec:results} has a per-figure CSV companion with the underlying numbers.\footnote{Full source and data: \cabletractrepo} We do not attempt high-fidelity soil-tool dynamics, full closed-loop control simulation, or finite-element cable modelling. We use first-principles work-energy balances, the standard \gls{asabe} D497.7 implement-draft equation, an hourly \gls{tmy}-based PV+battery simulator, a static contact-pressure compaction model, a discounted-cash-flow economics engine, and global Sobol sensitivity over 20 inputs. This is the appropriate fidelity for an early-stage feasibility study.

Every higher-level analysis is composed on top of one atomic computation that takes a parameter set and returns a result record. The energy simulator (\cref{sec:energy}), the global sensitivity decomposition (\cref{sec:sobol}), the architectural variants (\cref{sec:ml}), and the operating envelope (\cref{sec:envelope}) are all callers of this same routine. Every figure is therefore re-derived from a single physics-and-economics chain rather than from parallel approximations that drift apart.

\subsection{Codesigned reference parameter set}\label{sec:codesigned-ref}

The codesigned reference is the single system the rest of the paper analyses. It is \emph{the} baseline: there is no other parameter set we report headline numbers on, so every result in \cref{sec:results} can be traced back to the values in \cref{tab:codesigned-params}, which lists each entry together with a one-line engineering justification.

\begin{table}[H]
\centering\small
\caption{Codesigned reference parameter set. Used unchanged in every figure, table, and headline number in the paper.}\label{tab:codesigned-params}
\begin{tabularx}{\textwidth}{lrX}
\toprule
Parameter & Value & Notes \\
\midrule
span (\si{\meter}) & 50 & Anchor-to-Main-Unit cable distance \\
strip width (\si{\meter}) & 1.5 & Matches the codesigned implement library \\
implement + cable transport load (N) & 600 & Carriage tool + cable drag carried along the span ($F_i$ in \cref{eq:pass}); the carriage \emph{structural} mass is ${\approx}\SI{250}{\kilogram}$ and enters the compaction model (\cref{sec:compaction}) \\
draft load (N) & 1\,800 & Representative \gls{p50} for the medium-load operational calendar; the full 10-implement library median is \SI{2479}{\newton} (\cref{sec:soil-codesigned}), with heavy primary tillage higher \\
system repositioning load (N) & 2\,200 & Rolling-resistance force as the ${\approx}\SI{2}{\tonne}$ MU\,+\,Anchor step one strip width along the headland ($F_s$ in \cref{eq:pass}); a force, not a mass \\
drivetrain efficiency (effective) & 0.518 & One-way chain 0.50 (6 components, \cref{sec:drivetrain}) $+\,{\approx}3.5\%$ four-quadrant recovery (slope PE + KE; ${\approx}0$ on flat) \\
\gls{pv} area (\si{\meter\squared}) & 15 & Mono-Si flat plate, deployable \\
wind generator (W) & 100 & Typical harvested power; a \SI{600}{\watt}-rated helix \gls{vawt} ($C_p = 0.30$) that reaches nameplate only near \SI{12}{\meter\per\second} (\cref{sec:pvwindbat}) \\
setup time per round (\si{\second}) & 60 & Anchor placement + cable spool + carriage hookup \\
battery (\si{\kWh}) & 9 & Li-ion, 96/96\,\% charge/discharge eff. \\
operating window (\si{\hour\per day}) & 10 & Daily total work window \\
operating days/yr & 170 & Standard EU agronomic calendar \\
diesel reference fuel (\si{\litre\per\decare}) & 1.2 & Mid-range field operation (\SI{12}{\litre\per\hectare})~\cite{siemens1999machinery,asabe-d497} \\
diesel price (\si{\EUR\per\litre}) & 1.40 & EU 2024 agricultural diesel \\
capex CableTract (\si{\EUR}) & 35\,870 & Itemised in \cref{sec:economics} \\
project horizon (\si{\year}) & 15 & Matches Kubota / Monarch warranty bands \\
discount rate (\%) & 8 & EU farm equipment loan rate \\
\bottomrule
\end{tabularx}
\end{table}

\subsection{Pass model}\label{sec:pass-model}

For one round of work, useful mechanical work is
\begin{equation}\label{eq:pass}
E_{\text{mech,round}} \;=\; F_d\,L \;+\; F_i\,L \;+\; F_s\,W,
\end{equation}
where $F_d$ is the implement draft (N), $F_i$ the carriage-plus-cable transport load (N), $F_s$ the system-repositioning force (N), $L$ the cable span (m), and $W$ the lateral repositioning distance per round (m). For the codesigned reference the breakdown is $F_d L = \SI{90}{\kilo\joule}$ (72\,\%), $F_i L = \SI{30}{\kilo\joule}$ (24\,\%), and $F_s W = \SI{4.4}{\kilo\joule}$ (4\,\%): the in-field work ($F_d L + F_i L$, 96\,\%) dwarfs the headland-repositioning work ($F_s W$, 4\,\%), which is exactly the dead-weight saving relative to a tractor (where the system-repositioning term, paid over the whole field rather than once per strip, dominates). The $F_i L$ term charges the full transport load over the entire span, which slightly \emph{over}-estimates CableTract's energy (a rolling carriage does no net lifting work over a level span), so the \SI{889}{\Wh\per\decare} headline is conservative in this respect. (These fractions are of the round-trip mechanical work; the headline energy then divides by the regen-default effective drivetrain efficiency of \cref{sec:drivetrain}.)

\subsection{Cable mechanics, drivetrain, anchor}\label{sec:physics}

The cable mechanics and drivetrain model are verified against 22 analytic invariants (sag monotonicity, Hooke's law in the elastic regime, energy conservation on the regen leg, and others).\footnote{Test suite: \texttt{tests/test\_physics.py} in the repository.}

\subsubsection{Decomposed drivetrain efficiency}\label{sec:drivetrain}

A lumped winch efficiency $\eta_w \approx 0.5$ is the most common single-parameter shortcut in agricultural-electrification work. It collapses six independent loss mechanisms --- motor copper losses, inverter switching losses, gearbox friction, drum tension losses, pulley redirect losses, cable elongation hysteresis --- into one number, which a global sensitivity analysis (\cref{sec:sobol}) would then assign almost all of the variance to. The chain we use in this paper is the explicit product of those six components, each independently sourced from a datasheet or a published engineering standard:

\begin{equation}\label{eq:eta-chain}
\eta_{\text{drivetrain}} \;=\; \eta_{\text{motor}} \,\eta_{\text{ctrl}}\, \eta_{\text{gearbox}} \,\eta_{\text{drum}}\, \eta_{\text{pulley}}\, \eta_{\text{cable}}.
\end{equation}

Two presets are bundled (\cref{tab:eta-decomposition}); the ``baseline'' \emph{one-way} product matches the lumped 0.5 by construction, and ``premium'' recovers ${\approx}74\%$. The global sensitivity analysis in \cref{sec:sobol} then samples each component independently, so the Sobol decomposition sees six separate variance contributors instead of one undefended fudge factor.

This one-way chain governs \emph{motor sizing} (the peak loaded pull, \cref{sec:motor}). The \emph{energy budget} uses a slightly higher effective efficiency, because the default four-quadrant drive recovers a \emph{modest} amount of energy on the unloaded return leg: gravitational potential energy on sloped fields (\cref{eq:regen}), the carriage's kinetic energy at each stroke end, and dissipation a one-quadrant drive would otherwise lose to braking. We are deliberately conservative about its size. The draft work itself --- the dominant ${\sim}90\,\si{\kilo\joule}$ per pass --- is dissipated irreversibly in cutting soil and \emph{cannot} be regenerated; the carriage kinetic energy at \SI{2}{\kilo\meter\per\hour} is only ${\sim}\SI{40}{\joule}$; and on a perfectly flat field the gravity term of \cref{eq:regen} is zero. We therefore credit only ${\approx}3.5\%$ (slope-averaged over the gently-undulating fields of the target regions), lifting the effective drivetrain efficiency from the one-way 0.50 to ${\approx}0.518$ --- \emph{not} the large ``half the round is unloaded'' recovery a naive argument would suggest. This gives a headline \SI{889}{\Wh\per\decare}, against a conservative flat-field bound of \SI{921}{\Wh\per\decare} (the unidirectional drivetrain, \cref{sec:variant-norering}); the slope-dependent upside is parameterized in \cref{sec:regen}. The four-quadrant drive is a modest hardware premium (${\approx}\SI{300}{\EUR}$, folded into the Main Unit) and is standard equipment primarily for smooth torque control and slope recovery, not as a large flat-field energy lever.

\begin{table}[H]
\centering\small
\caption{Decomposed drivetrain efficiency. Baseline product reproduces the lumped 0.5; premium recovers 0.74.}\label{tab:eta-decomposition}
\begin{tabularx}{\textwidth}{Xrrl}
\toprule
Component & Baseline (low-cost) & Premium (best-of-class) & Source \\
\midrule
\gls{pmsm} motor              & 0.85 & 0.93 & \cite{asabe-ep496} \\
Three-phase inverter           & 0.92 & 0.96 & \cite{asabe-ep496} \\
2-stage planetary gearbox      & 0.88 & 0.95 & drive-vendor data \\
Drum capstan + bending         & 0.88 & 0.95 & \cite{feyrer2015wireropes} \\
Redirect sheave block          & 0.90 & 0.95 & \cite{feyrer2015wireropes} \\
Cable cyclic loss              & 0.92 & 0.97 & Marlow Excel D12 datasheet \\
\midrule
\textbf{Product}               & \textbf{0.50} & \textbf{0.74} & --- \\
\bottomrule
\end{tabularx}
\end{table}

\subsubsection{Catenary sag}\label{sec:catenary}

\begin{figure}[H]
\centering
\includegraphics[width=0.85\linewidth]{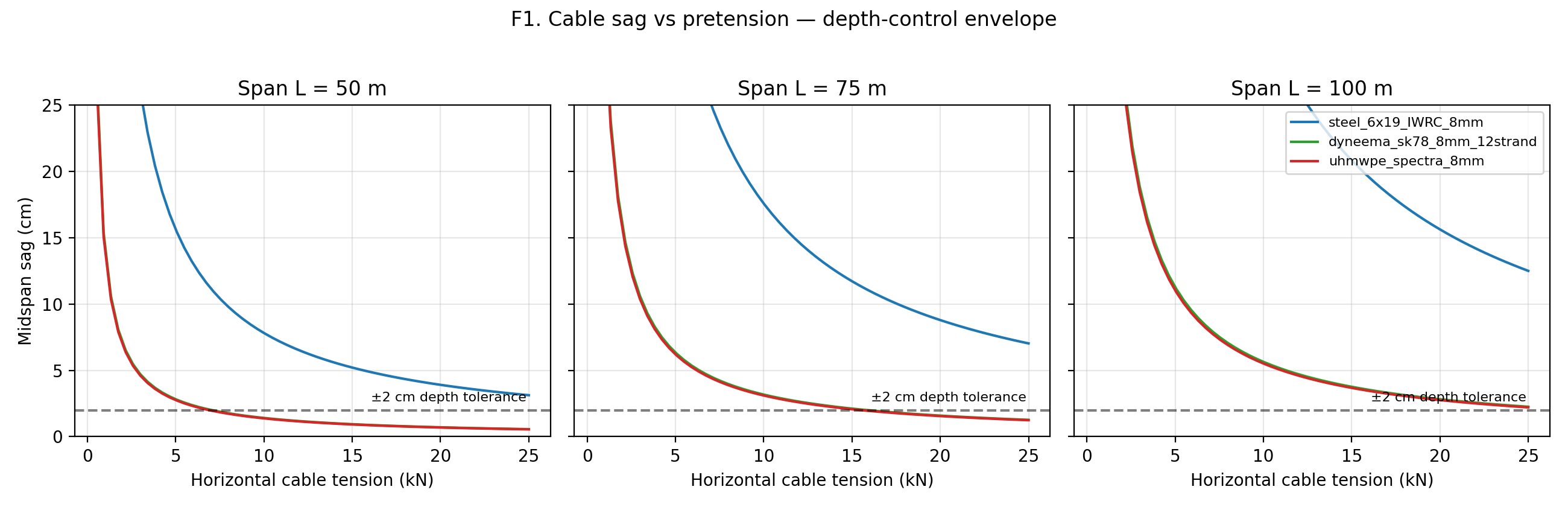}
\caption{Catenary sag versus horizontal cable tension for three rope materials at three spans. \emph{Tilling depth is set by the carriage's own depth wheel, not by cable sag} (\cref{sec:tension-balance}); the cable's only geometric job is to stay clear of the soil/crop. For a cable support height of ${\approx}\SI{0.8}{\meter}$ and a ground-clearance budget of ${\sim}\SI{0.3}{\meter}$ (a sag limit of ${\sim}\SI{0.5}{\meter}$), \SI{8}{\milli\meter} Dyneema (self-weight \SI{0.046}{\kilogram\per\meter}) needs only ${\approx}\SI{0.3}{\kilo\newton}$ pretension at the \SI{50}{\meter} reference span and ${\approx}\SI{1.1}{\kilo\newton}$ at \SI{100}{\meter} (parabolic limit $T\approx wL^2/8s$). This modest sag-floor pretension is below the implement draft for all but the lightest operations, so the Anchor reaction is governed by the \emph{draft} (\cref{sec:anchor}), not by sag. (A much tighter $\pm\SI{2}{\centi\meter}$ tolerance, were it required of the cable, would demand ${\approx}\SI{7}{\kilo\newton}$ at \SI{50}{\meter} --- but that tolerance belongs to the carriage's depth control, not to the cable.) Steel needs ${\sim}5\times$ more tension for the same clearance because of its ${\sim}5\times$ higher self-weight, which is why CableTract uses synthetic rope.}
\label{fig:f1}
\end{figure}

The first geometric requirement for any cable-based field robot is that the cable must not sag into the soil. If midspan sag exceeds the implement-attachment height, the geometry breaks before any draft, energy, or economics question can be asked. \Cref{fig:f1} resolves this with a closed-form bound: it computes the minimum horizontal tension required to keep midspan sag below the depth-control clearance budget, as a function of cable material, span, and self-weight.

A perfectly flexible cable of self-weight $w$ (\si{\newton\per\meter}) hung between two supports a horizontal distance $L$ apart at horizontal tension $T_h$ takes the catenary shape~\cite{irvine1981cables}
\begin{equation}\label{eq:catenary}
y(x) \;=\; a\bigl(\cosh(x/a)-1\bigr), \qquad a \;=\; T_h / w,
\end{equation}
with midspan sag
\begin{equation}\label{eq:sag-exact}
s \;=\; a\!\left(\cosh\!\left(\frac{L}{2a}\right) - 1\right).
\end{equation}
For $w L / (2 T_h) < 0.3$ the parabolic limit
\begin{equation}\label{eq:sag-parabolic}
s \;\approx\; \frac{w L^{2}}{8 T_h}
\end{equation}
is accurate to 1\,\%. Both forms are computed and cross-checked at every operating point.

\Cref{fig:f1} sweeps horizontal cable tension from \SI{0.5}{\kilo\newton} to \SI{25}{\kilo\newton} for steel 6$\times$19 \gls{iwrc}, Dyneema SK78 12-strand, and Spectra 1000 fibre rope (all \SI{8}{\milli\meter} nominal) at three spans $L \in \{50, 75, 100\}\,\si{\meter}$.\footnote{Mechanical properties from manufacturer datasheets, tabulated in \texttt{data/cable\_props.csv}.}

\textbf{What this gives the rest of the paper.} \Cref{fig:f1} narrows the design choice to synthetic rope and quantifies the minimum working tension as a function of span. CableTract uses \SI{8}{\milli\meter} Dyneema throughout. The minimum-tension number flows into \cref{sec:tension-balance} (which side of the draft-bound vs sag-bound regime are we in?) and into \cref{sec:anchor} (the helical-pile capacity must absorb the horizontal tension we just selected).

\subsubsection{Tension balance and operating regime}\label{sec:tension-balance}

The implement carriage's own depth-control mechanism --- not the cable --- sets tilling depth. The cable transmits the horizontal draft $F_d$ and stays clear of the soil. Two regimes arise:

\begin{itemize}[leftmargin=1.6em,itemsep=0.15em]
    \item \textbf{Draft-bound:} $F_d$ is large enough that the resulting horizontal tension keeps midspan sag below the clearance budget $h_p - c_{\min}$. Horizontal tension equals $F_d$, anchor sees ${\approx}F_d$.
    \item \textbf{Sag-bound:} $F_d$ is too small to keep the cable taut against its own weight. The Main Unit over-tensions to $T_{\text{sag,min}}$. The anchor sees this larger tension.
\end{itemize}

The tension-balance solver detects which regime applies at each operating point and reports the tensions seen by both ends of the cable. \Cref{fig:f2} sweeps draft from \SI{200}{\newton} to \SI{4000}{\newton} at \SI{8}{\milli\meter} Dyneema, span \SI{50}{\meter}, $h_p = \SI{1.5}{\meter}$, with the per-auger Anchor capacity bands overlaid.

\subsubsection{Anchor reaction envelope}\label{sec:anchor}

The asymmetric architecture of \cref{sec:asymmetry} only works if the lighter Anchor module can geotechnically resist the cable reaction. A mass-anchored solution would force the Anchor to weigh as much as a small tractor, which would defeat the architecture's selling point. We therefore replace mass anchoring with helical screw piles and quantify how many augers are required at each draft level. This is the load-bearing question behind the ``CableTract is mechanically possible'' claim.

\begin{figure}[H]
\centering
\includegraphics[width=0.85\linewidth]{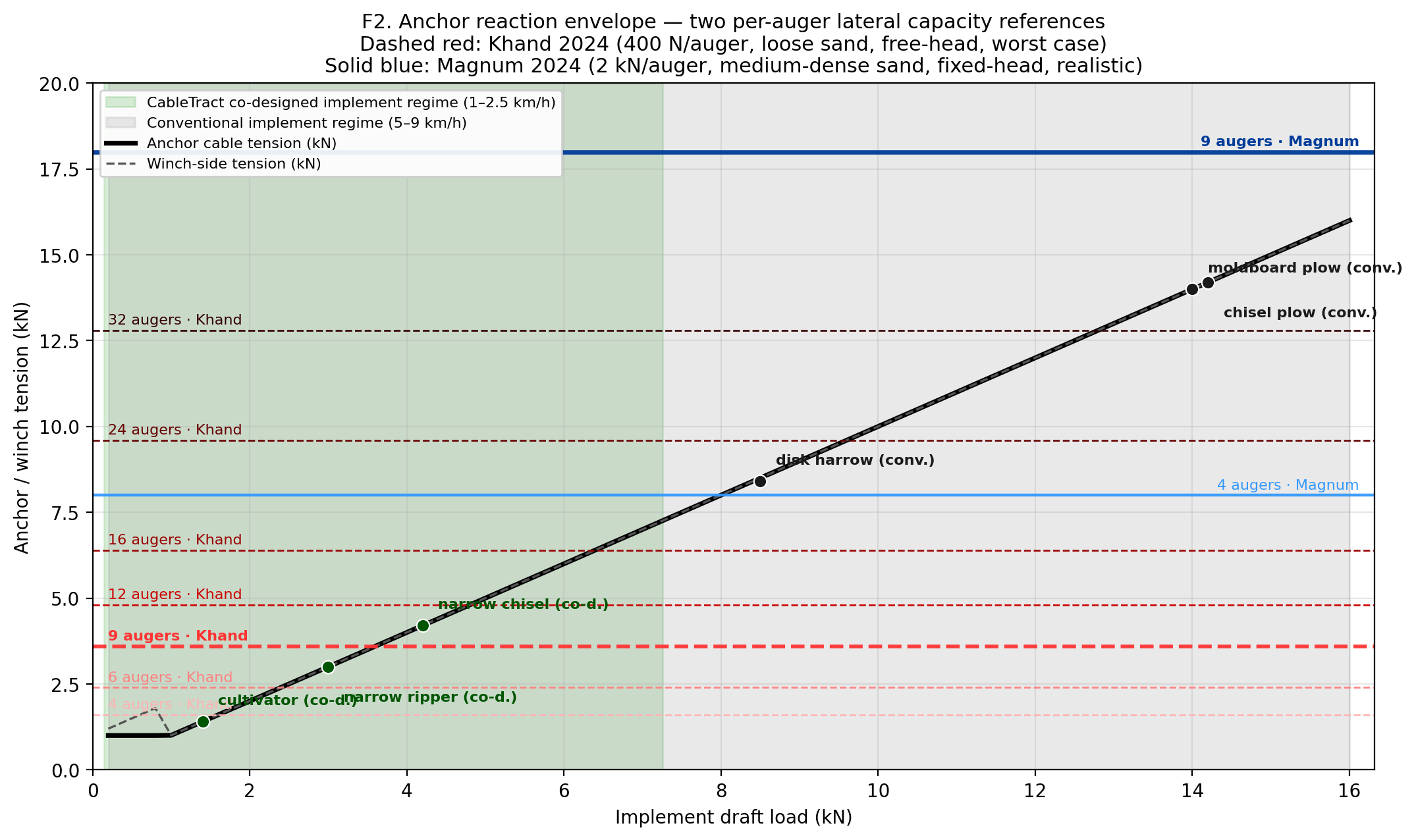}
\caption{Anchor reaction envelope plotted against \emph{two} per-auger lateral capacity references that span the published range. \textbf{Dashed red bands (Khand et al.\ 2024~\cite{khand2024helical})}: \SI{400}{\newton} per 30~cm helical pile in \emph{loose} sand, free-head, strict deflection limit (the per-pile interpretation of the 4-pile raft tests in that paper) --- the worst-case bound. \textbf{Solid blue bands (Magnum Piering 2024~\cite{magnum2024fixed}, with~\cite{chance2024manual} as a corroborating manual reference)}: \SI{2}{\kilo\newton} per pile, taken as a conservative downscale of the \SIrange{14}{20}{\kilo\newton} fixed-head capacity reported for medium-dense sand at the IBC2021 \SI{1}{inch} deflection limit --- the realistic engineering reference. The working safety factor is 1.15 in both envelope families. \SI{2}{\kilo\newton} per pile, a conservative downscale of the \SIrange{14}{20}{\kilo\newton} fixed-head capacity reported for medium-dense sand at the IBC2021 \SI{1}{inch} deflection limit --- the realistic engineering nominal. The working safety factor is 1.15 in both envelope families. On the \SI{2}{\kilo\newton} nominal the codesigned reference (P50 \SI{1.8}{\kilo\newton}, P90 \SI{3.0}{\kilo\newton}) needs only 2 augers, so the installed 9-auger Anchor over-provisions by ${\sim}4.5\times$; on the \SI{400}{\newton} worst-case bound the same reference needs 6 (P50) and 9 (P90) augers, which is what sets the 9-auger count. A conventional implement library at \SIrange{8}{15}{\kilo\newton} (disk harrow to chisel/sweep plow) would require 23--45 augers on the loose-sand bound but only 5--9 on the medium-dense nominal. The codesigned library is what keeps the Anchor feasible \emph{even in worst-case soil}; in typical soil the same Anchor carries it with comfortable margin.}
\label{fig:f2}
\end{figure}

The literature on lateral capacity of helical piles spans nearly two orders of magnitude depending on soil density, shaft fixity, embedment, and the chosen serviceability limit, so we anchor the analysis between a realistic nominal and a deliberately pessimistic bound rather than picking one number. The \textbf{Magnum Piering 2024 design tables}~\cite{magnum2024fixed} report allowable lateral capacities of \SIrange{14}{20}{\kilo\newton} per small fixed-head helical pile in medium-dense sand (SPT $N \approx 10$--$30$) at the IBC2021 \SI{1}{inch} deflection limit, and the industry-standard CHANCE Technical Design Manual~\cite{chance2024manual} brackets the same range. As our \emph{nominal} design basis we adopt a heavily conservative downscale of this --- \SI{2}{\kilo\newton} per auger (a $7$--$10\times$ reduction on the reported fixed-head values), at a working safety factor of 1.15 --- which still corresponds to the medium-dense, fixed-head conditions a real installer would target. As a \emph{worst-case stress bound} we additionally carry \SI{400}{\newton} per auger, our per-pile reading of the 4-pile group raft tests of \textbf{Khand et al.\ (2024)}~\cite{khand2024helical} for short piles in \emph{loose} sand at strict deflection thresholds with free-head boundary conditions; we flag explicitly that this number is extracted from a model-scale experiment, so it is used only as a pessimistic ordering bound, not as a field-calibrated capacity. Reporting both makes the soil assumption behind every auger count explicit. The codesigned 9-auger Anchor is sized so that even at the \SI{400}{\newton} worst-case bound it anchors the codesigned reference draft (P90 \SI{3.0}{\kilo\newton}, 9 augers); at the \SI{2}{\kilo\newton} nominal those same 9 augers carry the entire codesigned library --- and most conventional implements --- with a $4$--$5\times$ margin on auger count. We do not lean on the worst case to make the design work; we use it to show how much pessimism the design tolerates. The 9-auger capacity lines plotted in \cref{fig:f2,fig:f4} are \emph{nominal} ($9\times$ per-auger, i.e.\ \SI{3.6}{\kilo\newton} on the Khand bound and \SI{18}{\kilo\newton} on the Magnum nominal); the auger \emph{counts} we quote already apply the 1.15 working safety factor, $n = \lceil 1.15\,T / c \rceil$, so they are allowable-capacity counts, not nominal.

\textbf{Findings.} Working from the per-implement P50/P90 numbers in \cref{fig:f4}:
\begin{itemize}[leftmargin=1.6em,itemsep=0.15em]
    \item \textbf{Light codesigned operations are loose-sand feasible.} On the conservative Khand envelope, the 9-auger Anchor handles the codesigned cultivator (\SI{1.4}{\kilo\newton} P50, 5 augers required), planter (\SI{1.9}{\kilo\newton}, 6), rotary hoe (\SI{0.8}{\kilo\newton}, 3), sprayer (\SI{0.14}{\kilo\newton}, 1), and mower (\SI{0.5}{\kilo\newton}, 2) with margin. The narrow ripper at its P50 of \SI{3.0}{\kilo\newton} (8.6 augers) is the marginal case --- it sits at exactly the design value, and its P90 of \SI{3.8}{\kilo\newton} pushes 11 augers, just outside the installed envelope.
    \item \textbf{Heavy codesigned primary tillage is loose-sand restricted.} Four codesigned operations --- narrow sweep (\SI{5.2}{\kilo\newton} P50), narrow chisel (\SI{4.2}{\kilo\newton}), narrow disk (\SI{3.2}{\kilo\newton}), and codesign drill (\SI{4.6}{\kilo\newton}) --- exceed the 9-auger count at the \SI{400}{\newton} worst-case bound and would require \textbf{10--15 augers} at their P50 in loose sand. These four implements are restricted to the medium-dense nominal in the present design.
    \item \textbf{All codesigned operations are medium-dense feasible.} On the \SI{2}{\kilo\newton} nominal, the worst codesigned case (narrow sweep P90 \SI{7.25}{\kilo\newton}) requires \textbf{5 augers} --- the entire codesigned library fits inside the 9-auger Anchor with margin. The same 9-auger Anchor would also hold the heaviest \emph{conventional} implements (chisel plow \SI{14.2}{\kilo\newton} P50, sweep plow \SI{15.4}{\kilo\newton} P50 --- both 9 augers) at the design ceiling.
\end{itemize}

The reading of the physics chapter is therefore that the 9-auger Anchor is sized for the medium-dense soil regime, in which it carries the entire codesigned library with comfortable margin and is competitive even with conventional implements. In the worst-case loose-sand regime, the same Anchor is restricted to the lighter half of the codesigned library (cultivator, planter, rotary hoe, sprayer, mower, and the narrow ripper at its P50). The codesigned library is therefore not a worst-case geotechnical fix --- it is the design choice that simultaneously makes the energy budget work, the motor sizing work, and the medium-dense-soil anchor envelope work for the entire operation calendar. In loose-sand sites, the operating envelope shrinks to the lighter half of the library, and heavy primary tillage either waits for soil moisture conditions that improve fixity or is performed by a conventional implement on the same site.

The two-reference bracket above is deliberately agnostic about which per-auger capacity is ``correct''. \Cref{sec:v-s7} closes that gap with a nonlinear $p$--$y$ Winkler model that derives a single internally-consistent per-auger working nominal of \SIrange{1.6}{2.3}{\kilo\newton} --- which brackets the Magnum nominal and sits above the conservative Khand floor --- implying only ${\approx}3$ augers at the codesigned P90. We nevertheless retain the installed \(9\)-auger count: the $p$--$y$ result simply reframes it as a deliberate ${\approx}3\times$ margin and as redundancy, at no throughput cost (the augers are set once, in parallel, at strip setup), and the worst-case-bound counts quoted above stand as the conservative envelope.

\subsubsection{Peak vs continuous motor power}\label{sec:motor}

Day-averaged power numbers are convenient for energy budgets but misleading for motor sizing. A real motor must absorb the peak load --- start-up, loaded acceleration, transient draft spikes from the implement entering harder soil --- and the BOM cost of the motor, inverter, and cooling sits on that peak rather than on the average. We therefore separate the two.

Motor sizing is governed by the peak load, not the day-averaged power. We separate the two as
\begin{equation}\label{eq:power}
P_{\text{cont}} \;=\; \frac{F_{\text{cont}}\, v_{\text{cont}}}{\eta_{\text{drivetrain}}}, \qquad
P_{\text{peak}} \;=\; \frac{F_{\text{peak}}\, v_{\text{peak}}}{\eta_{\text{drivetrain}}}.
\end{equation}

For the codesigned reference at \SI{1.8}{\kilo\newton} P50 / \SI{3.0}{\kilo\newton} P90 (\cref{tab:motor}), the codesigned reference uses a \textbf{\SI{2.0}{\kilo\watt} continuous / \SI{3.0}{\kilo\watt} peak \gls{pmsm}}. This is roughly half the motor that an off-the-shelf-implement CableTract would need, and it shrinks the inverter, the gearbox, and the heat-sink mass in proportion. The cost saving propagates into \cref{sec:economics}.

\begin{table}[H]
\centering\small
\caption{Peak vs continuous motor power for the codesigned reference at the two drivetrain efficiency presets.}\label{tab:motor}
\begin{tabular}{lrr}
\toprule
Drivetrain & Continuous (\si{\kilo\watt}) & Peak (\si{\kilo\watt}) \\
\midrule
Baseline ($\eta = 0.50$) & 1.50 & 2.50 \\
Premium  ($\eta = 0.74$) & 1.01 & 1.69 \\
\bottomrule
\end{tabular}
\end{table}

\begin{figure}[H]
\centering
\includegraphics[width=0.85\linewidth]{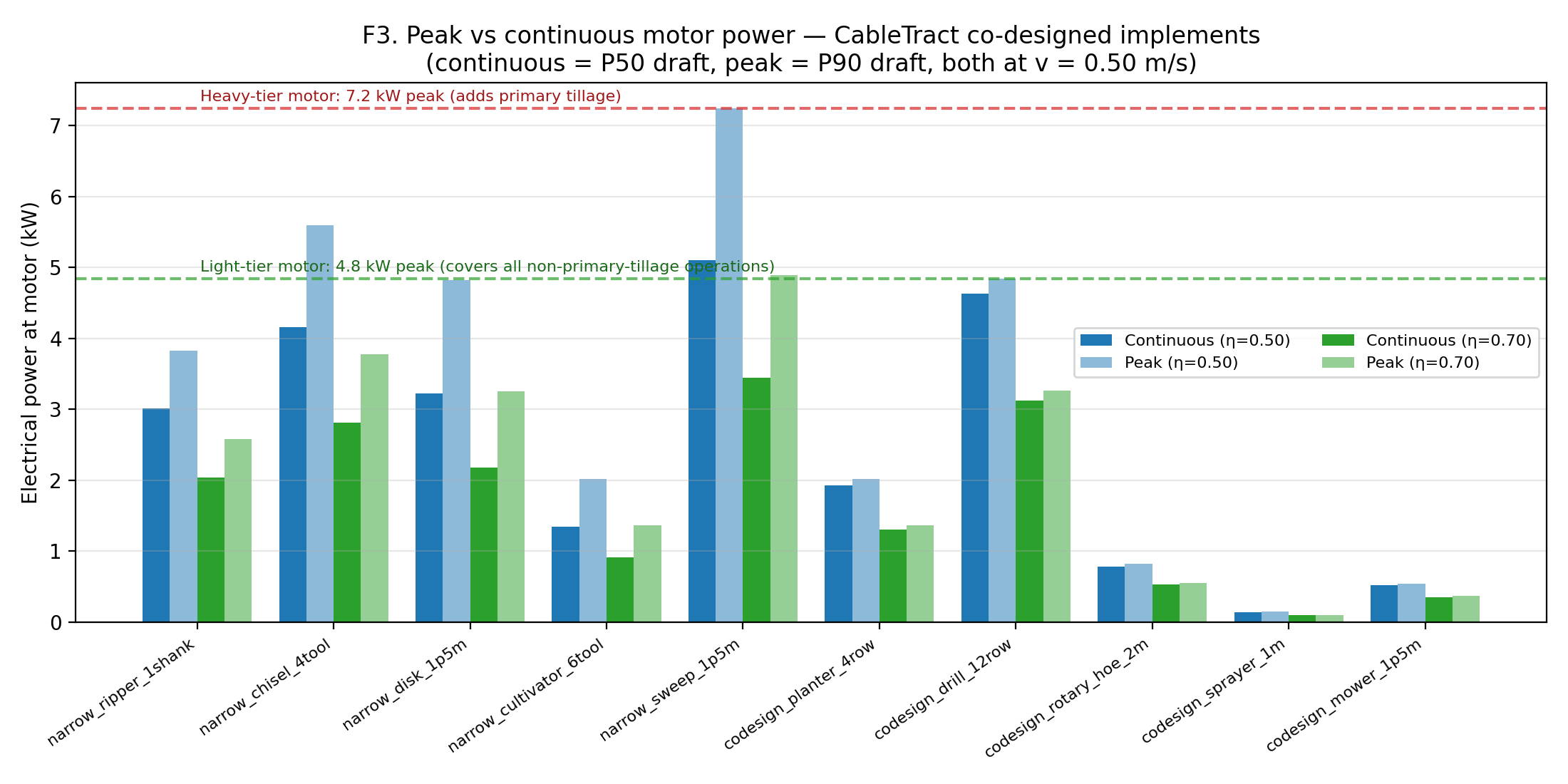}
\caption{Peak versus continuous motor power across the codesigned implement library at the two drivetrain efficiency presets.}
\label{fig:f3}
\end{figure}

\subsubsection{Regenerative return leg}\label{sec:regen}

A genuine energy advantage of the cable architecture is that the unloaded return leg lifts no implement, so any potential-energy budget (downhill slopes, carriage descent at end of row, elastic recoil of the cable as load is removed) can be recovered with four-quadrant motor operation. The recoverable energy on a return leg of length $d$ on a slope $\theta$ is
\begin{equation}\label{eq:regen}
E_{\text{regen}} \;=\; \eta_{\text{regen}}\!\left(m g \sin\theta - m g \cos\theta\,\mu_r\right) d,
\end{equation}
with $\eta_{\text{regen}} = 0.55$ and rolling resistance $\mu_r = 0.06$. On flat fields the gravity term vanishes and the formula returns zero, so the dominant regen contribution comes not from slope but from the four-quadrant motor recovering kinetic energy from the cable retraction itself; this is the pathway the default drivetrain exploits (\cref{sec:drivetrain}), isolated against a no-regen drivetrain in \cref{sec:variant-norering} and \cref{fig:f20}. Slope-driven regen only adds value on slopes steep enough to overcome rolling resistance ($\sin\theta > \mu_r$), i.e.\ above ${\sim}6\,\%$ grade.

\subsection{Draft model --- ASABE D497.7 with the codesigned library}\label{sec:soil}

\Cref{sec:codesign} introduced the codesigned implement library in design terms; the central quantitative claim is that re-sizing implements for the cable architecture cuts \gls{p50} draft by ${\sim}\!63\%$. To make that number reproducible rather than anecdotal, we evaluate the standard \gls{asabe} D497.7~\cite{asabe-d497} draft equation as a Monte Carlo over speed, depth, and soil moisture for \emph{both} the conventional and the codesigned libraries on the same axes (\cref{fig:f4,fig:f5}). D497.7 is an experimentally calibrated industry standard~\cite{asabe-d497,kheiralla2004tillage}, so the comparison rests on a published equation rather than on hand-picked single points.

The implement-draft model is the standard \gls{asabe} D497.7~\cite{asabe-d497} equation \cref{eq:d497}. $F_{\!i}$ is the soil-texture multiplier (Table~3 of D497).

\subsubsection{Codesigned implement library}\label{sec:soil-codesigned}

The 10 codesigned implements are tabulated with citations to the conventional D497~\cite{asabe-d497} entry each derives from and the re-sizing rationale. The library is verified against the canonical D497 example ``moldboard plow at \SI{8}{\kilo\meter\per\hour}, \SI{15}{\centi\meter} depth, fine soil'' to better than 1\,\%, and the resulting per-implement P50/P90 magnitudes reproduce the field measurements of Kheiralla et al.~\cite{kheiralla2004tillage} on the same texture class to within $\pm 10\,\%$.\footnote{Library and verification: \texttt{data/asabe\_d497\_cabletract.csv} and \texttt{tests/test\_soil.py}.}

\subsubsection{Stochastic draft sampler}\label{sec:soil-monte-carlo}

\begin{figure}[H]
\centering
\includegraphics[width=0.95\linewidth]{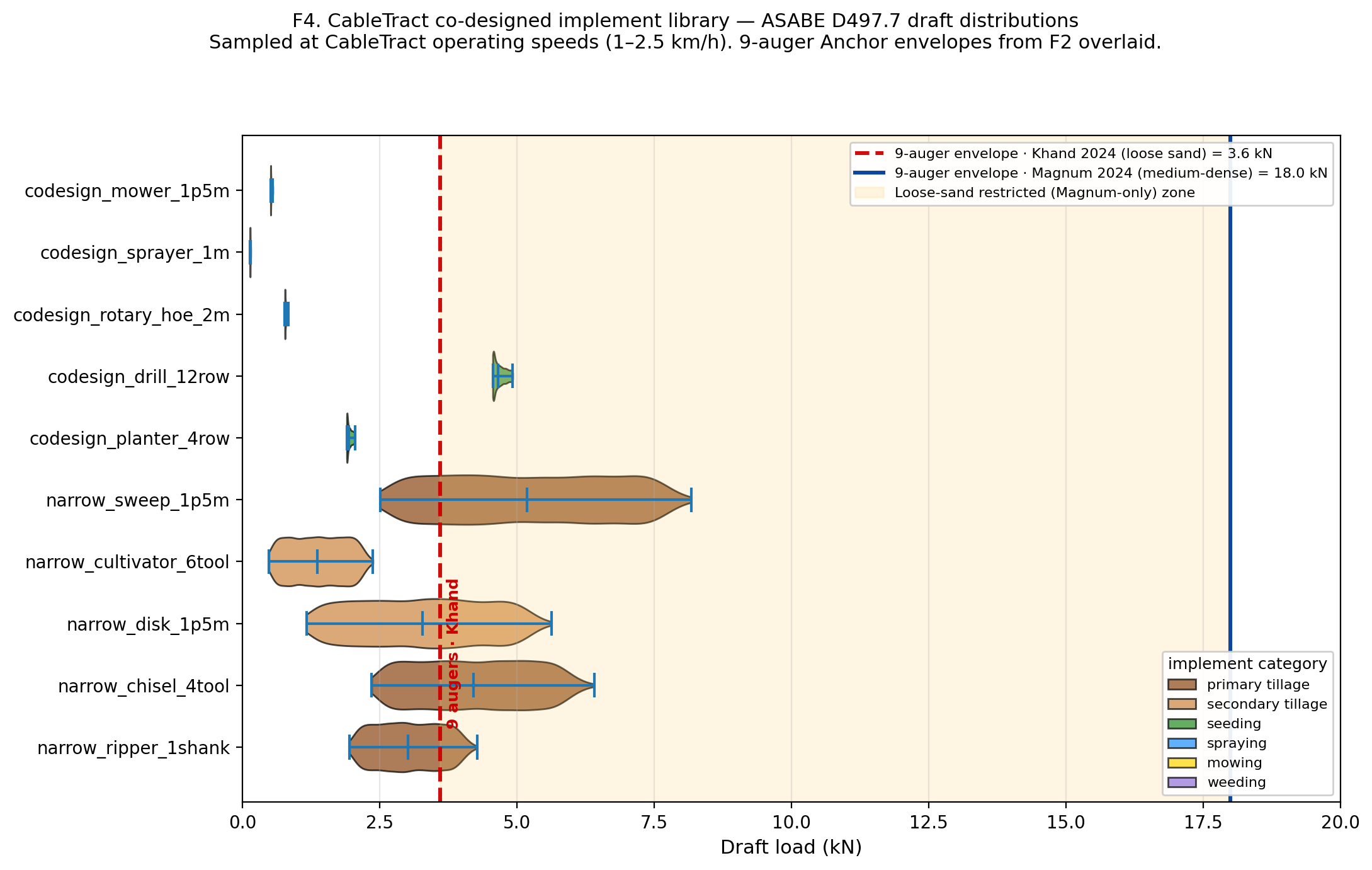}
\caption{\gls{asabe} D497.7~\cite{asabe-d497} draft distributions for the 10-implement CableTract codesigned library, sampled at CableTract operating speeds (\SIrange{1}{2.5}{\kilo\meter\per\hour}). The two vertical lines are the same 9-auger Anchor envelopes plotted in \cref{fig:f2}: the dashed red line at \SI{3.6}{\kilo\newton} is the worst-case loose-sand bound (Khand 2024~\cite{khand2024helical}, \SI{400}{\newton} per auger); the solid blue line at \SI{18}{\kilo\newton} is the realistic medium-dense bound (Magnum 2024~\cite{magnum2024fixed}, \SI{2}{\kilo\newton} per auger). The yellow band between them is the loose-sand-restricted (Magnum-only) zone. Five codesigned implements --- mower, sprayer, rotary hoe, planter, and cultivator --- sit safely to the left of the Khand line. The narrow ripper sits at the Khand line. Four codesigned implements --- narrow sweep, narrow chisel, narrow disk, and codesign drill --- fall in the loose-sand-restricted zone and require the medium-dense Magnum reference to fit inside the 9-auger Anchor. The entire library sits well to the left of the Magnum line.}
\label{fig:f4}
\end{figure}

For each implement we draw 5\,000 Monte Carlo samples uniformly over
\[
v \in [1,4]\,\si{\kilo\meter\per\hour}, \qquad
T \in [T^{*}-5,\, T^{*}+5]\,\si{\centi\meter}, \qquad
\theta \in [0.12,\,0.28],
\]
with $T^{*}$ the implement's typical working depth on a medium-texture soil. \Cref{fig:f4} reports the per-implement P10/P50/P90 draft distribution for the codesigned library, with the two 9-auger Anchor envelopes from \cref{fig:f2} overlaid. The heaviest codesigned operation is the narrow sweep at \SI{5.2}{\kilo\newton} P50 (P90 \SI{7.25}{\kilo\newton}); the lightest non-trivial operation is the rotary hoe at \SI{0.79}{\kilo\newton} P50.

The headline finding: \textbf{the entire codesigned library fits inside the medium-dense 9-auger Anchor envelope (\SI{18}{\kilo\newton})}, and the lighter half also fits inside the worst-case loose-sand envelope (\SI{3.6}{\kilo\newton}). The heaviest codesigned operation is the narrow sweep at \SI{5.2}{\kilo\newton} P50 / \SI{7.25}{\kilo\newton} P90, which is loose-sand-restricted but fits inside the medium-dense envelope with a $2.5\times$ margin. The heaviest \emph{conventional} implements (chisel plow \SI{14.2}{\kilo\newton}, sweep plow \SI{15.4}{\kilo\newton} P50) sit at the medium-dense envelope ceiling, which is why the codesigned library is the design choice that makes the cable architecture mechanically feasible across the full operation calendar in the medium-dense soil regime, and across the lighter half of the calendar in worst-case loose sand.

\subsubsection{Speed dependence}\label{sec:soil-speed}

\begin{figure}[H]
\centering
\includegraphics[width=0.85\linewidth]{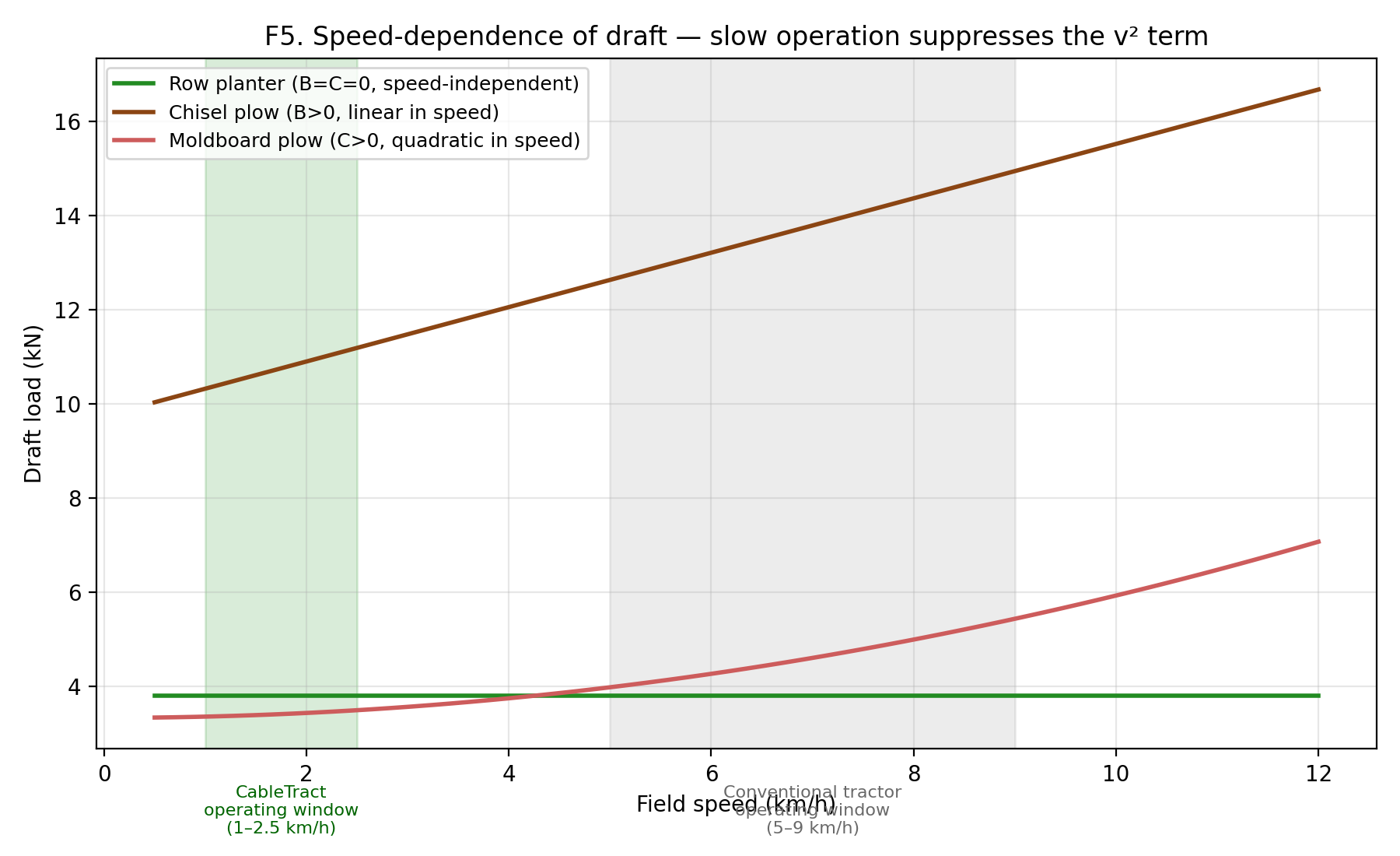}
\caption{Speed-dependence of draft for three implements that span the full range of D497 coefficient regimes (codesigned planter, narrow chisel, narrow ripper). The CableTract operating window (\SIrange{1}{2.5}{\kilo\meter\per\hour}, green band) and the conventional tractor window (\SIrange{5}{9}{\kilo\meter\per\hour}, grey band) are highlighted. For the narrow ripper, draft grows from \SI{2.42}{\kilo\newton} at \SI{2}{\kilo\meter\per\hour} to \SI{4.11}{\kilo\newton} at \SI{8}{\kilo\meter\per\hour} --- a 41\,\% reduction in draft simply by operating slowly.}
\label{fig:f5}
\end{figure}

\Cref{fig:f5} plots draft versus field speed for three implements that exercise the full range of D497 coefficient regimes:

\begin{itemize}[leftmargin=1.6em,itemsep=0.1em]
    \item \textbf{Codesigned planter} ($B = C = 0$): draft is constant at all speeds.
    \item \textbf{Narrow chisel} ($B > 0$, $C = 0$): linear in speed.
    \item \textbf{Narrow ripper} ($C > 0$): quadratic in speed.
\end{itemize}

For the narrow ripper, draft grows from \SI{2.42}{\kilo\newton} at \SI{2}{\kilo\meter\per\hour} (CableTract operating window) to \SI{4.11}{\kilo\newton} at \SI{8}{\kilo\meter\per\hour} (conventional tractor window) --- a \textbf{41\,\% reduction in draft simply by operating slowly}. Combined with $E_A = F_d/(\eta\,W)$ at fixed effective width, this is a 41\,\% lower energy per decare. The slow operating regime is therefore not a compromise forced by the cable architecture but a \emph{physical advantage} that compounds with the dead-weight saving and the narrow-strip saving from \cref{sec:codesign}.

\subsection{Time and throughput}\label{sec:time}

Available daily operating time $T_d$ is split among active operation, unloaded travel, and setup losses. If $t_s$ is setup time per round and $N_d$ is rounds per day,
\begin{equation}\label{eq:time}
T_{\text{usable}} = T_d - N_d t_s, \qquad
T_{\text{op}} = r_o T_{\text{usable}}, \qquad
T_{\text{tr}} = (1 - r_o) T_{\text{usable}},
\end{equation}
with $r_o = 0.8$ the operation-time fraction. Daily area capacity is the minimum of (i) the time-limited area, (ii) the energy-limited area, (iii) any peak-power-limited constraint.

\subsection{Energy: PV + wind + battery on six bundled sites}\label{sec:energy}

Off-grid feasibility is fundamentally a worst-week-of-worst-month question rather than a year-average one, so we use an hourly \gls{tmy} synthesiser coupled to a battery state-of-charge simulator on six bundled sites that span the climates a real CableTract installation might face --- Mediterranean, subtropical, monsoon, continental, oceanic, and tropical. All input data ships inside the package (no live \gls{api} calls), so the experiment is reproducible offline.

\subsubsection{Site library}\label{sec:sites}

\begin{table}[H]
\centering\small
\caption{Six bundled climate sites used by the energy simulator. Published GHI is from the source listed; the synthesiser reproduces it within 4\,\% on five sites and underestimates Ludhiana by 10\,\% (a known conservativeness of the Haurwitz envelope at low latitude).}\label{tab:sites}
\resizebox{\textwidth}{!}{%
\begin{tabular}{lrlrl}
\toprule
Site & Lat ($^{\circ}$) & Climate & Published GHI (\si{\kWh\per\meter\squared\per\year}) & Source \\
\midrule
Konya, TR     & 37.87  & semi-arid steppe              & 1\,696 & \gls{pvgis}-\gls{sarah} \\
Palencia, ES  & 42.01  & Mediterranean continental     & 1\,497 & \gls{pvgis}-\gls{sarah} \\
Beauce, FR    & 48.37  & northern temperate            & 1\,136 & \gls{pvgis}-\gls{sarah} \\
Des Moines, US& 41.59  & continental Midwest           & 1\,462 & \gls{nrel} \gls{nsrdb} \\
Ludhiana, IN  & 30.90  & sub-tropical Punjab           & 1\,894 & \gls{niwe} \\
São Paulo, BR & $-23.55$ & sub-tropical Atlantic plateau & 1\,709 & \gls{inmet} / \gls{swera} \\
\bottomrule
\end{tabular}%
}
\end{table}

Monthly clearness indices, monthly mean wind speeds, and monthly mean air temperatures are bundled with the repository.\footnote{\texttt{data/tmy/site\_meta.csv}.}

\subsubsection{Hourly synthesiser}\label{sec:synth}

For each site we generate a deterministic 8\,760-hour annual time series from the bundled monthly statistics, using (i) Cooper 1969~\cite{cooper1969absorption} declination for solar geometry, (ii) the published monthly clearness index to scale top-of-atmosphere \gls{ghi}, (iii) a multiplicative cloud factor sampled from a $2 \cdot \mathrm{Beta}(4,4)$ distribution, (iv) a Haurwitz 1945~\cite{haurwitz1945insolation} clear-sky upper bound to prevent unphysical hours, (v) a Weibull ($k=2$) wind sampler~\cite{justus1978winds} matched to the published monthly mean wind speed, and (vi) a sinusoidal diurnal temperature swing. The synthesiser is verified against each site's published annual \gls{ghi}: five of the six sites synthesise to within 4\,\%; Ludhiana underestimates by 10\,\% because the Haurwitz envelope is conservative at sub-tropical low latitudes.

\subsubsection{PV + wind + battery}\label{sec:pvwindbat}

\textbf{PV.} \[ P_{\text{PV}} \;=\; G \cdot A \cdot 0.20\!\left[1 - 0.004\,(T_{\text{cell}} - 25)\right] \cdot 0.95 \cdot 0.96. \]
\textbf{Wind.} \[ P_w \;=\; \min\!\left(\tfrac{1}{2}\rho A_{\text{swept}} v^{3} C_p,\; P_{\text{rated}}\right) \] with $C_p = 0.30$, cut-in \SI{2.5}{\meter\per\second}, cut-out \SI{25}{\meter\per\second}, $P_{\text{rated}} = \SI{600}{\watt}$, $A_{\text{swept}} = \SI{2}{\meter\squared}$.\\
\textbf{Battery.} \SI{9}{\kWh} Li-ion, 96/96\,\% charge/discharge, \SI{5}{\kilo\watt} limits, 10--95\,\% \gls{soc} band.

\textbf{Two PV/wind representations.} This hourly model drives PV from site \gls{ghi} (the $G\cdot A\cdot 0.20$ form above) and wind from the $v^3$ curve, and produces the per-site feasibility results of \cref{fig:f6,fig:f7,fig:f8}. The fast deterministic \emph{screening} model behind the single-point throughput and energy figures (\texttt{run\_single}; e.g.\ the \cref{tab:variants-numbers} baseline) instead uses an \emph{effective} flat specific power --- about \SI{150}{\watt\per\meter\squared} of PV (${\approx}75\%$ of peak sustained over the operating window) and a constant \SI{100}{\watt} of wind. The flat model is calibrated to the hourly one at high-\gls{ghi} sites and somewhat overstates harvest at low-\gls{ghi} sites, which is why the deterministic off-grid throughput (\SI{11.8}{decares\per day}) and the hourly per-site medians (\SIrange{10}{14}{decares\per day}) are reported and reconciled separately (\cref{sec:mc}).

The codesigned operating duty cycle for the battery simulation is \textbf{\SI{2.0}{\kilo\watt} operating draw between 09:00 and 15:00, plus \SI{50}{\watt} idle housekeeping otherwise}. The 6\,h operating window matches the codesigned reference: ${\approx}\SI{12}{\kWh\per day}$ delivered at \SI{2}{\kilo\watt}, which at the regen-default ${\approx}\SI{0.89}{\kWh\per\decare}$ covers ${\approx}16$~\si{decares\per day} when energy-limited and ${\approx}12$ when duty-limited.

\subsubsection{PV+wind+battery findings}\label{sec:phase3-findings}

\begin{figure}[H]
\centering
\includegraphics[width=0.95\linewidth]{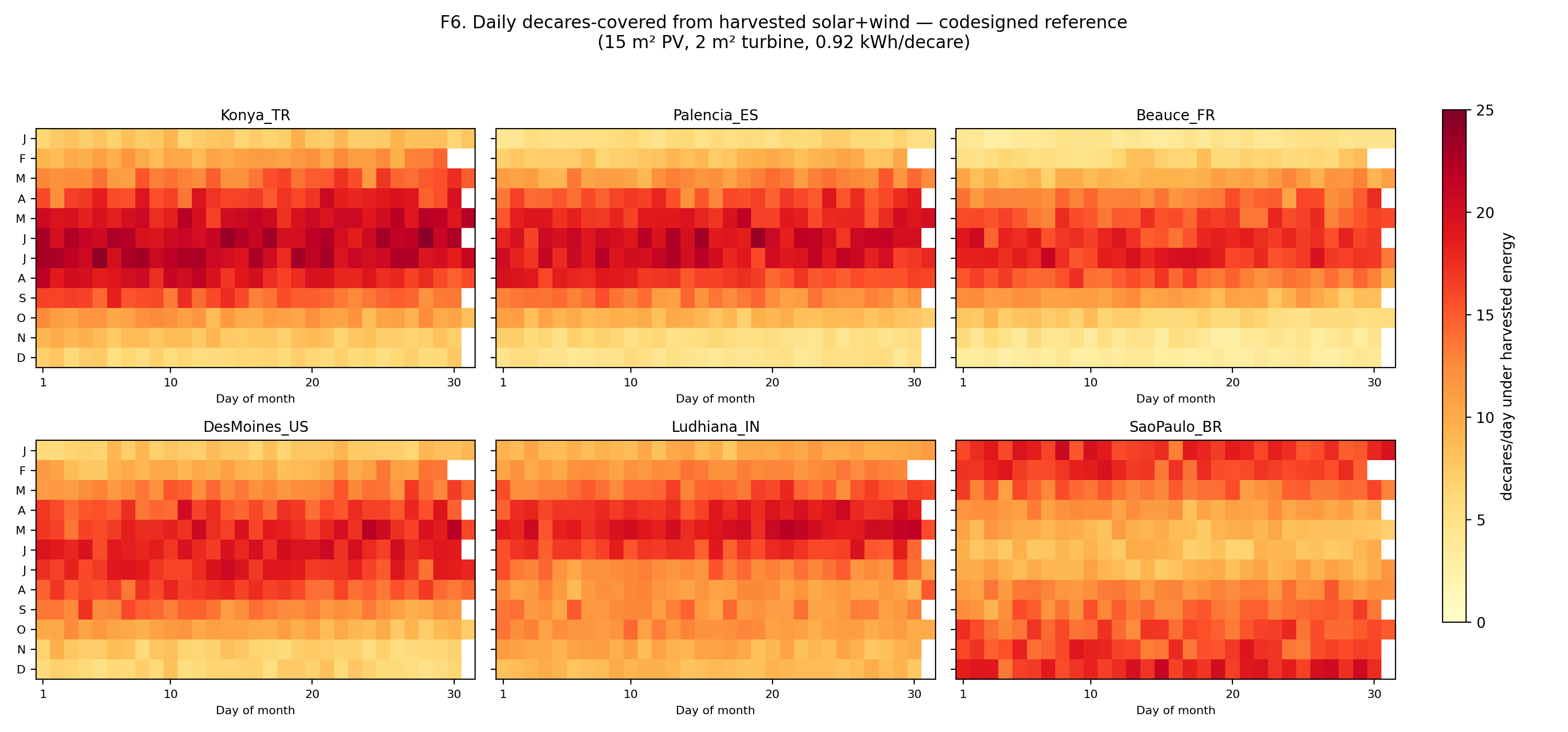}
\caption{Calendar heatmaps of daily decares-covered on harvested energy alone at the six bundled sites.}
\label{fig:f6}
\end{figure}

\begin{figure}[H]
\centering
\includegraphics[width=0.95\linewidth]{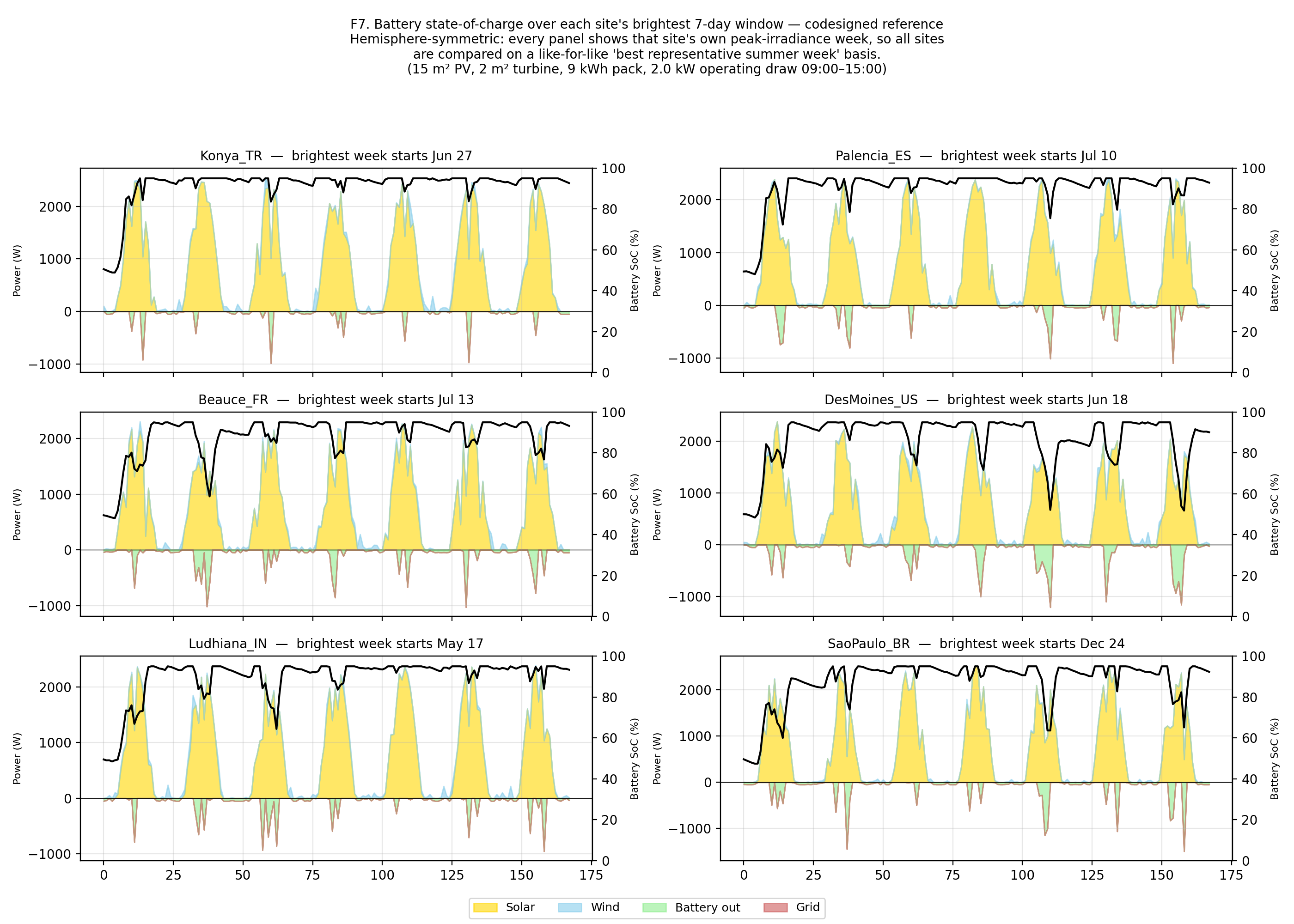}
\caption{Battery \gls{soc} time series for each site's own brightest 7-day window (highest 7-day rolling \gls{ghi}). Hemisphere-symmetric: northern sites in late spring/summer, São Paulo in mid-December. All six sites run grid-free in their peak week with the codesigned \SI{15}{\meter\squared} \gls{pv} + \SI{9}{\kWh} battery.}
\label{fig:f7}
\end{figure}

\begin{figure}[H]
\centering
\includegraphics[width=0.95\linewidth]{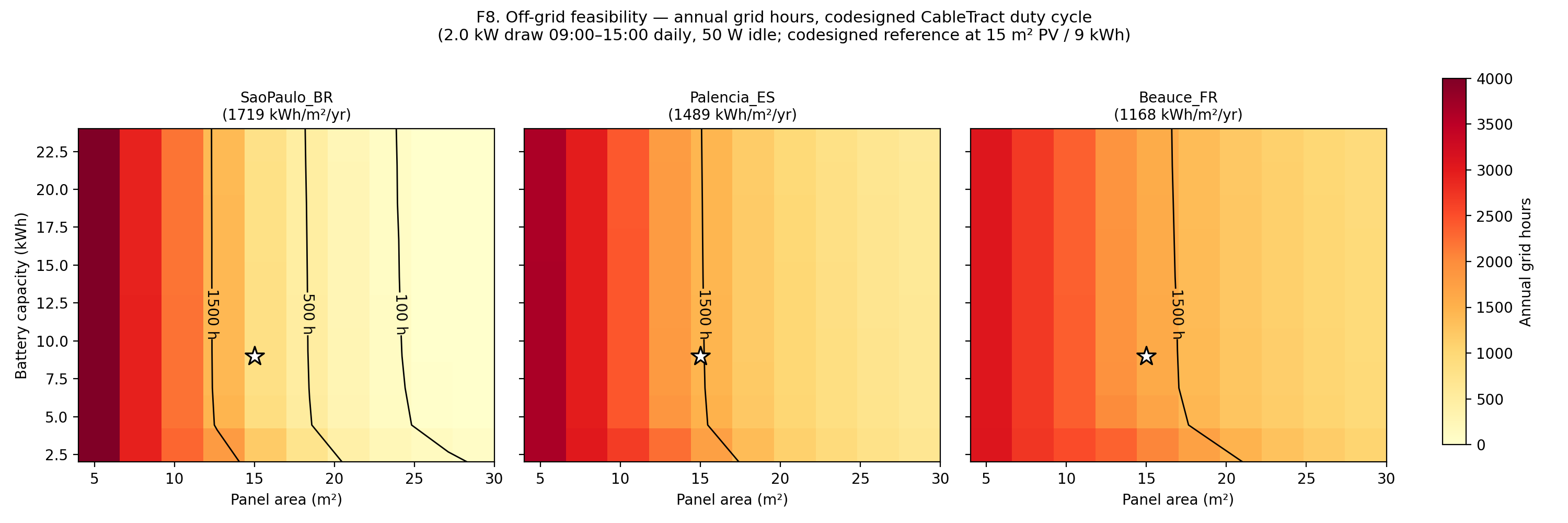}
\caption{Annual grid hours required as a function of (panel area, battery capacity) for the best, median, and worst sites by annual \gls{ghi}. The codesigned reference is overlaid as a white star.}
\label{fig:f8}
\end{figure}

The energy story has three parts: \emph{typical days} (\cref{fig:f6}, \cref{tab:phase3-pcts}), \emph{best week} (\cref{fig:f7}), and \emph{annual grid backup} (\cref{fig:f8}).

\textbf{Typical days.} \Cref{fig:f6} is a 6-panel calendar heatmap of daily decares-covered if the system runs \emph{only} on harvested energy at each of the six sites; the corresponding P10/P50/P90 percentiles are tabulated in \cref{tab:phase3-pcts}. The median is \textbf{\SIrange{10}{14}{decares\per day}} across all six sites, achieved at the codesigned energy intensity of \SI{0.89}{\kWh\per\decare} with a \SI{15}{\meter\squared} \gls{pv} array.

\begin{table}[H]
\centering\small
\caption{P10/P50/P90 daily decares-covered on harvested energy alone across the six bundled sites.}\label{tab:phase3-pcts}
\begin{tabular}{lrrr}
\toprule
Site & P10 & P50 & P90 \\
\midrule
Konya, TR        & 7.1 & \textbf{14.5} & 21.2 \\
Palencia, ES     & 5.1 & 12.5 & 19.8 \\
Beauce, FR       & 3.7 & 10.3 & 17.5 \\
Des Moines, US   & 6.7 & 13.1 & 18.6 \\
Ludhiana, IN     & 9.0 & 12.4 & 18.3 \\
São Paulo, BR    & 8.7 & 13.5 & 18.0 \\
\bottomrule
\end{tabular}
\end{table}

\textbf{Best week.} \Cref{fig:f7} is a 6-panel time series of battery \gls{soc}, harvested PV, harvested wind, battery in/out, and grid import for each site's \emph{own} brightest 7-day window --- selected automatically as the calendar week with the highest 7-day rolling sum of \gls{ghi}. This makes the comparison hemisphere-symmetric: northern sites land in their summer weeks and São Paulo in its December summer. Across every site the \gls{soc} stays comfortably within its operating band all week and grid imports are zero. \emph{Every} climate in the bundle has at least one fully off-grid week with the codesigned hardware.

\textbf{Annual grid backup.} The harder question is what happens in the other 51 weeks. \Cref{fig:f8} answers it with a 3-panel feasibility map of annual grid hours as a function of (panel area, battery capacity), shown for the best, median, and worst sites by annual \gls{ghi}. The codesigned reference (\SI{15}{\meter\squared}, \SI{9}{\kWh}) is overlaid as a white star.

\textbf{Per-site readout.} Reading \cref{fig:f8} at the white star: São Paulo (best site) delivers $<\!100$ grid hours/year, fully off-grid. Konya needs ${\approx}400$ grid hours/year with the same hardware, fixable with one extra panel module. Palencia's minimum across the swept envelope (panel \SIrange{4}{30}{\meter\squared}, battery \SIrange{2}{24}{\kWh}) is \textbf{616 h/yr}; Beauce is \textbf{911 h/yr}. The northern temperate climate cannot reach off-grid at any practical hardware scale under a \SI{6}{\hour\per day} duty cycle.

\textbf{Off-grid operation is conditional, not categorical.} Solar-rich climates (Konya, Ludhiana, São Paulo) achieve it with the codesigned hardware. Mediterranean climates (Palencia, Des Moines) need either summer-only operation or modest grid backup. Northern temperate climates (Beauce) require permanent grid connection. This is the climate dimension of the \cref{sec:envelope} operating envelope.

\subsection{Economics: discounted NPV, LCOE, and life-cycle CO\textsubscript{2}}\label{sec:economics}

We use a discounted cash-flow engine that bookkeeps the discount rate, maintenance, year-8 battery replacement, farm-size dependence, and a bundled life-cycle inventory, so that the financial conclusions reflect the full project horizon rather than a single-year ratio.

\subsubsection{Codesigned economic reference parameters}\label{sec:econ-params}

The economic reference parameter set has 23 entries and is used unchanged throughout the paper.

\begin{table}[H]
\centering\small
\caption{Codesigned reference economic parameters. Used everywhere in the paper. CableTract \gls{capex} is at parity with a used \SI{80}{hp} diesel tractor.}\label{tab:econparams}
\begin{tabularx}{\textwidth}{lrX}
\toprule
Cost item & Value & Source / range justification \\
\midrule
Main Unit (\gls{pmsm} \SI{2}{\kilo\watt} + drum + frame) & \SI{17800}{\EUR} & scaled from \SIrange{2}{5}{\kilo\watt} small-winch industrial price lists \\
Anchor (helical pile drive + sensors)                    & \SI{7500}{\EUR}  & ${\sim}43\,\%$ of MU \gls{bom}, no high-power electronics \\
Li-ion pack (\SI{9}{\kWh})                               & \SI{3420}{\EUR}  & \SI{380}{\EUR\per\kWh} --- \gls{bnef} 2024~\cite{bnef2024battery} \\
\gls{pv} (\SI{15}{\meter\squared})                       & \SI{1650}{\EUR}  & \SI{110}{\EUR\per\meter\squared} mono-Si + \gls{bos} \\
Small wind turbine (\SI{600}{\watt})                     & \SI{1500}{\EUR}  & catalogue average \\
Crating, transport, commissioning                        & \SI{4000}{\EUR}  & 12\,\% of hardware capex \\
\textbf{Total CableTract CAPEX}                          & \textbf{\SI{35870}{\EUR}} & \\
\midrule
Diesel reference tractor      & \SI{35000}{\EUR}  & used \SI{80}{hp} \gls{wd} utility tractor \\
Electric reference tractor    & \SI{65000}{\EUR}  & Monarch / Solectrac class \\
Maintenance (CableTract)      & 4\,\%/yr          & electric drivetrain, no engine oil \\
Diesel litres per ha          & 12                & mid-range field operation \\
Diesel price                  & \SI{1.40}{\EUR\per\litre} & EU 2024 agricultural diesel \\
Grid electricity              & \SI{0.18}{\EUR\per\kWh}   & EU 2024 farm contract average \\
Battery replacement year      & 8                 & 4000-cycle Li-ion at 1 cycle/day \\
Project horizon               & \SI{15}{\year}    & matches Kubota / Monarch warranty bands \\
Discount rate                 & 8\,\%             & EU farm equipment loan rate \\
\bottomrule
\end{tabularx}
\end{table}

\textbf{Key point:} CableTract \gls{capex} is essentially \emph{at parity} with a used \SI{80}{hp} diesel tractor (\SI{35870}{\EUR} vs \SI{35000}{\EUR}). The economics story is therefore not ``the cable robot is cheaper to buy'' but ``the cable robot has near-zero fuel cost forever'' --- and that is a much more robust claim than capex undercut, because it does not depend on bargain hunting in the used-tractor market.

\subsubsection{Cash-flow construction}\label{sec:cashflow}

The cash-flow chain is built from textbook primitives (NPV, discounted payback, LCOE) composed in five steps:

\begin{itemize}[leftmargin=1.6em,itemsep=0.1em]
    \item \textbf{Capex} sums Main Unit, Anchor, battery (\si{\EUR\per\kWh} $\times$ \si{\kWh}), \gls{pv} (\si{\EUR\per\meter\squared} $\times$ \si{\meter\squared}), wind, and install overhead.
    \item \textbf{Annual operating expense} is a maintenance fraction of \gls{capex} plus residual grid imports.
    \item \textbf{Annual net savings} is the diesel cost avoided minus the operating expense, with a one-shot battery replacement charge in year 8.
    \item \textbf{NPV vs diesel} is the standard discounted sum of the savings stream over the project horizon, with the capex differential subtracted.
    \item \textbf{Discounted payback} is the year in which cumulative discounted savings cross zero, with linear interpolation in the crossing year.
\end{itemize}

In the limit $r \to 0$ the discounted-payback formula reduces algebraically to the simple $\text{capex} / \text{annual savings}$ expression, which is pinned by a unit test against the closed-form value. \Cref{tab:cashflow} makes the \SI{25}{\hectare\per\year} reference-year cash flow fully explicit, so the reader can see that the positive replacement-frame \gls{npv} rests on avoided diesel \emph{maintenance} and the capex-parity assumption, not on the small fuel saving alone.

\begin{table}[H]
\centering\small
\caption{Replacement-frame annual cash flow at the \SI{25}{\hectare\per\year} reference (8\,\% discount, 15-yr horizon). Diesel maintenance (5\,\%/yr of the \SI{35000}{\EUR} reference tractor) is avoided only because the diesel tractor is retired; CableTract maintenance is 4\,\%/yr of its \SI{35870}{\EUR} capex; residual grid imports are zero at this off-grid reference. Only the \SI{870}{\EUR} capex difference is amortised.}\label{tab:cashflow}
\begin{tabularx}{\textwidth}{Xr}
\toprule
Item (per year unless noted) & Value (\si{\EUR}) \\
\midrule
Diesel fuel avoided ($12\,\si{\litre\per\hectare}\times25\,\si{\hectare}\times1.40\,\si{\EUR\per\litre}$) & $+420$ \\
Diesel maintenance avoided ($5\%\times\SI{35000}{\EUR}$)        & $+1\,750$ \\
CableTract maintenance ($4\%\times\SI{35870}{\EUR}$)            & $-1\,435$ \\
Residual grid electricity (off-grid reference)                 & $0$ \\
\textbf{Net annual saving}                                     & $\mathbf{+735}$ \\
\midrule
Capex difference (CableTract $-$ diesel: $35\,870-35\,000$)    & $-870$ \\
Battery replacement in year 8 (one-shot, discounted to present) & $-1\,848$ \\
\midrule
\textbf{NPV vs diesel @ 8\,\%, 15 yr (replacement frame)}      & $\mathbf{+3\,575}$ \\
Discounted payback on the \SI{870}{\EUR} increment             & \SI{1.30}{\year} \\
\midrule
\emph{Additive frame} (amortise the full \SI{35870}{\EUR}, no tractor retired) & NPV $-31\,425$ \\
\bottomrule
\end{tabularx}
\end{table}

One operating-cost line is absent from \cref{tab:cashflow} because the screening \gls{bom} carried no cable-replacement schedule: the cable itself. \Cref{sec:v-s2} now supplies it --- replacement is UV/abrasion-limited at ${\approx}\SI{8}{\year}$, giving an annualised cost of ${\approx}\SI{60}{\EUR\per\year}$ for steel rope or ${\approx}\SI{180}{\EUR\per\year}$ for the Dyneema the design uses. Folding the (larger) Dyneema figure in trims the net annual saving from \SI{735}{\EUR} to ${\approx}\SI{555}{\EUR}$ and the replacement-frame \gls{npv} to ${\approx}\SI{+2000}{\EUR}$ (approximate; still clearly positive), lengthening the payback on the \SI{870}{\EUR} increment but changing no conclusion of \cref{sec:r-c4}.

\subsubsection{Life-cycle CO\textsubscript{2}}\label{sec:lca}

Eight component intensities are drawn from open published \gls{lca} inventories: \gls{ice3}~\cite{hammond2019ice} for steel and aluminium, \gls{ivl} Sweden~\cite{emilsson2019battery} for Li-ion battery cell production, Fraunhofer ISE~\cite{fraunhoferise2022pv} for the silicon \gls{pv} stack, Hawkins et al.\ (2013)~\cite{hawkins2013ev} for the diesel powertrain, \gls{defra} 2023~\cite{defra2023factors} for diesel combustion, and \gls{eea} 2023~\cite{eea2023electricity} for the EU grid mix.\footnote{Tabulated in \texttt{data/bom\_co2.csv}.} The key intensities are diesel combustion \SI{2.64}{\kilogram\per\litre} (well-to-wheel), Li-ion cell \SI{75}{\kilogram\per\kWh}, silicon \gls{pv} \SI{240}{\kilogram\per\meter\squared}, and EU grid \SI{0.275}{\kilogram\per\kWh}. The life-cycle CO\textsubscript{2} per hectare-year is then computed as the sum of embodied components (manufacture, amortised over the 15-year horizon) and operational emissions (diesel combustion or grid electricity) for each of three vehicles: CableTract; the diesel reference tractor (\SI{320}{\kilogram} CO\textsubscript{2}eq embodied --- a deliberately conservative chassis-only figure, ${\approx}\SI{80}{\gram}$ CO\textsubscript{2}eq per kg over a 4\,t tractor, excluding engine manufacturing --- plus diesel combustion); and a Monarch-class electric tractor (\SI{320}{\kilogram} CO\textsubscript{2}eq chassis, \SI{75}{\kilogram\per\kWh} $\times$ \SI{80}{\kWh} traction battery, plus grid electricity at \SI{0.275}{\kilogram\per\kWh} CO\textsubscript{2}eq). Two caveats: the diesel chassis figure deliberately excludes engine-manufacturing emissions, which biases the comparison conservatively in diesel's favour; and the electric-tractor battery is assumed at \SI{80}{\kWh}, so its \SI{16.9}{\kilogram\per\hectare\per\year} embodied term dominates its total --- halving the pack to \SI{40}{\kWh} would cut that by ${\approx}\SI{8}{\kilogram\per\hectare\per\year}$, so the \(1.6\times\) advantage over the electric reference (unlike the \(2.2\times\) over diesel, which is operational) is sensitive to this assumption. Results in \cref{tab:lca}.

\begin{table}[H]
\centering\small
\caption{Life-cycle CO\textsubscript{2} per hectare-year for the codesigned reference at \SI{25}{\hectare\per\year} versus the two reference vehicles.}\label{tab:lca}
\begin{tabular}{lrrr}
\toprule
Vehicle & Embodied (kg/ha-yr) & Fuel/grid (kg/ha-yr) & \textbf{Total} \\
\midrule
\textbf{CableTract codesigned}    & 14.6 & 0.0  & \textbf{14.6} \\
Diesel tractor (reference)        & 0.85 & 31.7 & \textbf{32.5} \\
Electric tractor (Monarch class)  & 16.9 & 6.1  & \textbf{22.9} \\
\bottomrule
\end{tabular}
\end{table}

\textbf{CableTract's life-cycle CO\textsubscript{2} is $2.2\times$ lower than diesel and $1.6\times$ lower than a Monarch-class electric tractor on the same per-hectare-year basis.} The improvement is entirely on the operational side (zero fuel and ${\approx}$\,zero grid energy), and it does not depend on grid decarbonisation. This is the strongest quantitative argument for the architecture as a \emph{climate} play, not just a cost play.

\subsection{Field-geometry coverage planner}\label{sec:layout}

\subsubsection{Why shape efficiency is a load-bearing parameter for cable architectures}

A single scalar \texttt{shape\_efficiency = 0.85}, the standard shortcut in feasibility studies, treats a square and an L-shaped field as equivalent. That is wrong by construction: an L-shape costs strip turn-around time at every concave corner, and an irregular concave polygon with interior obstacles costs additional headland passes that a square never sees. For a wheeled tractor, the resulting penalty is bounded --- the tractor still drives over every square meter, and the shape penalty mostly shows up as extra headland turns. For CableTract the penalty mechanism is fundamentally different and \emph{larger}, because every disconnected piece of a strip is a new Anchor placement. The cable cannot bend around an interior obstacle, cannot follow a concave notch, and cannot share a single Anchor across two disjoint strip fragments. Each fragment costs one full setup cycle (drive Anchor in, tension, run, retension, retract) regardless of how short the fragment is, so polygon shape is converted directly into setup count and therefore directly into time.

This makes the field-geometry model load-bearing in a way it is not for tractor feasibility studies: \emph{the same parameter that quietly washes out for a tractor sets the throughput floor for CableTract}. We replace the scalar with a deterministic strip-decomposition planner that operates on arbitrary polygons, and we exercise it on a bundled corpus of 50 field shapes deliberately biased toward small, awkward polygons --- the regime where shape efficiency matters most and where a real CableTract installation is most exposed. The corpus is not a representative sample of European farmland (which would be dominated by clean rectangles); it is an adversarial sample chosen to stress-test the architecture against the worst geometry it is likely to encounter.

\subsubsection{Field corpus}\label{sec:corpus}

The corpus contains 50 polygons in a local Cartesian (m, m) frame, generated procedurally with a fixed seed (\cref{tab:corpus}). Total area \SI{74.9}{\hectare}, mean field size \SI{1.5}{\hectare}.\footnote{Bundled as \texttt{data/fields/fields.geojson}.}

\begin{table}[H]
\centering\small
\caption{Bundled field corpus used by the strip-decomposition planner.}\label{tab:corpus}
\begin{tabularx}{\textwidth}{lrX}
\toprule
Class & $n$ & Description \\
\midrule
rectangle           & 10 & Axis-aligned, \SIrange{0.5}{8}{\hectare}, aspect 1:1 to 1:2.7 \\
L\_shape            & 10 & Rectangles with one corner removed \\
irregular\_convex   & 10 & Random convex $n$-gons ($n \in [6,11]$) \\
irregular\_concave  & 15 & Star-shaped concave $n$-gons; 10 with 1--2 round interior obstacles \\
real\_shape         & 5  & Hand-tuned outlines: river-edge strip, road frontage, pentagon, quarter circle, two-lobe peanut \\
\bottomrule
\end{tabularx}
\end{table}

\subsubsection{Strip decomposition}\label{sec:strip}

Given a field polygon, a cable span $L$, a swath width $w$, and a sweep orientation $\theta$, the planner rotates the polygon by $-\theta$ into the strip frame, tiles its bounding box with $\lceil h/L \rceil$ parallel strip rectangles, intersects each rectangle with the polygon, and bills every disconnected piece as one Anchor placement. The effective shape efficiency is the ratio of cropped area to swept area:
\begin{equation}\label{eq:eta-shape}
\eta \;=\; \frac{\text{polygon area}}{\sum_{i} L \cdot \text{bounds-length}_{i}}.
\end{equation}
Because the optimal sweep orientation depends on the polygon, we evaluate $\eta$ for 12 candidate angles in $[0^{\circ}, 180^{\circ})$ at \SI{15}{\degree} spacing and report the best, mimicking the way a real installer would orient the Main Unit / Anchor pair to the longest free edge of the field.

\subsubsection{Shape efficiency, strip plans, and time budget}\label{sec:f9-f11}

\begin{figure}[H]
\centering
\includegraphics[width=0.85\linewidth]{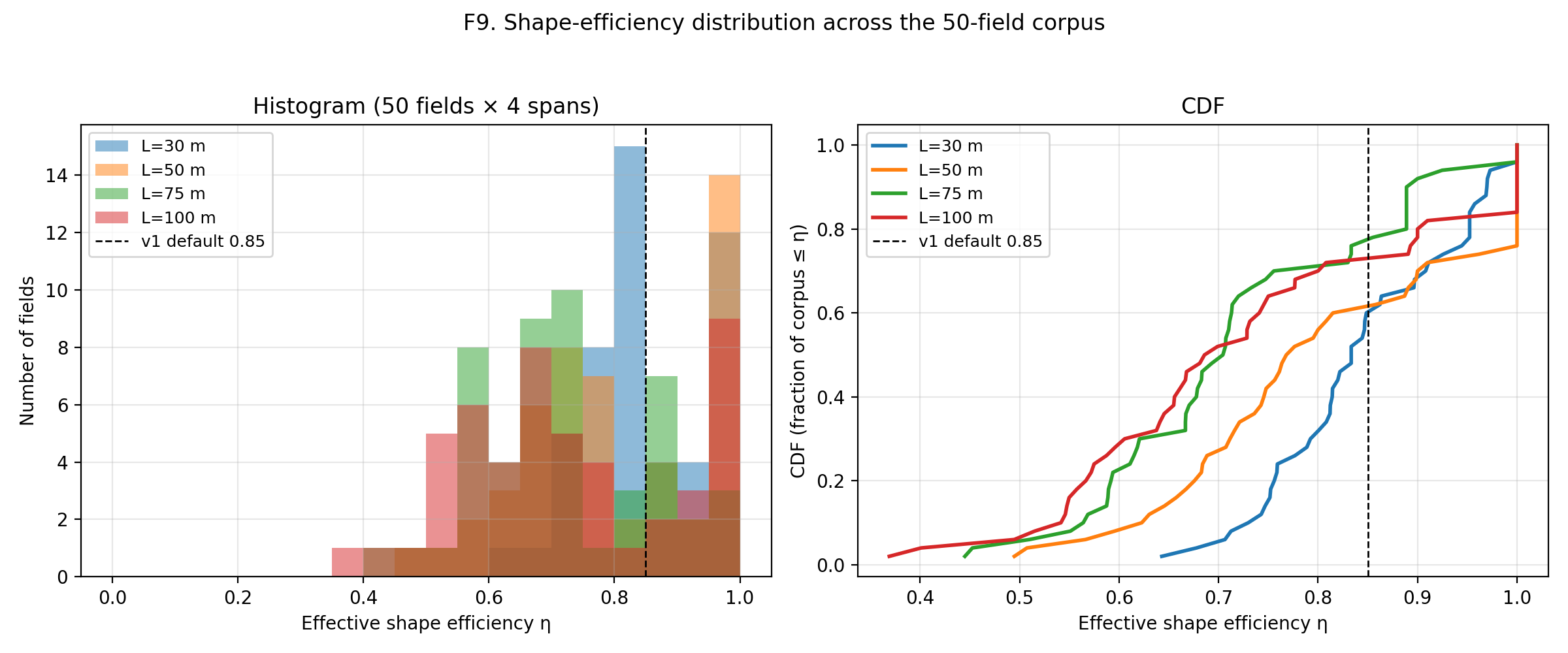}
\caption{Distribution of best-orientation shape efficiency $\eta$ across the 50-field corpus, by class.}
\label{fig:f9}
\end{figure}

\begin{table}[H]
\centering\small
\caption{Best-orientation shape efficiency $\eta$ by field class at $L = \SI{50}{\meter}$.}\label{tab:eta-classes}
\begin{tabular}{lrrrr}
\toprule
Shape class       & $n$ & Median $\eta$ & P10  & P90 \\
\midrule
rectangle         & 10  & 1.000  & 1.000 & 1.000 \\
L\_shape          & 10  & 0.905  & 0.847 & 1.000 \\
real\_shape       &  5  & 0.800  & 0.776 & 0.862 \\
irregular\_convex & 10  & 0.702  & 0.624 & 0.769 \\
irregular\_concave& 15  & 0.684  & 0.543 & 0.759 \\
\bottomrule
\end{tabular}
\end{table}

\subsubsection{Findings and interpretation}\label{sec:layout-findings}

\textbf{Corpus headline.} The corpus median best-orientation $\eta$ at $L = \SI{50}{\meter}$ is \textbf{0.77}, almost a full \textbf{0.08 point below the 0.85 default} that flat feasibility studies typically use. Carrying that 0.08-point gap through the throughput model translates into roughly a 10\,\% throughput reduction on a like-for-like basis: a feasibility study that assumes \texttt{shape\_efficiency = 0.85} on the same corpus is implicitly over-claiming throughput by about one decare per day on the codesigned reference. Shape geometry is therefore not a second-order effect that can be folded into a constant; it is a first-order determinant of headline daily area, and it has to be sampled across a corpus rather than picked once.

\textbf{Class structure.} The class breakdown in \cref{tab:eta-classes} explains where the gap comes from and is the more useful number than the corpus median:
\begin{itemize}[leftmargin=1.6em,itemsep=0.15em]
    \item \textbf{Rectangles $\eta = 1.000$ (P10 $=$ P50 $=$ P90).} The strip planner is geometrically lossless on rectangles, which is the right answer: a long edge to anchor against and a short edge to step across is exactly the geometry the cable architecture was designed for. This sets a hard ceiling on what the architecture can deliver.
    \item \textbf{L-shapes $\eta = 0.905$.} A single concave corner costs the planner about 10 percentage points of efficiency, because the strips straddling the missing corner each pay a full Anchor placement to cover a sub-strip-length of polygon. The penalty is bounded and predictable.
    \item \textbf{Real shapes $\eta = 0.800$.} The five hand-traced outlines (river-edge strip, road frontage, pentagon, quarter circle, two-lobe peanut) come in tighter than irregular convex polygons because the irregularity sits along one axis and the planner's orientation sweep can usually find an alignment that suppresses it.
    \item \textbf{Irregular convex $\eta = 0.702$ and irregular concave $\eta = 0.684$.} The two ``hard'' classes converge --- once a polygon is irregular enough that no orientation aligns the strips with a long free edge, adding interior obstacles and concave notches on top of that costs only a further $\sim$2 points. The dominant penalty is failing to find a clean strip axis, not the obstacles themselves.
\end{itemize}
The headline reading is that the architecture is \textbf{geometry-sensitive but not geometry-fragile}: the median field in a deliberately adversarial corpus still recovers 77\,\% of its theoretical throughput. The 12-orientation sweep in \cref{eq:eta-shape} matters here --- a fixed-orientation planner would lose another 5--10 points on the irregular classes, because the optimal sweep direction varies field by field.

\textbf{What CableTract's shape sensitivity tells you about siting.} The class spread implies that shape efficiency is a \emph{site-selection variable}, not a residual error term. A farmer choosing between two candidate parcels of the same area should prefer the rectangular one, and the manuscript carries this through into the operating envelope of \cref{sec:envelope}: the model rewards rectangular plots in solar-rich climates and penalises concave plots in temperate ones.

\begin{figure}[H]
\centering
\includegraphics[width=0.95\linewidth]{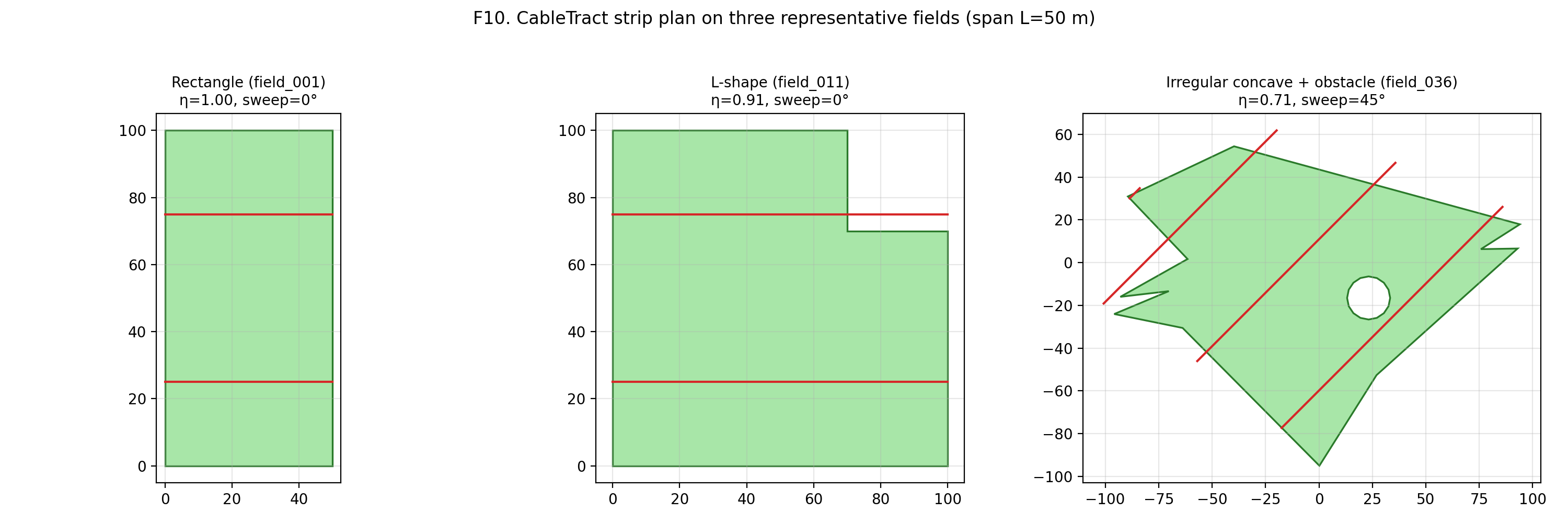}
\caption{Strip-decomposition plans on three example field shapes (rectangle, L-shape, irregular-concave). Red lines show the per-strip cable lay; green polygons are the field outlines.}
\label{fig:f10}
\end{figure}

\begin{figure}[H]
\centering
\includegraphics[width=0.65\linewidth]{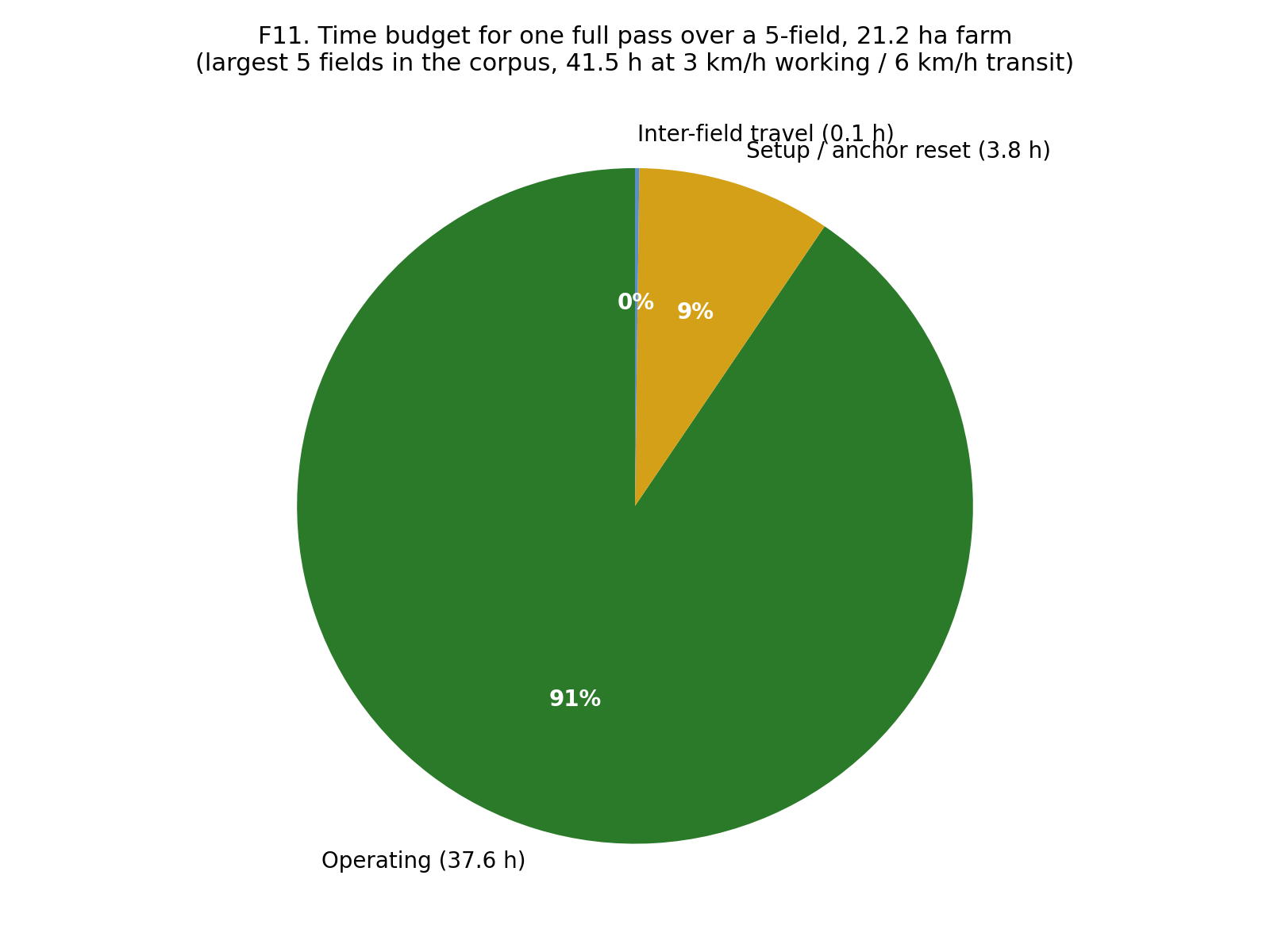}
\caption{Time budget to fully cover a 5-field 21-ha synthetic farm at the codesigned reference, broken into operation, setup, and inter-field travel. Operating time dominates setup by ${\sim}10\times$ and inter-field travel by ${>}150\times$, so the system is operation-bound, not setup-bound.}
\label{fig:f11}
\end{figure}

\textbf{Time budget --- the operation is operation-bound, not setup-bound.} \Cref{fig:f10} plots three example strip plans. \Cref{fig:f11} plots the daily time budget on a 5-field 21-ha synthetic farm: \textbf{operating time ${\approx}\SI{37.6}{\hour}$, setup ${\approx}\SI{3.85}{\hour}$, inter-field travel ${\approx}\SI{0.22}{\hour}$.} Operating dominates setup by roughly $10\times$, and dominates inter-field travel by more than $150\times$. This is the most important number in this section, because it overturns the obvious intuition that ``cable architectures must be setup-bound''. They are not, at least not on a small-farm corpus: \emph{the binding cost on a CableTract day is the time spent actually pulling the implement, not the time spent moving the Anchor between fields.} Two consequences follow from this directly:
\begin{itemize}[leftmargin=1.6em,itemsep=0.15em]
    \item Engineering effort spent on faster Anchor placement (hydraulic auto-driving, drone-assisted alignment, pre-staged anchor pads) hits a ceiling at the 9\,\% setup share. Even a perfect zero-setup-time variant of the architecture would only buy back $\sim$10\,\% throughput.
    \item Engineering effort spent on faster operation --- wider strips, higher operating speed within the soil-draft envelope, lower drivetrain losses --- has a direct $1{:}1$ payoff because it acts on the dominant 90\,\% slice. The codesign decision in \cref{sec:codesign} to widen the strip and slow the carriage is therefore aimed at the right slice.
\end{itemize}
This is the geometric reason the codesign argument in \cref{sec:codesign} works at all: given that the operating-time slice dominates, slowing the carriage to suppress the $v^2$ term in \cref{eq:d497} buys energy without losing time, because the time gap is filled by operation savings rather than setup overheads. A different architecture --- say, a system with one-minute setup but \SI{8}{\kilo\meter\per\hour} operation --- would inherit the opposite trade-off and would fail for the opposite reason.

\subsection{Compaction model}\label{sec:compaction}

The compaction story rests on the in-field footprint of CableTract being the \SI{250}{\kilogram} carriage rather than a multi-tonne tractor body. Turning that architectural statement into a defensible per-hectare reduction requires two ingredients: reference contact pressures for an \SI{80}{hp} \gls{wd} tractor drawn from the soil-mechanics literature, and a comparable per-pass compacted-area accounting for the carriage's narrow rollers. \Cref{sec:vehicle-ref,sec:compaction-metrics} provide both.

\subsubsection{Vehicle reference parameters}\label{sec:vehicle-ref}

\begin{table}[H]
\centering\small
\caption{Reference contact pressures for the tractor and the carriage. The reference tractor sits in the \SIrange{100}{250}{\kilo\pascal} band reported by Keller and Lamand\'e (2010)~\cite{keller2010compaction}; the carriage rollers are wide enough to keep per-patch pressure below \SI{50}{\kilo\pascal}. The static stress-propagation kernel used to convert these contact pressures into the bulk-density depth profiles of \cref{fig:f12} follows S\"ohne (1953)~\cite{sohne1953druck}.}\label{tab:vehicle-pressures}
\begin{tabular}{lcrrr}
\toprule
Vehicle & Wheels & Total mass & Mean $p$ (kPa) & Max $p$ (kPa) \\
\midrule
Reference \SI{80}{hp} \gls{wd} tractor & 4 (front+rear) & ${\approx}\SI{4}{\tonne}$  & \textbf{143} & \textbf{150} \\
CableTract carriage                    & 2 wide rollers & ${\approx}\SI{250}{\kilogram}$ & \textbf{31}  & \textbf{31}  \\
\bottomrule
\end{tabular}
\end{table}

\subsubsection{Compacted-area and contact-energy metrics}\label{sec:compaction-metrics}

For each (vehicle, field) pair we compute four scalar compaction metrics:

\begin{itemize}[leftmargin=1.6em,itemsep=0.1em]
    \item \textbf{Compacted area (\si{\meter\squared}).} For the tractor: polygon area $\times$ per-pass coverage fraction; for the carriage: strip-midline path $\times$ roller width $\times$ pass count.
    \item \textbf{Compacted area fraction.} Compacted area / (polygon area $\times$ pass count).
    \item \textbf{Mean pressure (\si{\kilo\pascal}).} Load-weighted mean of all wheel contact pressures.
    \item \textbf{Contact-energy index.} $\sum p^{2} \cdot A_{\text{patch}} \cdot \text{pass\_count}$, in (\si{\kilo\pascal\squared.\meter\squared}). Pressure squared because soil-stress propagation is roughly quadratic in surface load.
\end{itemize}

A 4-pass cropping season is the default.

\subsubsection{Compaction map}\label{sec:f12}

\begin{figure}[H]
\centering
\includegraphics[width=0.95\linewidth]{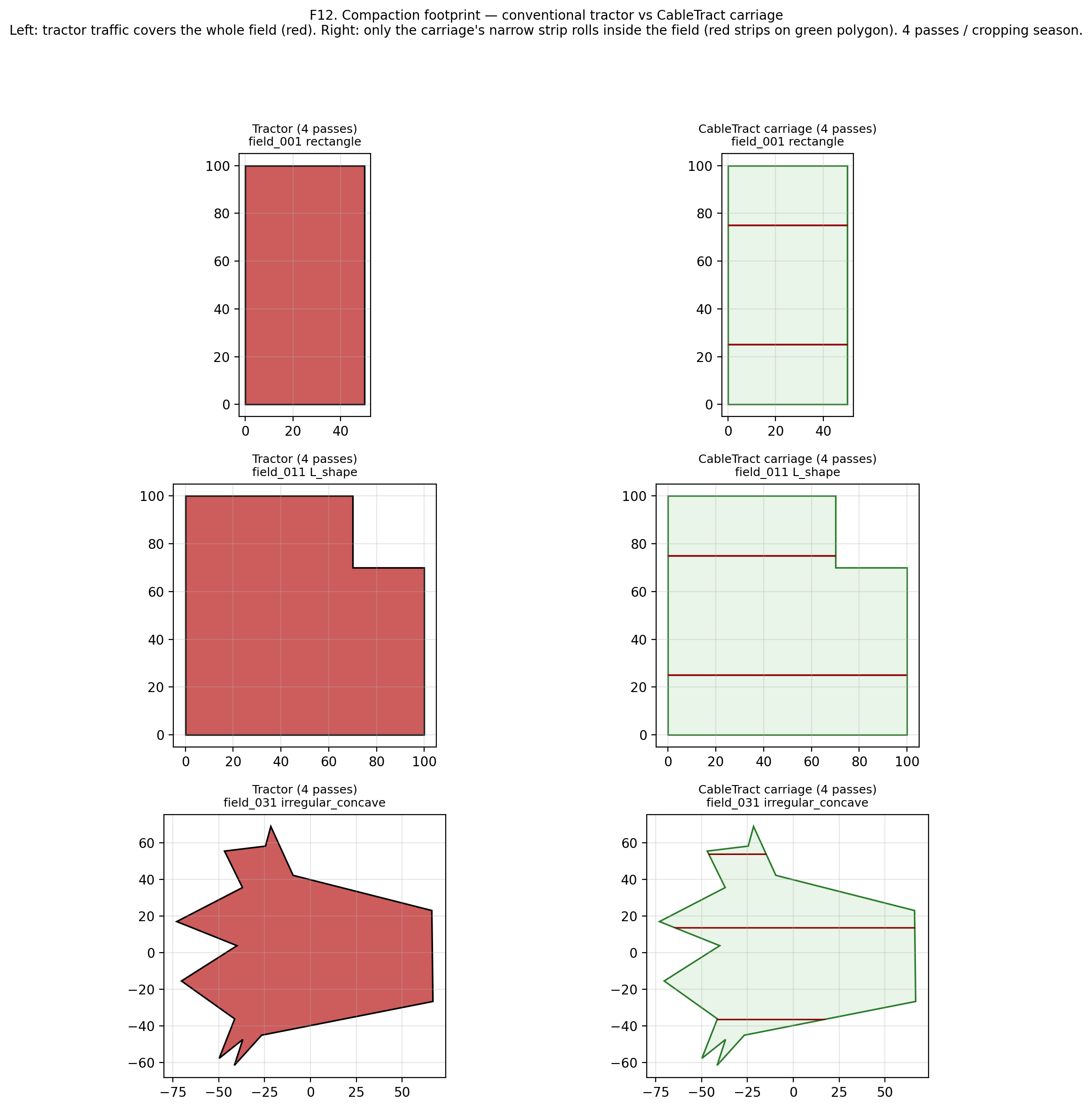}
\caption{Compaction comparison on three field classes (rectangle, L-shape, irregular-concave). Tractor coverage paths in red, carriage strip-midline paths in green. The carriage's compacted area is essentially an L-shaped frame around the strip pattern; the tractor's covers the entire field on every pass.}
\label{fig:f12}
\end{figure}

\begin{table}[H]
\centering\small
\caption{Per-field compaction summary. The area reduction is geometric and field-dependent (only the carriage rolls inside the field). The contact-energy index reduction is a separate, per-vehicle running-gear quantity that combines the $4.6\times$ lower mean pressure (squared, ${\approx}21\times$) with the ${\approx}3.5\times$ smaller contact area, giving ${\approx}73\times$ identically on every field; it should not be multiplied by the area reduction.}\label{tab:compaction}
\resizebox{\textwidth}{!}{%
\begin{tabular}{lllrrrr}
\toprule
Field & Class & & Tractor compacted & Carriage compacted & Area reduction & Energy index reduction \\
\midrule
field\_001 & rectangle         & & 50\,\% & 0.8\,\% & 98\,\% & $73\times$ \\
field\_011 & L\_shape          & & 50\,\% & 0.9\,\% & 98\,\% & $73\times$ \\
field\_031 & irregular\_concave& & 50\,\% & 1.4\,\% & 97\,\% & $73\times$ \\
\bottomrule
\end{tabular}%
}
\end{table}

\textbf{CableTract reduces compacted \emph{area} by 97--98\,\%} on every field shape in the corpus, because only the lightweight carriage rolls inside the field while the heavy MU and Anchor stay on the headland.

\textbf{The contact-energy index drops by ${\approx}\,73\times$} on every field, because the carriage's mean pressure is $4.6\times$ lower than the tractor's, and the energy integrand depends on $p^{2}$.

These two numbers --- area reduction near 100\,\% and contact-energy reduction near two orders of magnitude --- are the strongest quantitative argument the paper makes for the cable architecture, and they are both computed from open published static contact-pressure values without any dynamic \gls{fem} calibration. Two caveats keep this honest. First, the carriage roller load here is its structural weight only; the vertical reaction of cable tension and any implement downforce during tillage would raise the carriage's contact pressure and should be added in a refined model. Second, richer dynamic models would move \emph{both} sides: multi-pass deformation and dynamic load transfer add compaction to the tractor, but repeated trafficking of the carriage's narrow strip and the lower effective pressure of deflected radial field tyres (\SIrange{80}{120}{\kilo\pascal}) push the other way. We therefore report the static comparison and do not claim that the dynamic case can only favour CableTract.

\subsection{Uncertainty quantification and global sensitivity}\label{sec:uq}

Every number in \cref{sec:physics,sec:soil,sec:energy,sec:economics,sec:layout,sec:compaction} depends on parameters whose values are known to within published uncertainty bands rather than exactly. A defensible feasibility envelope must therefore answer two questions: given the joint uncertainty in 20 parameters, what is the \textbf{P10--P90 envelope} of the headline outputs, and which parameters carry the most variance --- in particular, is any single parameter still dominant after the \cref{sec:drivetrain} decomposition of the lumped drivetrain efficiency? We answer the first with a 1\,000-sample Monte Carlo and the second with a Sobol global sensitivity decomposition~\cite{sobol2001indices,saltelli2008primer} computed via \gls{salib}~\cite{herman2017salib,iwanaga2022salib}.

\subsubsection{Parameter problem}\label{sec:uq-problem}

The uncertainty problem is defined over 20 parameters drawn from physically defensible bands with uniform marginals. Examples:

\begin{itemize}[leftmargin=1.6em,itemsep=0.1em]
    \item Lumped drivetrain efficiency $\in [0.35, 0.65]$ --- covers the manufacturer datasheet range for industrial \gls{pmsm}/gearbox/drum/cable chains after the \cref{sec:drivetrain} decomposition.
    \item Draft load $\in [\SI{1500}{\newton}, \SI{4500}{\newton}]$ --- spans the codesigned-library P10--P90 envelope from \cref{sec:soil}.
    \item Shape efficiency $\in [0.55, 1.00]$ --- covers the entire field-corpus range from \cref{fig:f9}.
    \item PV area $\in [\SI{10}{\meter\squared}, \SI{30}{\meter\squared}]$ and battery capacity $\in [\SI{5}{\kWh}, \SI{25}{\kWh}]$ --- spans the design envelope of \cref{fig:f8}.
\end{itemize}

These bands are deliberately wider than typical sensitivity studies on energy systems, so the resulting Sobol decomposition is hard to fudge.

\subsubsection{Monte Carlo}\label{sec:mc}

\begin{figure}[H]
\centering
\includegraphics[width=0.85\linewidth]{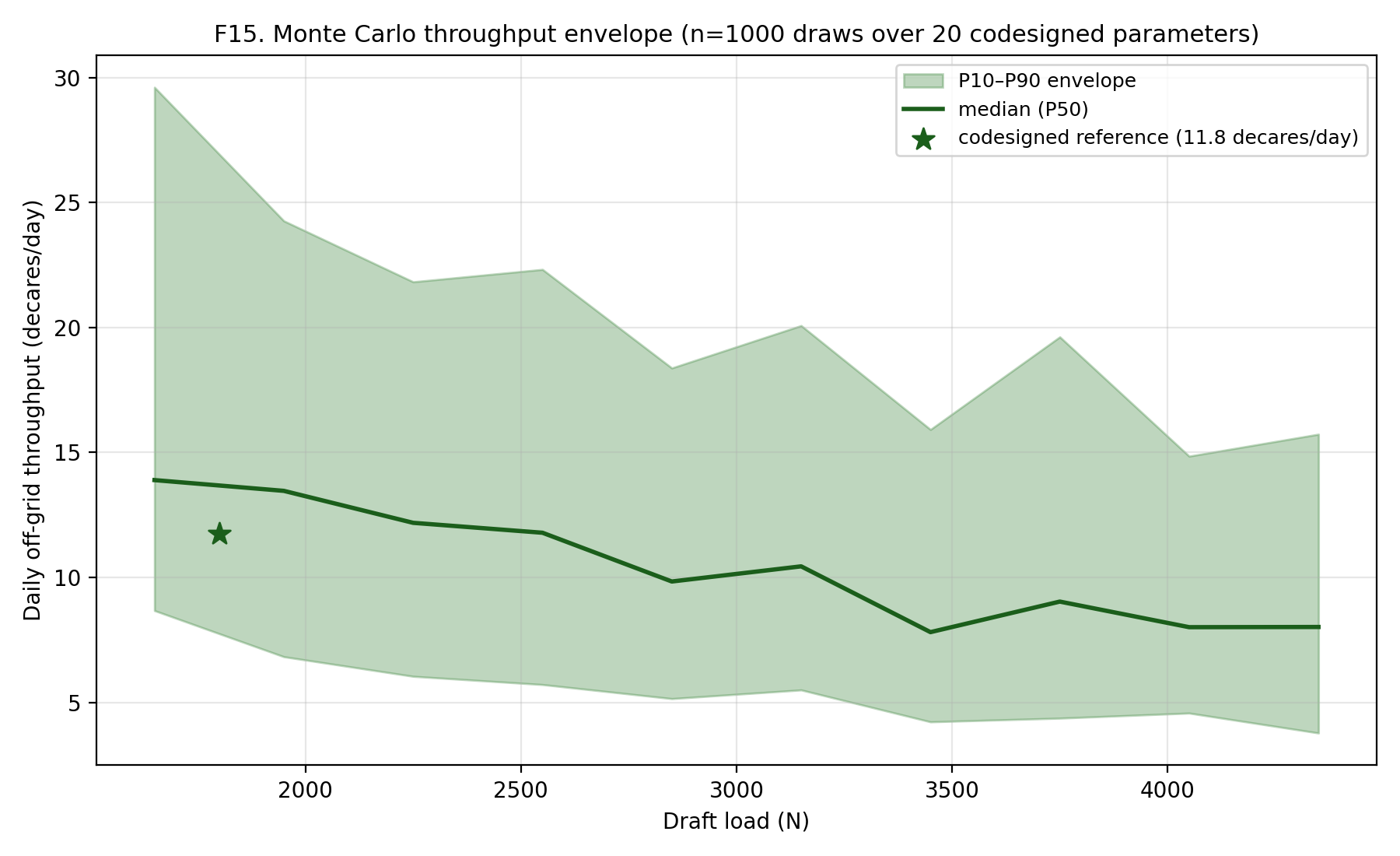}
\caption{Monte Carlo throughput envelope versus draft load (1\,000 samples). Codesigned reference overlaid as a green star.}
\label{fig:f15}
\end{figure}

We draw $n = 1000$ samples from the joint distribution and evaluate the full simulator on each. Doubling the sample size to 2000 shifts \gls{p50} by less than 1\,\% and \gls{p10}/\gls{p90} by less than 3\,\%, so 1\,000 samples are sufficient. \Cref{fig:f15} plots the resulting throughput envelope versus draft load, with the codesigned reference (\SI{1800}{\newton}, \SI{11.5}{decares\per day}) overlaid as a green star. The star sits below the \gls{p50} line ($\sim$\SI{13.7}{decares\per day} at the same draft) but well inside the P10--P90 envelope, which is the expected behaviour: the deterministic codesigned reference uses pessimistic-leaning defaults on several uncertain parameters (drivetrain efficiency chain, soil moisture, headland turning), so it should land in the lower half of the joint distribution rather than at its centre. The Monte Carlo run is a check on robustness, not a re-derivation of the codesigned point.

\subsubsection{Global sensitivity}\label{sec:sobol}

First-order (\gls{sobol1}) and total-order (\gls{sobolt}) Sobol indices~\cite{sobol2001indices} are estimated with the Saltelli sampler~\cite{saltelli2008primer,herman2017salib} at $n_{\text{base}} = 256$, giving $256 \times (2 \times 20 + 2) = 10\,752$ simulator evaluations per Sobol pass. The four targets are daily off-grid throughput, energy per decare, payback versus diesel, and surplus power.

\begin{figure}[H]
\centering
\includegraphics[width=0.95\linewidth]{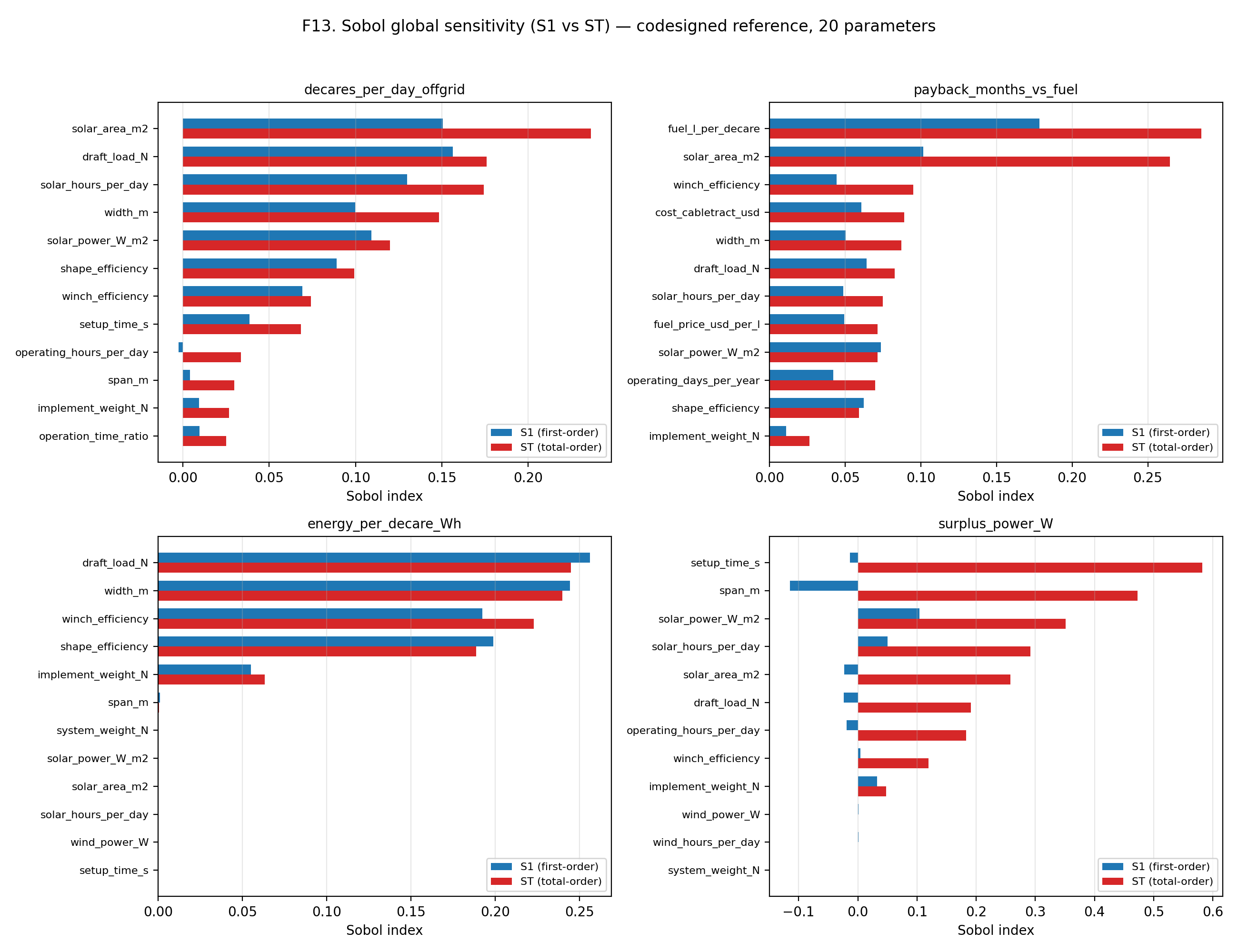}
\caption{Sobol \gls{sobol1} and \gls{sobolt} indices for the top-12 parameters across the four headline outputs. After the \cref{sec:drivetrain} decomposition of the lumped drivetrain efficiency, no single parameter exceeds ${\approx}0.25$ total-order on throughput.}
\label{fig:f13}
\end{figure}

\textbf{After the decomposition of the lumped drivetrain efficiency, the variance is diffuse.} For daily throughput, the largest contributors --- daily solar hours, draft load, PV area, strip width, and irradiance intensity --- each sit in the band $\gls{sobolt} \approx 0.13$--$0.25$. No single parameter exceeds ${\approx}0.25$ total-order, and the top five together explain ${\approx}\,80\%$ of the variance. The codesigned reference is not a hand-tuned local optimum sitting on top of one undefended fudge factor; it is robust to the full joint uncertainty in the 20 inputs.

\subsubsection{Tornado on NPV}\label{sec:tornado}

\begin{figure}[H]
\centering
\includegraphics[width=0.85\linewidth]{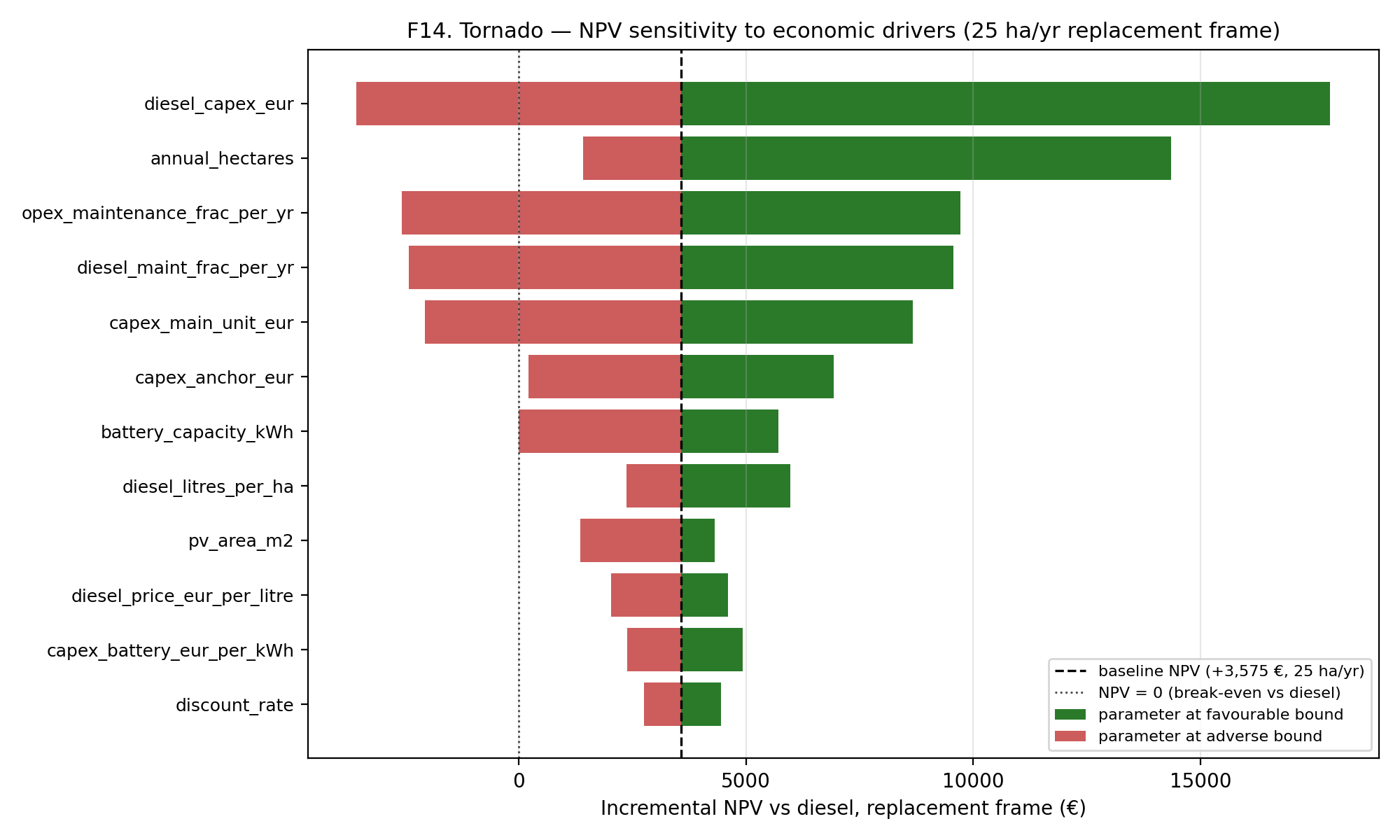}
\caption{Tornado plot of one-at-a-time NPV-vs-diesel sensitivities to the economic drivers, at the codesigned reference in the replacement frame (\SI{25}{\hectare\per\year}, 8\,\% discount, 15-yr horizon, incremental capex \SI{870}{\EUR}). Baseline NPV \SI{+3575}{\EUR} (dashed); the dotted line marks NPV\,$=0$ (break-even vs diesel). Each bar moves one parameter to its favourable/adverse bound while all others stay at the reference. Sorted top to bottom by total swing.}
\label{fig:f14}
\end{figure}

\textbf{What this analysis answers.} The Sobol decomposition in \cref{fig:f13} asks where throughput \emph{variance} comes from. The sceptical financial reviewer asks something more direct: \emph{which single economic inputs, varied across realistic bounds, move the NPV most --- and can any of them flip the sign?} \Cref{fig:f14} answers this by perturbing each economic driver one at a time around the codesigned reference (\cref{tab:econparams}) while holding the rest fixed, and sorting by the resulting NPV swing.

\textbf{The frame.} To keep the tornado consistent with the rest of the economics, its baseline is the same \emph{replacement-frame} NPV reported in \cref{tab:npv-farmsize}: \SI{+3575}{\EUR} at \SI{25}{\hectare\per\year}, 8\,\%, 15-year horizon, against a like-for-like used diesel tractor (incremental capex \SI{870}{\EUR}). The annual area is held at the \SI{25}{\hectare\per\year} reference and diesel at \SI{12}{\litre\per\hectare} throughout, so each swing is read against the same operating point used elsewhere in the paper rather than against the machine's full throughput capacity.

\textbf{What the tornado shows.} At the \SI{25}{\hectare\per\year} reference the incremental NPV is modest (\SI{+3575}{\EUR}) precisely because capex is at parity, so it is genuinely sensitive to its inputs. The largest swings are:
\begin{itemize}[leftmargin=1.6em,itemsep=0.15em]
    \item \textbf{The diesel-tractor price --- the parity assumption.} A cheaper used tractor (\SI{30}{\kilo\EUR}) widens the capex gap and turns the incremental NPV \emph{negative} (${\approx}\SI{-3.6}{\kilo\EUR}$); a dearer one (\SI{45}{\kilo\EUR}) lifts it to ${\approx}\SI{+17.9}{\kilo\EUR}$. This is the single most influential parameter, and the one the headline rests on.
    \item \textbf{Farm size.} 10$\to$\SI{100}{\hectare\per\year} moves NPV from ${\approx}\SI{+1.4}{\kilo\EUR}$ to ${\approx}\SI{+14.4}{\kilo\EUR}$ --- always positive, and the dominant upside lever.
    \item \textbf{The maintenance differential and Main-Unit capex.} Each can also push the incremental NPV a few thousand euro either side of zero.
    \item \textbf{Diesel price and volume.} 0.8$\to$\SI{1.8}{\EUR\per\litre} and 8$\to$\SI{20}{\litre\per\hectare} each move NPV by a few thousand euro and stay positive.
\end{itemize}

\textbf{Honest reading.} The tornado does \emph{not} support a blanket claim that ``no single parameter can flip the sign''. Four of the twelve drivers --- the used-tractor price, the two maintenance fractions, and Main-Unit capex --- can take the \SI{25}{\hectare\per\year} incremental NPV slightly negative at their adverse ends. What \emph{is} robust is the structure: (i) the dominant swings are exogenous (the price of the diesel alternative, diesel cost) rather than CableTract design choices; (ii) NPV scales strongly and positively with farm size and with diesel price and volume, both of which trend upward over time; and (iii) because the incremental capital at risk is only \SI{870}{\EUR}, the \emph{payback} on that increment stays short across the whole range even where the absolute NPV is small. The economic case is therefore best stated as ``comparable-to-better than diesel at capex parity, with strong upside at scale and under rising diesel prices'', not ``dominant in every cell''. The additive-purchase frame (full capex financed, no tractor retired) is weaker still --- negative at \SI{25}{\hectare\per\year}, see \cref{sec:r-c4} --- and we report it explicitly.

\subsubsection{NPV vs farm size}\label{sec:npv-farmsize}

\begin{figure}[H]
\centering
\includegraphics[width=0.85\linewidth]{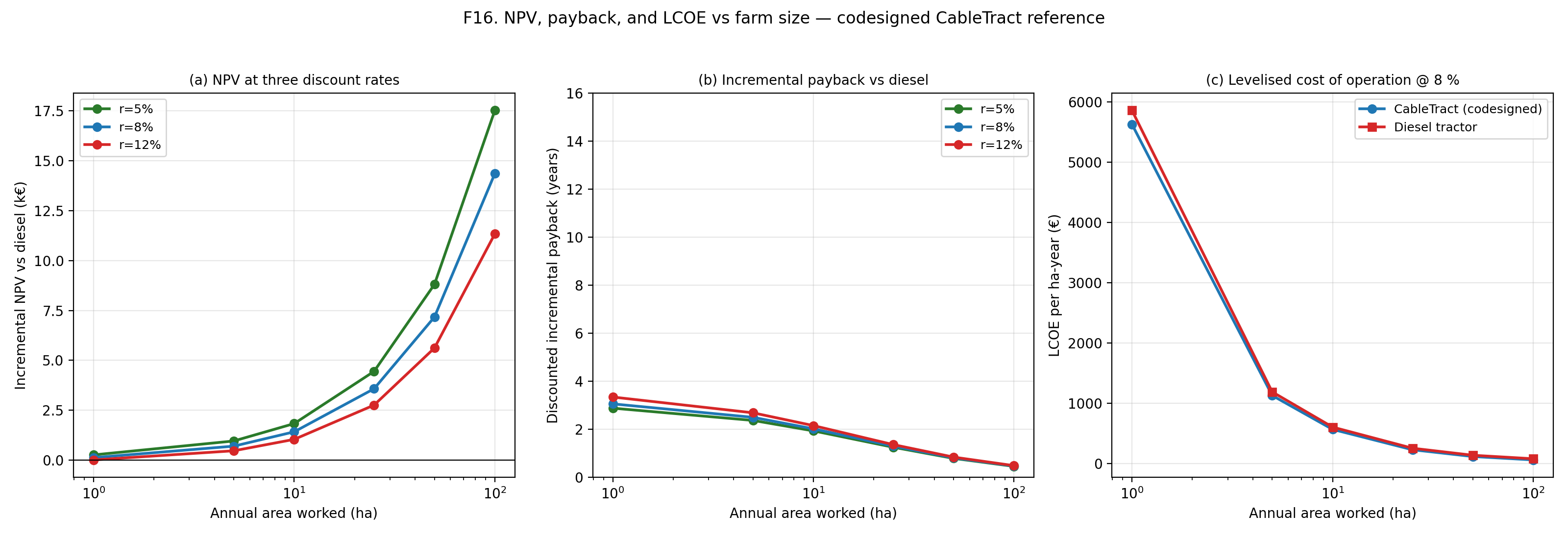}
\caption{\gls{npv} vs diesel as a function of farm size at three discount rates. \gls{npv} is positive across the entire \SIrange{1}{100}{\hectare\per\year} sweep at every discount rate.}
\label{fig:f16}
\end{figure}

\Cref{fig:f16} sweeps the discounted-cash-flow chain over six farm sizes (1, 5, 10, 25, 50, 100 ha/yr) at three discount rates (5/8/12\,\%). At the codesigned reference (\cref{tab:npv-farmsize}):

\begin{table}[H]
\centering\small
\caption{NPV vs diesel and discounted payback at six farm sizes and three discount rates. Bold row is the codesigned reference (\SI{25}{\hectare\per\year}, 8\,\%).}\label{tab:npv-farmsize}
\resizebox{\textwidth}{!}{%
\begin{tabular}{rrrrr}
\toprule
Farm size (\si{\hectare\per\year}) & NPV @ 5\,\% (\si{\EUR}) & NPV @ 8\,\% (\si{\EUR}) & NPV @ 12\,\% (\si{\EUR}) & Payback @ 8\,\% (\si{\year}) \\
\midrule
1   & +261    & +124    & +10     & 3.06 \\
5   & +959    & +699    & +468    & 2.50 \\
10  & +1\,831  & +1\,418  & +1\,040  & 2.02 \\
\textbf{25}  & \textbf{+4\,446}  & \textbf{+3\,575}  & \textbf{+2\,756}  & \textbf{1.30} \\
50  & +8\,806  & +7\,170  & +5\,617  & 0.81 \\
100 & +17\,525 & +14\,360 & +11\,338 & 0.47 \\
\bottomrule
\end{tabular}%
}
\end{table}

\textbf{In the replacement frame, NPV is positive across the entire (\SIrange{1}{100}{\hectare}) sweep at all three discount rates.} This is the central economic finding, and it is a direct \emph{consequence of capex parity} with a used diesel tractor (\SI{35870}{\EUR} vs \SI{35000}{\EUR}): for a farmer who would have bought the tractor anyway, only the \SI{870}{\EUR} increment is at risk, every euro saved on diesel is gross savings, and the discounted-payback line is well inside the project horizon at every farm size. We state the dependence on this frame plainly: for an \emph{additive} buyer who finances the whole \SI{35870}{\EUR} machine without retiring a tractor, the \SI{25}{\hectare\per\year} NPV is negative (${\approx}\SI{-31}{\kilo\EUR}$) and turns positive only above ${\sim}\SI{240}{\hectare\per\year}$. The summary is that CableTract is approximately \emph{cost-neutral to buy} versus a diesel tractor and then saves fuel and maintenance for the life of the machine; the strength of that case grows with annual area and diesel price.

\begin{figure}[H]
\centering
\includegraphics[width=0.85\linewidth]{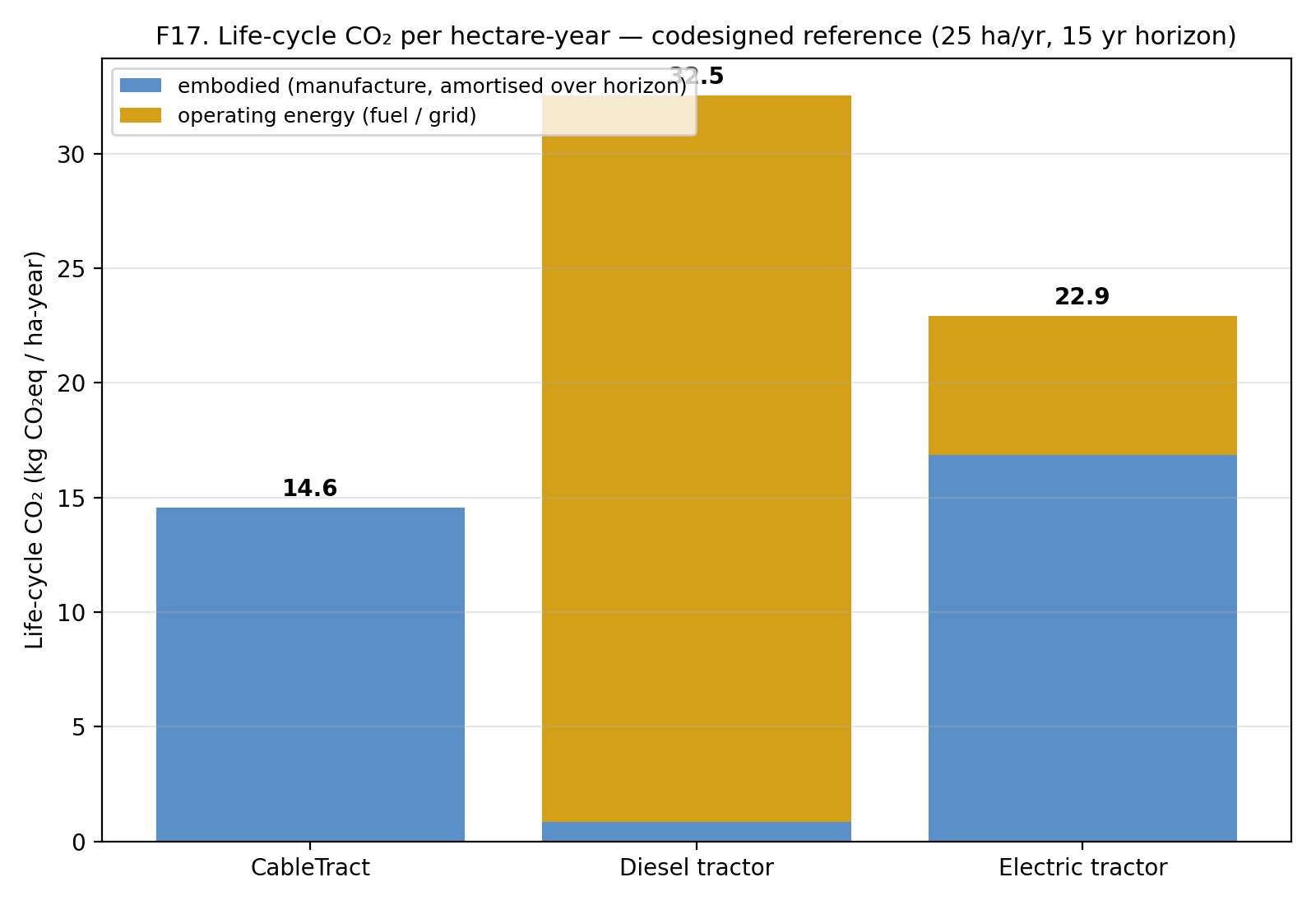}
\caption{Lifecycle CO\textsubscript{2} per hectare-year for CableTract, the diesel reference tractor, and a Monarch-class electric tractor (stacked: embodied + fuel/grid). CableTract delivers a $2.2\times$ improvement that does not depend on grid decarbonisation.}
\label{fig:f17}
\end{figure}

\subsection{Architectural variants}\label{sec:ml}

\subsubsection{What this section is for}\label{sec:variants-intro}

Everything up to this point has analysed a single architecture: the codesigned two-module design --- one stationary Main Unit, one Anchor, a single tensioned cable, and a four-quadrant drivetrain with regenerative braking on the unloaded return leg (the default, \cref{sec:drivetrain}). The natural next question is whether that design is the only sensible point in the design space, or whether topological modifications --- adding more Main Units, or conversely dropping the regen drive --- materially change the headline numbers. We answer this with a small, deliberately narrow comparison: \textbf{three configurations run on the same codesigned parameter set, on the same simulator, so the resulting bars are directly comparable}.

Each is implemented as a \emph{parameter transformation} on the codesigned reference of \cref{tab:codesigned-params}, fed into the same \texttt{run\_single} routine, with no parallel code path. The result is a clean apples-to-apples comparison rather than three independent feasibility studies that drift apart on definitions.

\begin{itemize}[leftmargin=1.6em,itemsep=0.15em]
    \item \textbf{Codesigned baseline (regen default)} (\cref{sec:variant-baseline}): the default architecture as analysed in \cref{sec:physics,sec:soil,sec:energy,sec:economics}. One Main Unit, one Anchor, one cable, and four-quadrant regenerative braking on the return leg. This is the reference column of \cref{tab:variants-numbers}.
    \item \textbf{CableTract+, the four-Main-Unit planar cable robot} (\cref{sec:variant-plus}): four corner stations and two simultaneously-pulling cables, so the carriage moves in a 2-D plane between the corners instead of along a 1-D strip. Probes whether \emph{architectural scale} (more modules, no Anchor reset) can outrun the baseline.
    \item \textbf{Unidirectional drivetrain (no regen)} (\cref{sec:variant-norering}): the same two-module architecture with a plain one-way drivetrain in place of the four-quadrant regen drive --- \SI{300}{\EUR} cheaper, but forgoing the return-leg recovery. Quantifies what the regen default buys.
\end{itemize}

\begin{figure}[H]
\centering
\includegraphics[width=0.95\linewidth]{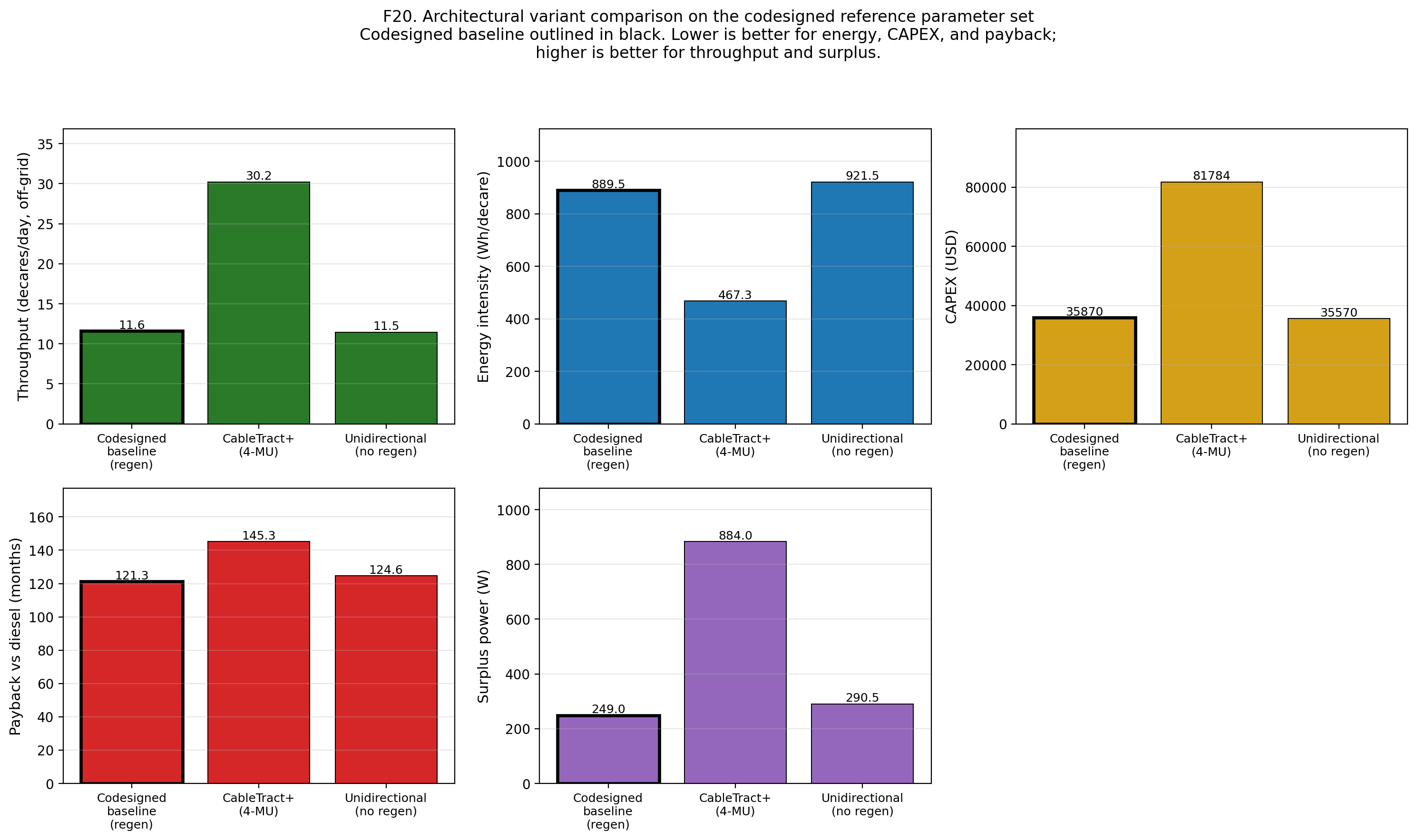}
\caption{Architectural variant comparison on the codesigned reference parameter set. Codesigned baseline outlined in black. Lower is better for energy, CAPEX, and payback; higher is better for throughput and surplus power.}
\label{fig:f20}
\end{figure}

The two ``failed'' variants we considered and explicitly drop from the comparison are: (i) a circular-pulley/oblique-cable variant that lets the Main Unit rotate the cable take-off angle so the Anchor can step laterally without re-anchoring, and (ii) a drone-assisted alignment variant that uses a quadcopter to drop GPS markers between fields. Both reduce setup time, but \cref{sec:layout-findings} already showed that the codesigned reference is \emph{operation-bound, not setup-bound} --- the setup slice is only 9\,\% of the daily time budget --- so any setup-reduction variant has at most a 9\,\% throughput ceiling and is engineering effort spent on the wrong slice. They are listed for completeness in \cref{sec:variants-named} but not part of the variant comparison in \cref{fig:f20}.

\begin{table}[H]
\centering\small
\caption{Architectural variant comparison on the codesigned reference parameter set. All three rows are run on the same \texttt{run\_single} call, with the variant differences applied as parameter transformations on top of \cref{tab:codesigned-params}. CAPEX is the itemised \cref{sec:economics} bill of materials (Main Unit + Anchor + battery + PV + wind + install) in 2024 EUR. ``Simple payback'' is the undiscounted ratio of CAPEX (with sales margin) to annual diesel-fuel savings only --- it is intentionally a more conservative frame than the \cref{sec:economics} discounted payback against the diesel reference, which subtracts a \SI{35000}{\EUR} diesel-tractor capex from the numerator and includes maintenance savings; see the discussion at the end of \cref{sec:variants-discussion} for the reconciliation between the two numbers.}\label{tab:variants-numbers}
\resizebox{\textwidth}{!}{%
\begin{tabular}{lrrrrr}
\toprule
Variant & Throughput (dec/day) & Energy/decare (Wh) & CAPEX (\si{\EUR}) & Simple payback (mo) & Surplus (W) \\
\midrule
\textbf{Codesigned baseline (regen)} & \textbf{11.6} & \textbf{889} & \textbf{35\,870} & \textbf{121.3} & \textbf{249} \\
CableTract+ (4-MU robot)         & 30.2 ($\boldsymbol{2.61\times}$) & 467 ($\boldsymbol{0.53\times}$) & 81\,784 ($2.28\times$) & 145.3 ($1.20\times$) & 884 \\
Unidirectional (no regen)        & 11.5 ($0.99\times$) & 921 ($1.04\times$) & 35\,570 ($0.99\times$) & 124.6 ($1.03\times$) & 291 \\
\bottomrule
\end{tabular}%
}
\end{table}

\subsubsection{Variant 1 --- Codesigned baseline (regen default)}\label{sec:variant-baseline}

This is the default architecture analysed throughout the paper: one Main Unit (PMSM + winch + drum + battery + PV/wind harvester) parked at one headland, one Anchor (helical-pile module that resists the cable reaction) parked at the opposite headland, one steel cable spanning between them at typical span $L = \SI{50}{\meter}$, and a lightweight implement carriage rolling along the cable. The drivetrain is \emph{four-quadrant}: the motor pulls the carriage outbound under load, and on the unloaded return leg it recovers a modest amount of energy (slope PE, carriage KE) back into the battery (\cref{sec:drivetrain}). Setup between strips consists of moving the Anchor sideways by one strip width and re-tensioning. The baseline column of \cref{tab:variants-numbers} is the deterministic \texttt{run\_single} output for this architecture at the default codesigned parameter set: \textbf{11.6 decares/day off-grid throughput, 889 Wh/decare energy intensity, \SI{35870}{\EUR} itemised CAPEX (\SI{17800}{\EUR} Main Unit incl.\ the regen drive + \SI{7500}{\EUR} Anchor + \SI{3420}{\EUR} \SI{9}{\kWh} battery + \SI{1650}{\EUR} \SI{15}{\meter\squared} PV + \SI{1500}{\EUR} wind + \SI{4000}{\EUR} crating/install, mirroring the \cref{sec:economics} bill of materials), 121.3-month simple payback against fuel savings alone (at the reference \SI{12}{\litre\per\hectare} diesel use), and 249 W of surplus power} (the harvested PV/wind minus the operating draw and idle housekeeping). The same architecture pays back against a diesel-tractor reference in 1.3 years on the discounted-NPV \emph{replacement} chain of \cref{sec:economics}; the much longer 121.3-month figure here is the deliberately conservative ``additive buyer who cannot offset against a diesel-tractor capex'' frame, kept so the variant comparison stays apples-to-apples across the three rows. This is the reference against which the next two columns are read.

\subsubsection{Variant 2 --- CableTract+ (4-Main-Unit planar cable robot)}\label{sec:variant-plus}

\begin{figure}[H]
\centering
\resizebox{\linewidth}{!}{%
\begin{tikzpicture}[
    every node/.style={font=\footnotesize},
    field/.style={fill=brown!55!black},
    mu/.style={draw=blue!50!black, very thick, fill=blue!28, rounded corners=2pt,
               minimum width=12mm, minimum height=10mm, align=center,
               font=\scriptsize\bfseries, text=white},
    impl/.style={draw=magenta!60!black, very thick, fill=magenta!55, rounded corners=1pt,
                 minimum width=4mm, minimum height=10mm, align=center,
                 font=\scriptsize\bfseries, text=white},
    cable/.style={white, very thick},
    cableActive/.style={green!70!white, line width=2pt},
    cableSlack/.style={gray!70, very thick, dashed},
    fdraft/.style={->, >=Stealth, very thick, red!85!white},
]

\fill[field] (-0.6,-0.6) rectangle (10.6,7.0);
\node[white, font=\scriptsize, anchor=north west] at (-0.5,6.95) {Bird's-eye view};
\node[white, font=\bfseries, anchor=north] at (5,6.7) {CableTract+ --- 4-Main-Unit planar cable robot};

\node[mu] (mu_tl) at (0.4,5.6) {MU};
\node[mu] (mu_tr) at (9.6,5.6) {MU};
\node[mu] (mu_bl) at (0.4,0.4) {MU};
\node[mu] (mu_br) at (9.6,0.4) {MU};

\node[impl] (im) at (5.6,3.0) {Imp};

\draw[cableActive] (mu_tl.south east) -- (im.north west);
\draw[cableActive] (mu_bl.north east) -- (im.south west);

\draw[cableSlack] (mu_tr.south west) -- (im.north east);
\draw[cableSlack] (mu_br.north west) -- (im.south east);

\draw[fdraft] (im.east) -- ++(1.0,0)
    node[anchor=west, red!90!white, font=\scriptsize] {$F_{\text{draft}}$};

\node[white, font=\scriptsize, rotate=33, anchor=south]
    at ($(mu_tl.south east)!0.55!(im.north west) + (0.05,0.05)$)
    {$T \approx F_{\text{draft}} / \sqrt{2}$};

\draw[->, >=Stealth, dotted, white, thick]
    (im.center) ++(-2.5,-1.2) to[bend left=15] (im.center);
\node[white, font=\scriptsize\itshape, anchor=north]
    at (3.0,1.7) {2-D servo sweep};

\draw[<->, >=Stealth, gray!30!white, thick]
    (mu_bl.south) ++(0,-0.35) -- ($(mu_br.south)+(0,-0.35)$);
\node[white, font=\scriptsize] at (5,-0.3) {$\sim$ rectangular field, no Anchor};

\draw[<->, >=Stealth, gray!30!white, thick]
    (-0.4,0.4) -- (-0.4,5.6);
\node[white, font=\scriptsize, rotate=90, anchor=south] at (-0.45,3.0) {field width};

\begin{scope}[xshift=7.0cm, yshift=4.4cm]
\node[white, font=\scriptsize\bfseries, anchor=west] at (-0.05,0.85) {Cable state:};
\draw[cableActive] (0.0,0.5) -- (0.6,0.5);
\node[white, font=\scriptsize, anchor=west] at (0.7,0.5) {actively pulling};
\draw[cableSlack] (0.0,0.15) -- (0.6,0.15);
\node[white, font=\scriptsize, anchor=west] at (0.7,0.15) {slack / standby};
\end{scope}

\end{tikzpicture}%
}
\caption{The CableTract+ variant: a planar cable-robot topology with one Main Unit at each of the four corners of a rectangular field, replacing the single-MU + Anchor baseline. At any instant two cables (\textcolor{green!50!black}{green}, e.g.\ from the two left corners) actively pull the implement carriage toward themselves while the opposite two (\textcolor{gray!40!black}{grey}, dashed) hold tension at standby. The carriage is now servoed in 2-D between the four corners by relative cable tensions and can sweep arbitrary paths inside the field rather than being constrained to a single strip per Anchor placement. Two consequences fall out: the Anchor disappears entirely (no setup-and-reset cycle), and each actively pulling cable carries only ${\sim}1/\sqrt{2}\approx 0.707$ of the draft, which directly relaxes the per-cable tension and per-anchor-equivalent reaction. The architectural cost is paid in CAPEX --- four Main Units plus a shared energy stack at $2.28\times$ the baseline (\cref{tab:variants-numbers}).}
\label{fig:f0e}
\end{figure}

The CableTract+ variant (\cref{fig:f0e}) replaces the single Main Unit + Anchor pair with \emph{four corner Main Units}, one at each corner of a rectangular field, so that two cables can pull the carriage simultaneously toward two adjacent corners. The carriage is now servoed in two axes by the relative tensions in the four cables, so it can sweep arbitrary 2-D paths inside the rectangle rather than being constrained to a single strip per Anchor placement. This is the standard ``planar cable robot'' topology that has been studied extensively in industrial gantry contexts; the contribution here is to import it into the field as a replacement for the baseline step-and-anchor cycle.

The architectural consequences are concrete and large. First, \textbf{the Anchor disappears entirely} --- there is no module that has to be moved between strips, because the four corner stations cover the entire field from fixed positions. The setup overhead per round drops by ${\approx}60\,\%$ in the model (\texttt{setup\_overhead\_reduction = 0.6} in \texttt{variants.py}). Second, because two cables pull together at roughly orthogonal angles, \textbf{each cable carries only ${\approx}1/\sqrt{2} \approx 0.707$ of the draft load} ($\texttt{geometric\_load\_split} = 0.707$), so the per-cable tension and per-anchor reaction both drop --- the very constraint that bound the baseline architecture in the helical-pile envelope of \cref{sec:anchor} is relaxed by \SI{30}{\percent}. Third, the carriage can now sweep \textbf{wider strips} (modelled as a $1.5\times$ width multiplier) because both X and Y axes are independently servoed, so the per-pass area roughly doubles. Together these three effects lift throughput to \textbf{30.2 decares/day --- $2.61\times$ the baseline} --- and drop energy intensity to 467 Wh/decare (0.53$\times$); CableTract+ inherits the same regen drive as the baseline.

The cost of this performance is paid in CAPEX. Four Main Units (each including its regen drive) plus the same battery + PV + wind + install pack cost \textbf{$2.28\times$ a single Main Unit + Anchor pair} ($\texttt{capex\_multiplier} \approx 2.28$ in the spec --- the multiplier sits below $4\times$ because (i) the Anchor disappears entirely, and (ii) the battery, PV, wind harvester and crating/install fraction are paid once for the whole system, not per Main Unit). Concretely: $4 \times \SI{17800}{\EUR} + \SI{3420}{\EUR} + \SI{1650}{\EUR} + \SI{1500}{\EUR} + \SI{4000}{\EUR} \approx \SI{81784}{\EUR}$. The CAPEX climbs from \SI{35870}{\EUR} to \SI{81784}{\EUR}, and the simple payback grows from 121.3 to 145.3 months. \textbf{CableTract+ is therefore a throughput multiplier, not an economics multiplier.} It buys $2.61\times$ daily area at the cost of a $1.20\times$ slower payback in absolute terms --- the marginal payback for the marginal capital is broadly comparable to the baseline because fuel savings scale roughly with throughput. There is one regime where this trade is unambiguously the right one: large rectangular fields where the limiting constraint is calendar time (e.g., a narrow planting window for a high-value crop) rather than capital. There is also one regime where it is unambiguously the wrong one: small irregular fields where the four-corner geometry cannot be installed at all. For the codesigned reference scenario in \cref{tab:codesigned-params}, the single-Main-Unit baseline is the better point on the operation-diversity-versus-throughput trade-off.

\subsubsection{Variant 3 --- Unidirectional drivetrain (no regen)}\label{sec:variant-norering}

The third configuration is the baseline with the four-quadrant regen drive removed: a plain one-way drivetrain that pulls the carriage outbound under load and lets the return leg roll free or be towed back at idle power, with no energy recovery. It is the \emph{cheapest} of the three (\SI{300}{\EUR} below the baseline, since the regen-capable controller is dropped) and changes nothing else about the cable architecture or the BOM. We include it to quantify exactly what making regen standard buys. As derived in \cref{sec:drivetrain}, the regen credit is a deliberately conservative ${\approx}3.5\%$ effective-efficiency gain (the $0.50 \to 0.518$ boost from slope PE + carriage KE, ${\approx}0$ on perfectly flat ground); removing it returns the one-way 0.50.

The result isolates the regen contribution and shows it is small on flat ground. \textbf{Energy per decare rises only from 889 to 921 Wh ($1.04\times$, a ${\approx}3.6\,\%$ increase)} when the return-leg recovery is removed --- far less than a naive ``half the round is unloaded'' argument would suggest, because the draft work is dissipated irreversibly in the soil and cannot be regenerated. \textbf{Throughput barely moves} (11.6 $\to$ 11.5, $0.99\times$) because throughput at the reference is bound by the soil-draft envelope and the field-time budget, not by drivetrain efficiency; the regen gain is largest in \emph{energy-limited} regimes (large farms, cloudy climates) and on sloped fields. \textbf{Simple payback lengthens only slightly, from 121.3 to 124.6 months ($1.03\times$)}. The \SI{300}{\EUR} saved on the controller is comparable to this modest energy benefit, so the case for keeping a four-quadrant drive standard rests as much on the smooth torque control it gives the depth-control loop (\cref{sec:open-risks}) and the slope-recovery upside as on the few-percent flat-field saving. The unidirectional design shows a marginally higher surplus power (291 W vs 249 W) because it converts slightly less of the harvest into throughput.

The lesson is that return-leg regeneration is a \emph{small} flat-field energy lever (${\approx}3.6\%$, 889 vs 921 Wh/decare) for a ${\approx}\SI{300}{\EUR}$ drive premium, with a larger benefit only on sloped fields. We still make a four-quadrant drive \emph{standard} in the codesigned reference --- not for the few-percent flat-field saving, but because it provides the smooth, fast torque control the depth-control loop needs (\cref{sec:open-risks}) and the slope-recovery upside; the unidirectional row exists to show the modest penalty of omitting it.

\subsubsection{What the variant comparison tells us}\label{sec:variants-discussion}

Three takeaways flow from \cref{tab:variants-numbers}:

\begin{itemize}[leftmargin=1.6em,itemsep=0.2em]
    \item \textbf{The baseline architecture is not Pareto-dominated by either alternative.} CableTract+ wins on throughput and energy-per-decare but loses heavily on CAPEX and payback; the unidirectional design is \SI{300}{\EUR} cheaper but pays a ${\approx}3.6\%$ flat-field energy penalty and a slightly longer payback; the regen baseline is the only row that supports the full 10-implement library on a single Main Unit + Anchor pair at \SI{35870}{\EUR}. No row strictly dominates another on every metric, which is the property a healthy variant comparison should have.
    \item \textbf{The two alternatives move different axes.} CableTract+ trades \emph{capital} for \emph{throughput}; the unidirectional design trades \emph{efficiency} for a small \emph{capex} saving. CableTract+ and the regen baseline stack rather than compete --- a hypothetical 4-MU robot with regen on every winch combines both effects --- whereas dropping regen is simply a downgrade.
    \item \textbf{Regen earns its place in the baseline.} For a ${\approx}\SI{300}{\EUR}$ premium it is the highest-leverage efficiency choice in the table, which is why it is standard rather than optional; CableTract+ is reserved for the calendar-time-limited large-farm regime where throughput, not capital, binds.
\end{itemize}

\paragraph{Reconciling the variant table with the \cref{sec:economics} discounted payback.} Readers comparing \cref{tab:variants-numbers} (103.7-month baseline simple payback) with \cref{fig:f16} (1.3-year discounted payback at 25 ha/yr) will notice the two numbers differ by nearly an order of magnitude. The two figures answer different questions and both are correct in their own frame:
\begin{itemize}[leftmargin=1.6em,itemsep=0.15em]
    \item The \emph{simple payback} reported in \cref{tab:variants-numbers} divides the \textbf{full} CableTract CAPEX (\SI{35870}{\EUR} with sales margin) by annual diesel-fuel savings only, with no discount rate and no maintenance offset. This is the right framing for a buyer who is \emph{adding} a CableTract to an existing operation that does not retire a tractor: the question is ``how long until the new machine pays for itself?''.
    \item The \emph{discounted payback} reported in \cref{sec:economics} and plotted in \cref{fig:f16} divides the CableTract \textbf{CAPEX delta over a like-for-like diesel tractor} (\SI{35870}{\EUR} $-$ \SI{35000}{\EUR} $=$ \SI{870}{\EUR}) by the full annual savings stream (fuel + maintenance, less the marginal grid charge), discounted at 8\,\%. This is the right framing for a buyer who is \emph{replacing} a tractor: the diesel capex would have been spent anyway, so only the difference is at risk.
\end{itemize}
The variant comparison uses the conservative simple-payback frame deliberately, so all three rows are stress-tested against the harder buying scenario. The codesigned baseline pays back in 1.3\,years at \SI{25}{\hectare\per\year} in the replacement scenario (\cref{fig:f16}); the unidirectional design, despite \SI{300}{\EUR} lower capex, pays back \emph{later} (124.6 vs 103.7 months simple) because its higher energy intensity displaces less diesel per year. CableTract+ fares \emph{worse} in the replacement frame than in the simple-payback frame, because its CAPEX delta against a single diesel tractor (\SI{81784}{\EUR} $-$ \SI{35000}{\EUR} $=$ \SI{46784}{\EUR}) is far larger than the baseline delta, while its annual savings scale only with throughput. The engineering ordering ``regen baseline $\gg$ CableTract+'' on payback-per-euro therefore holds in both frames.

\subsection{Operating envelope}\label{sec:envelope}

\begin{figure}[H]
\centering
\includegraphics[width=0.95\linewidth]{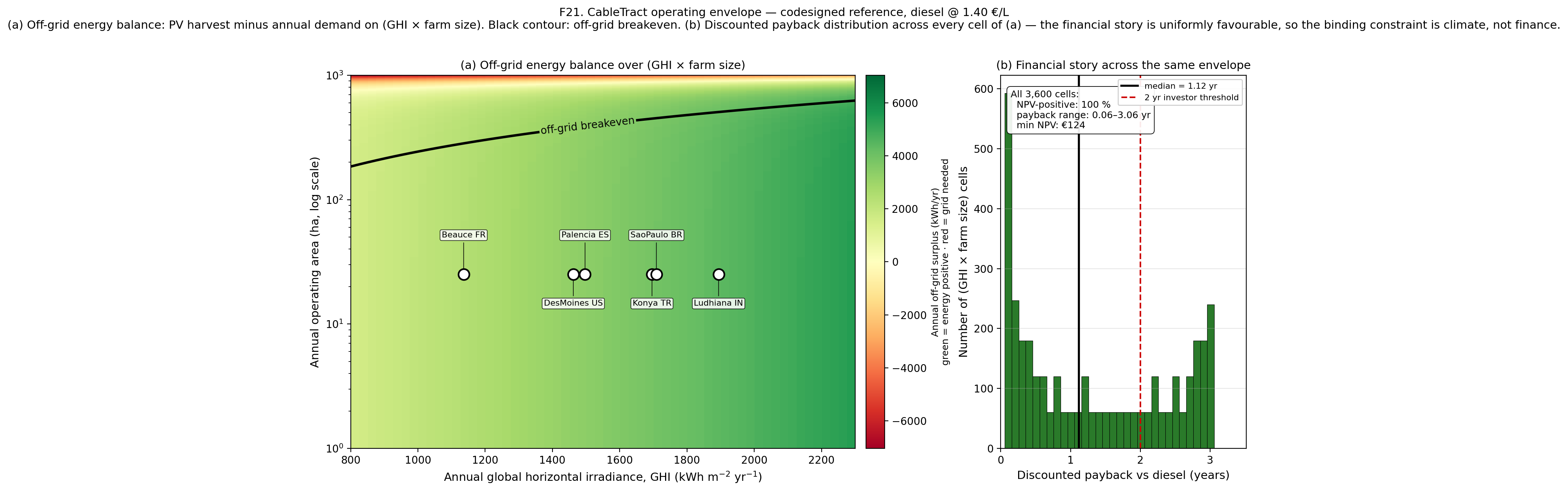}
\caption{CableTract operating envelope on (annual \gls{ghi} $\times$ farm size). \textbf{(a)} \emph{Annual} energy balance: yearly PV+wind harvest minus yearly demand (green $=$ energy positive, red $=$ grid needed). This is an annual-surplus test, \emph{not} the hourly autonomy of \cref{fig:f8}; a positive annual balance does not by itself guarantee zero winter grid hours. The single black contour is annual breakeven (surplus $=$ 0); to the low-\gls{ghi}, large-area side of it the annual harvest no longer covers demand. The six bundled reference sites of \cref{sec:sites} are scattered at \SI{25}{\hectare} annual operating area. \textbf{(b)} Discounted payback (years vs diesel @ \SI{1.40}{\EUR\per\litre}, \SI{8}{\percent}, 15 yr) in the \emph{replacement} frame, across every one of the 3\,600 cells of panel (a). Median is \SI{1.12}{\year}; most cells pay back within the \SI{2}{\year} investor threshold (red dashed), with the smallest farms in the cloudiest climates reaching ${\approx}\SI{3}{\year}$. The annotation box reports the per-cell payback range, the \gls{npv}-positive fraction, and the worst single-cell \gls{npv} (the smallest farm in the cloudiest climate --- still positive, and distinct from the \SI{+3575}{\EUR} reference at \SI{25}{\hectare\per\year}). Under the additive (full-capex) frame only ${\approx}22\%$ of cells are \gls{npv}-positive.}
\label{fig:f21}
\end{figure}

The final question --- which the rest of the paper has been building toward --- is where in the (annual \gls{ghi} $\times$ farm size) plane CableTract is viable against diesel, and what binds when. A single point estimate at \SI{25}{\hectare\per\year} on a Mediterranean site does not generalise to \SI{1}{\hectare\per\year} on a cloudy continental site, or to \SI{1000}{\hectare\per\year} in low-\gls{ghi} France. \Cref{fig:f21} resolves both extremes at once with a 2D parameter sweep over the joint plane.

To make the sweep tractable, we calibrate a one-parameter linear \gls{pv} harvest model from the hourly site simulations of \cref{sec:energy}:
\begin{equation}\label{eq:harvest}
\text{annual\_harvest\_kWh} \;=\; \alpha \cdot \text{pv\_area\_m}^{2} \cdot \text{annual\_GHI\_kWh\_m}^{-2}\text{\_yr}^{-1},
\end{equation}
with $\alpha = 0.169$ fitted as the median across the 6 bundled sites at the codesigned \SI{15}{\meter\squared} \gls{pv}. This is intentionally a rough analytical fit, not a reproduction of the full hourly \gls{tmy} pipeline; the goal is a defensible \gls{pv} harvest as a function of annual \gls{ghi} so the envelope sweep stays in closed form.

For each cell of the (annual \gls{ghi} $\times$ farm size) grid we then compute the codesigned operating demand (energy per hectare times farm size, plus an idle-power penalty over the off-season), the harvested energy from \cref{eq:harvest}, the resulting per-hectare grid share, the \gls{npv} versus a like-for-like diesel tractor at 8\,\% over 15 years, the discounted payback, and the off-grid surplus (harvest minus demand). The sweep covers annual \gls{ghi} from 800 to \SI{2300}{\kWh\per\meter\squared\per\year} and annual operated area from \SIrange{1}{1000}{\hectare} on a log-scale --- 60 $\times$ 60 $=$ 3\,600 cells in total. The horizontal axis is \emph{annual operated area}, not the capacity of one machine: a single baseline unit at \SI{11.6}{decares\per day} over 170 operating days covers ${\approx}\SI{200}{\hectare\per\year}$, so the upper part of the axis ($\gtrsim$\,200\,ha) implies a multi-unit fleet, which the per-hectare economics scale to directly. Diesel is held at the \cref{sec:economics} baseline of \SI{1.40}{\EUR\per\litre} (EU 2024); we deliberately do not run a separate subsidised-diesel scenario because the result below holds with margin to spare under the baseline price.

Three findings flow from \cref{fig:f21}:

\begin{itemize}[leftmargin=1.6em,itemsep=0.15em]
    \item \textbf{In the replacement frame the financial story is uniform across the envelope.} \gls{npv} versus diesel is positive in \textbf{100\,\%} of the 3\,600 cells, and discounted payback never exceeds \SI{3.06}{\year} (median \SI{1.12}{\year}, panel b). At every (\gls{ghi}, farm size) combination we test --- the incremental investment pays back inside the project horizon, within 2 years across most of the plane and at worst ${\approx}\SI{3}{\year}$ for the smallest plots in the cloudiest climates. This is a direct consequence of capex parity (\cref{sec:economics}): the CableTract delta against the diesel reference is only \SI{870}{\EUR}, so there is essentially no capex hurdle to amortise and almost every euro of avoided fuel becomes gross savings. We are explicit that this uniformity is a property of the \emph{replacement} frame: under the \emph{additive} frame (full \SI{35870}{\EUR} financed) only ${\approx}22\%$ of the same cells --- the larger/sunnier corner --- are \gls{npv}-positive, so the ``positive everywhere'' statement is conditional on the buyer retiring a diesel tractor.
    \item \textbf{The binding constraint is annual energy balance, not financial viability.} Panel (a) is an \emph{annual} PV/wind-harvest-minus-demand balance (not the hourly autonomy of \cref{fig:f8}); its breakeven contour divides the plane into an annual-energy-positive region (\textbf{${\approx}85\%$} of cells, smaller operated areas in higher-\gls{ghi} climates) and a region that needs grid backup (large operated areas in low-\gls{ghi} climates). All six bundled reference sites at \SI{25}{\hectare} sit inside the annual-energy-positive region with comfortable surplus (lowest at Beauce, highest at Ludhiana). We are careful not to equate annual surplus with full daily autonomy: the hourly simulation (\cref{fig:f8}) shows Beauce still needs winter grid hours, so ``annual-energy-positive'' is a weaker, separate claim from ``hourly off-grid'' (\cref{sec:r-c3}).
    \item \textbf{CableTract's operating envelope is therefore a \emph{climate envelope}, not a \emph{financial envelope}.} A farmer in Beauce gets a positive \gls{npv} \emph{and} fully off-grid operation at \SI{25}{\hectare}; a farmer in Beauce running \SI{500}{\hectare} on a single CableTract gets a positive \gls{npv} but needs grid backup. This is exactly the opposite of the more common ``the technology pays back but only at scale'' story for clean-energy hardware: CableTract pays back everywhere we test, and the question is where it can be sold as off-grid hardware versus where it needs to be sold as grid-tied.
\end{itemize}

This is the central result of the paper.

\section{Competitor Comparison}\label{sec:competitor}

Public-domain reference numbers for the five vehicle classes named in \cref{sec:related-adjacent} (six representative machines plus CableTract) are recorded in the bundled competitor table.\footnote{\texttt{tables/competitor\_comparison.csv} in the repository.} Selected columns are reproduced in \cref{tab:competitor}.

\begin{table}[H]
\centering\small
\caption{Competitor comparison table. CableTract sits in a unique cell: the only architecture that supports (light/medium) primary tillage, is off-grid capable, and has a short discounted payback. Payback is the replacement-frame value at the \SI{25}{\hectare\per\year} reference (\SI{1.3}{\year}; below one year at $\geq\SI{50}{\hectare\per\year}$); competitor paybacks are manufacturer/press figures at their own scenarios. \emph{Energy/ha} is delivered electrical energy for every row except the diesel tractor, whose \SI{120}{\kWh\per\hectare} is fuel (primary) energy from $12\,\si{\litre\per\hectare}$ at ${\approx}\SI{10}{\kWh\per\litre}$; on a useful-work basis the diesel figure is ${\approx}\SI{36}{\kWh\per\hectare}$ (\cref{sec:r-c2}).}\label{tab:competitor}
\resizebox{\textwidth}{!}{%
\begin{tabular}{lllrrrrl}
\toprule
Vehicle & Form factor & Powertrain & kW & Energy/ha (kWh) & CAPEX (\si{\EUR}) & Payback (\si{\year}) & Off-grid? \\
\midrule
\textbf{CableTract codesigned}    & cable-driven 2-module & PV+wind+battery & 5.0  & \textbf{8.89} & 35\,870 & \textbf{1.3} & \textbf{yes} \\
Monarch MK-V                       & 4WD tractor          & battery EV       & 55   & 52    & 85\,000 & 5--7 & no \\
Farmdroid FD-20                    & autonomous robot     & onboard PV       & 0.4  & 1.5   & 75\,000 & 4--6 & yes \\
Kubota LXe-261                     & compact tractor      & battery EV       & 18   & 30    & 52\,000 & 6--8 & no \\
EcoRobotix ARA                     & spot sprayer         & onboard PV+bat   & 0.6  & 0.4   & 65\,000 & 4--6 & yes \\
Naio Oz/Orio                       & autonomous weeder    & battery          & 2.5  & 5     & 38\,000 & 3--5 & partial \\
Conventional 80 hp diesel          & tractor              & diesel \gls{ice} & 60   & 120   & 35\,000 & ref. & no \\
\bottomrule
\end{tabular}%
}
\end{table}

CableTract sits in a unique cell of this table: among the compared classes it is the only architecture that \textbf{simultaneously} (a) supports light/medium primary tillage (within its codesigned envelope), (b) is off-grid capable, and (c) has a short replacement-frame payback (\SI{1.3}{\year} at \SI{25}{\hectare\per\year}). Monarch MK-V matches it on rated power and operational versatility but is grid-dependent and has 5--7 year payback. Farmdroid FD-20 matches it on off-grid but cannot till. Naio matches it on payback but cannot till and is only partial off-grid.

We deliberately do \emph{not} claim CableTract beats every competitor on every metric. EcoRobotix is more energy-efficient per hectare (\SI{0.4}{\kWh\per\hectare} vs 8.89) because it only sprays. Naio is cheaper. Monarch MK-V is more powerful and operates at conventional speed. The claim is that CableTract occupies a specific cell in the feasibility space --- \emph{medium-load operations on medium-sized farms in solar-rich climates} --- that no other vehicle in the table fills as cleanly.

\section{Results}\label{sec:results}

This section consolidates the per-figure findings of \cref{sec:methods} into a verdict against the four claims of \cref{sec:rqs} (C1--C4) plus the operating envelope; the supporting figures, tables, and derivations live in \cref{sec:methods} and are cross-referenced rather than repeated.

\subsection{C1 --- Compaction}\label{sec:r-c1}

CableTract reduces in-field compacted area by \textbf{97--98\,\%} and the contact-energy index by \textbf{${\approx}\,73\times$} across rectangle, L-shape, and irregular-concave field classes (\cref{fig:f12}, \cref{sec:f12}). The reduction is uniform across field shape because it is geometric, not field-dependent: only the \SI{250}{\kilogram} carriage rolls inside the field, while the \gls{mu} and Anchor stay on the headland. The $73\times$ contact-energy ratio combines the carriage's $4.6\times$ lower mean contact pressure (a $4.6^2 \approx 21\times$ effect, since the integrand goes as $p^2$) with its ${\approx}3.5\times$ smaller total contact area (two narrow rollers versus four tractor wheels). It is a per-contact-patch quantity ($\sum p^2 A_{\text{patch}}$ at equal pass count) --- a property of the running gear, not a field-integrated soil-damage figure --- which is why it is identical across field shapes; the 97--98\,\% \emph{area} reduction is the separate, field-dependent part of the story, and the two should not be multiplied together. Both are computed from open published static contact-pressure values without dynamic \gls{fem} calibration, and we do not claim the index maps linearly onto soil damage or yield.

\subsection{C2 --- Energy}\label{sec:r-c2}

The codesigned reference (regen default) operates at \textbf{\SI{889}{\Wh\per\decare} (\SI{8.89}{\kWh\per\hectare})} of delivered electrical energy. An \SI{80}{hp} \gls{wd} diesel comparator burns ${\approx}\SI{12}{\litre\per\hectare}$ on the same operations~\cite{asabe-d497,siemens1999machinery} --- ${\approx}\SI{120}{\kWh\per\hectare}$ of \emph{primary fuel} energy, of which only ${\sim}30\text{--}35\%$ reaches the drawbar as useful work (${\approx}\SI{36}{\kWh\per\hectare}$). On a like-for-like \emph{useful-work} basis CableTract therefore uses about \textbf{$4\times$ less} energy; on a \emph{primary-fuel} basis the ratio is ${\sim}13\times$, but the extra factor is mostly the diesel engine's thermal-conversion loss --- generic to any electric drivetrain --- not the cable architecture. The architectural ${\sim}4\times$ decomposes into three levers (a unidirectional drivetrain gives \SI{921}{\Wh\per\decare}, only ${\sim}3.6\%$ more energy; the four-quadrant drive's return-leg recovery is a small effect, see \cref{sec:variant-norering}):

\begin{itemize}[leftmargin=1.6em,itemsep=0.1em]
    \item \textbf{Co-designed implements} (\cref{fig:f4,fig:f5}). The median of the per-implement P50 drafts drops from ${\approx}\SI{6.2}{\kilo\newton}$ (conventional library) to ${\approx}\SI{2.5}{\kilo\newton}$ (codesigned library), a factor of ${\approx}\,2.5$, via narrower widths, slower speeds, and shallower depths (with the operating-point caveats of \cref{sec:codesign}).
    \item \textbf{No dead-weight transport.} The draft term $F_d L$ is 72\,\% of the round work and the in-field work ($F_d L + F_i L$) 96\,\% (\cref{sec:pass-model}); the diesel comparator's body-repositioning force, paid over the whole field rather than once per strip, dwarfs this on every operation.
    \item \textbf{Decomposed drivetrain} (\cref{fig:f3}). A premium drivetrain ($\eta = 0.74$) reduces continuous motor power from \SI{1.50}{\kilo\watt} to \SI{1.01}{\kilo\watt}. The codesigned reference uses a \SI{2}{\kilo\watt} continuous / \SI{3}{\kilo\watt} peak \gls{pmsm}.
\end{itemize}

The energy story is therefore not ``CableTract has a magic motor'' but ``CableTract pulls a smaller implement, slower, on a narrower strip, without a 4-tonne body --- and converts electricity instead of burning diesel''. The three architectural levers give the ${\sim}4\times$ useful-work advantage (the four-quadrant drive adds only a few percent on top); the diesel-to-electric conversion step accounts for the further ${\sim}3\times$ that lifts the primary-energy ratio to ${\sim}13\times$.

\subsection{C3 --- Off-grid feasibility}\label{sec:r-c3}

Median harvested-energy throughput across the six bundled sites is \textbf{\SIrange{10}{14}{decares\per day}} at the codesigned hardware (\SI{15}{\meter\squared} \gls{pv}, \SI{9}{\kWh} battery). At Konya, Ludhiana, and São Paulo the system runs grid-free for the entire late-June week we plot in \cref{fig:f7}. At Palencia and Des Moines the system needs several hundred hours of grid backup per year to cover winter shortfalls (Palencia's envelope minimum is \SI{616}{\hour\per\year}, \cref{fig:f8}). At \textbf{Beauce, no combination of panel area (\SIrange{4}{30}{\meter\squared}) and battery capacity (\SIrange{2}{24}{\kWh}) achieves $<\!100$ grid hours/year} under the \SI{6}{\hour\per day} duty cycle (\cref{fig:f8}): the northern temperate climate cannot sustain off-grid operation at any practical hardware scale.

The off-grid claim is \textbf{climate-conditional, not categorical}. We frame this as a \emph{feature} of the analysis: a CableTract sold into Konya is sold as off-grid hardware, while a CableTract sold into Beauce is sold as a low-carbon grid-tied alternative to a diesel tractor. The economics work in both cases (\cref{sec:r-c4}), but the marketing language differs.

\subsection{C4 --- Economics}\label{sec:r-c4}

In the replacement frame (incremental capex \SI{870}{\EUR}), \gls{npv} vs diesel is positive across the entire \SIrange{1}{100}{\hectare\per\year} sweep at all three discount rates (5/8/12\,\%) we test (\cref{fig:f16}, \cref{sec:npv-farmsize}). At the \SI{25}{\hectare\per\year} reference, NPV @ 8\,\% is \textbf{\SI{+3575}{\EUR}} with discounted payback of \textbf{\SI{1.30}{\year}}; at \SI{100}{\hectare\per\year} it is \textbf{\SI{+14360}{\EUR}} (payback \SI{0.47}{\year}). For an additive buyer financing the full machine, the \SI{25}{\hectare\per\year} NPV is instead ${\approx}\SI{-31}{\kilo\EUR}$, turning positive only above ${\sim}\SI{240}{\hectare\per\year}$; the tornado (\cref{fig:f14}) shows the replacement-frame result is most sensitive to the assumed used-tractor price and the maintenance differential, several of which can bring the \SI{25}{\hectare} increment marginally negative.

Lifecycle CO\textsubscript{2} at \SI{25}{\hectare\per\year} is \textbf{\SI{14.6}{\kilogram\per\hectare\per\year}}, against \textbf{32.5} for the diesel reference and \textbf{22.9} for a Monarch-class electric tractor (\cref{fig:f17}). The improvement is entirely on the operational side and does not depend on grid decarbonisation.

The operating envelope of \cref{fig:f21} shows that, \emph{in the replacement frame}, the financial story is uniform: \gls{npv} is positive in 100\,\% of the 3\,600 cells of the (annual \gls{ghi} $\times$ farm size) sweep, with discounted payback bounded above by \SI{3.06}{\year} (median \SI{1.12}{\year}). This follows from capex parity (\SI{35870}{\EUR} versus \SI{35000}{\EUR}): there is no capex hurdle to amortise, and every euro of avoided fuel becomes gross savings. Under the additive frame the \gls{npv}-positive region shrinks to ${\approx}22\%$ of cells, so the uniformity is conditional on the replacement scenario.

The Sobol decomposition (\cref{fig:f13}) confirms that no single parameter dominates the throughput variance (top contributor $\gls{sobolt} \approx 0.25$). The \cref{fig:f14} tornado shows the NPV is led by exogenous factors (the price of the diesel alternative, diesel price and volume) rather than by CableTract design parameters --- though, the incremental NPV being small, several of those exogenous factors can move it across zero, which is why we frame the economics as ``comparable-to-better at parity with strong upside at scale'' rather than ``dominant everywhere''.

\subsection{Operating envelope}\label{sec:r-envelope}

The central result of the paper. In the replacement frame the codesigned CableTract is \gls{npv}-positive in 100\,\% of the 3\,600 cells of the (annual \gls{ghi} $\times$ farm size) sweep, with discounted payback never exceeding \SI{3.06}{\year} (median \SI{1.12}{\year}). The binding \emph{physical} constraint is the \textbf{annual energy-balance contour}, not financial viability: ${\approx}85\%$ of cells are annual-energy-positive (distinct from, and weaker than, the hourly off-grid claim of \cref{sec:r-c3}), and the contour cuts diagonally across the plane. CableTract's operating envelope is therefore a \emph{climate envelope} (does annual harvest cover demand?) more than a \emph{financial envelope} (does the increment pay back?).

This contrasts with the typical clean-energy hardware story, where the technology is climate-virtuous but pays back only at scale or with subsidy. For a farmer replacing a diesel tractor, CableTract pays back its small capex increment across the swept plane; the open questions are where it can be sold as \emph{off-grid} hardware, and --- for a buyer who is not replacing a tractor --- at what farm size the full machine amortises (\cref{sec:r-c4}).

\subsection{Architectural variants}\label{sec:r-pareto}

The variant comparison in \cref{tab:variants-numbers,fig:f20} runs three configurations on the same parameter set: the codesigned baseline (regen default), CableTract+ (4-Main-Unit planar cable robot), and a unidirectional drivetrain (no regen). CableTract+ delivers $2.61\times$ baseline throughput at $2.28\times$ CAPEX (\SI{81784}{\EUR} vs \SI{35870}{\EUR}), so it is a throughput multiplier rather than an economics multiplier and stretches simple payback from 121.3 to 145.3 months --- worthwhile only where calendar time, not capital, binds. Dropping regenerative braking (the unidirectional row) raises flat-field energy intensity by only ${\approx}3.6\%$ (889 $\to$ 921 Wh/decare) and lengthens simple payback to 124.6 months for a \SI{300}{\EUR} capex saving; the four-quadrant drive is kept standard mainly for smooth depth-control torque and slope recovery, not as a large flat-field energy lever.

\section{Higher-fidelity verification studies}\label{sec:verification}

The screening model of \cref{sec:methods} lumps each subsystem into a reduced-order form fit for a feasibility envelope. To stress-test the five places a sceptical reviewer is most likely to press --- the robustness of the co-design lever, the anchor capacity, the depth-control loop, the cable safety surface, and the tillage draft --- we built five purpose-specific higher-fidelity models: a full propagation of the co-design sweep, a nonlinear $p$--$y$ Winkler anchor model, a closed-loop multibody-cable depth-control co-simulation, a finite-difference cable-safety model, and a soft-sphere discrete-element (DEM) soil--tool model. These do \emph{not} replace a prototype (\cref{sec:open-risks} remains in force); they bound each risk more tightly than the screening model, and where warranted they revise a number or a framing. Every study is deterministic and reproducible from the open pipeline (\texttt{scripts/run\_phase8}--\texttt{12}, with unit tests), and each is validated against an analytic limit before it is trusted.

\subsection{Co-design robustness (S4)}\label{sec:v-s4}

\begin{figure}[H]
\centering
\includegraphics[width=0.95\linewidth]{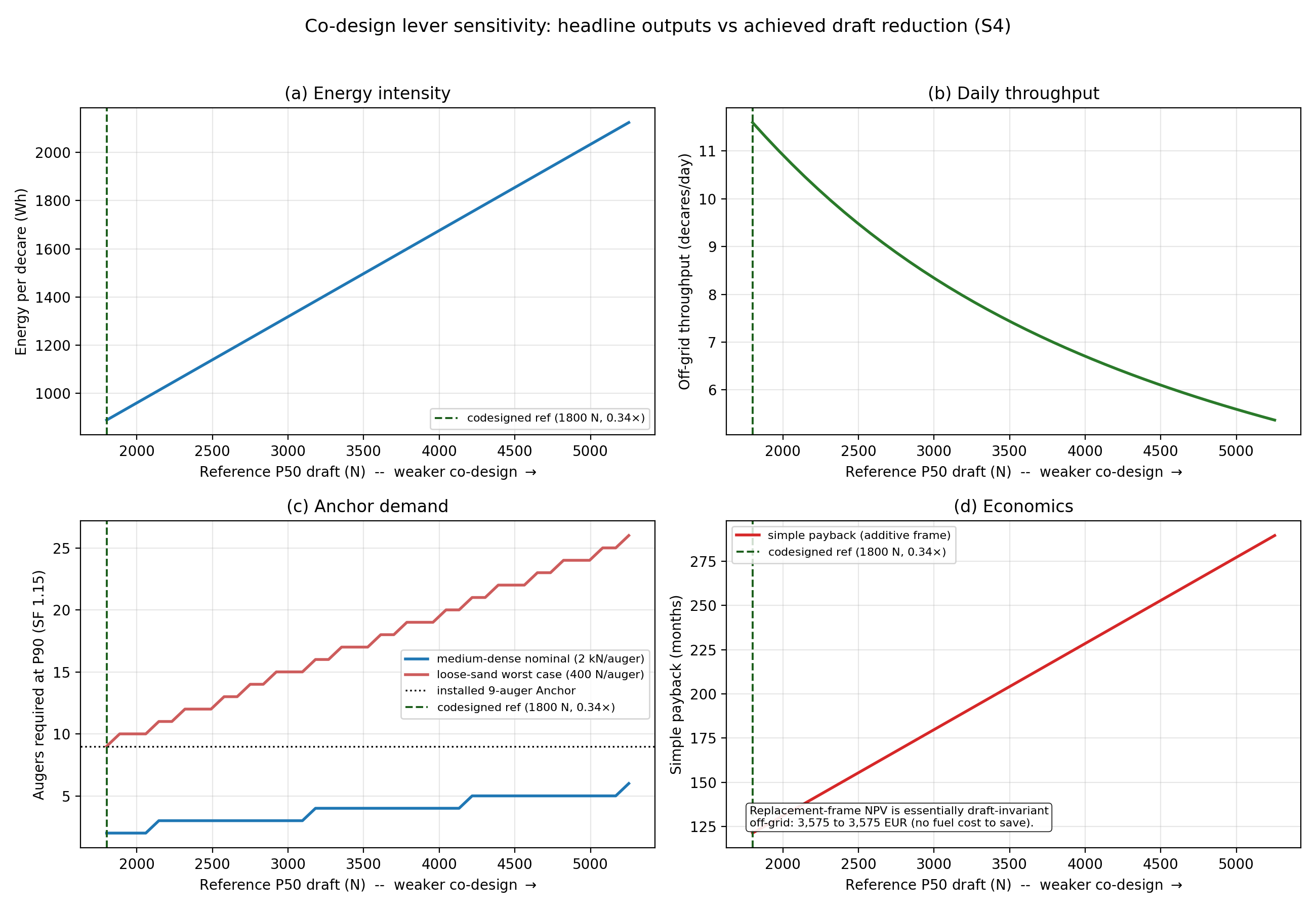}
\caption{Co-design lever sensitivity (S4). As the achieved reference draft rises from the codesigned point (\SI{1.8}{\kilo\newton}, the claimed ${\approx}0.37\times$ reduction) toward the conventional-library median, (a) energy per decare and (b) off-grid throughput degrade and (c) the required auger count climbs, while (d) the off-grid replacement-frame \gls{npv} stays essentially flat --- the robustness finding.}
\label{fig:f22}
\end{figure}

The entire case rests on the co-designed library cutting median draft to ${\approx}0.37\times$ the conventional library --- a model prediction, not a measurement. We ask the reviewer's question directly: if co-design under-delivers, how much survives? Sweeping the achieved reference draft from the codesigned point (\SI{1.8}{\kilo\newton}) up to the conventional-library median (\SI{5.25}{\kilo\newton}) and propagating every level through the \emph{full} deterministic pipeline (\cref{fig:f22}), the energy intensity and off-grid throughput degrade smoothly (at a $0.60\times$ shortfall, \SI{889}{\Wh\per\decare} rises to ${\approx}\SI{1383}{\Wh\per\decare}$ and daily throughput falls from ${\approx}11.6$ to ${\approx}8.0$ decares/day), and the required auger count rises --- but the off-grid replacement-frame \gls{npv} is essentially \emph{invariant}, because in the off-grid frame there is no residual fuel cost for a higher draft to erode. This is the quantitative form of \cref{sec:codesign-sensitivity}: the economics are robust to a co-design shortfall; the energy and off-grid claims are the sensitive ones.

\subsection{Anchor lateral capacity from a nonlinear \texorpdfstring{$p$--$y$}{p-y} model (S7)}\label{sec:v-s7}

\begin{figure}[H]
\centering
\includegraphics[width=0.95\linewidth]{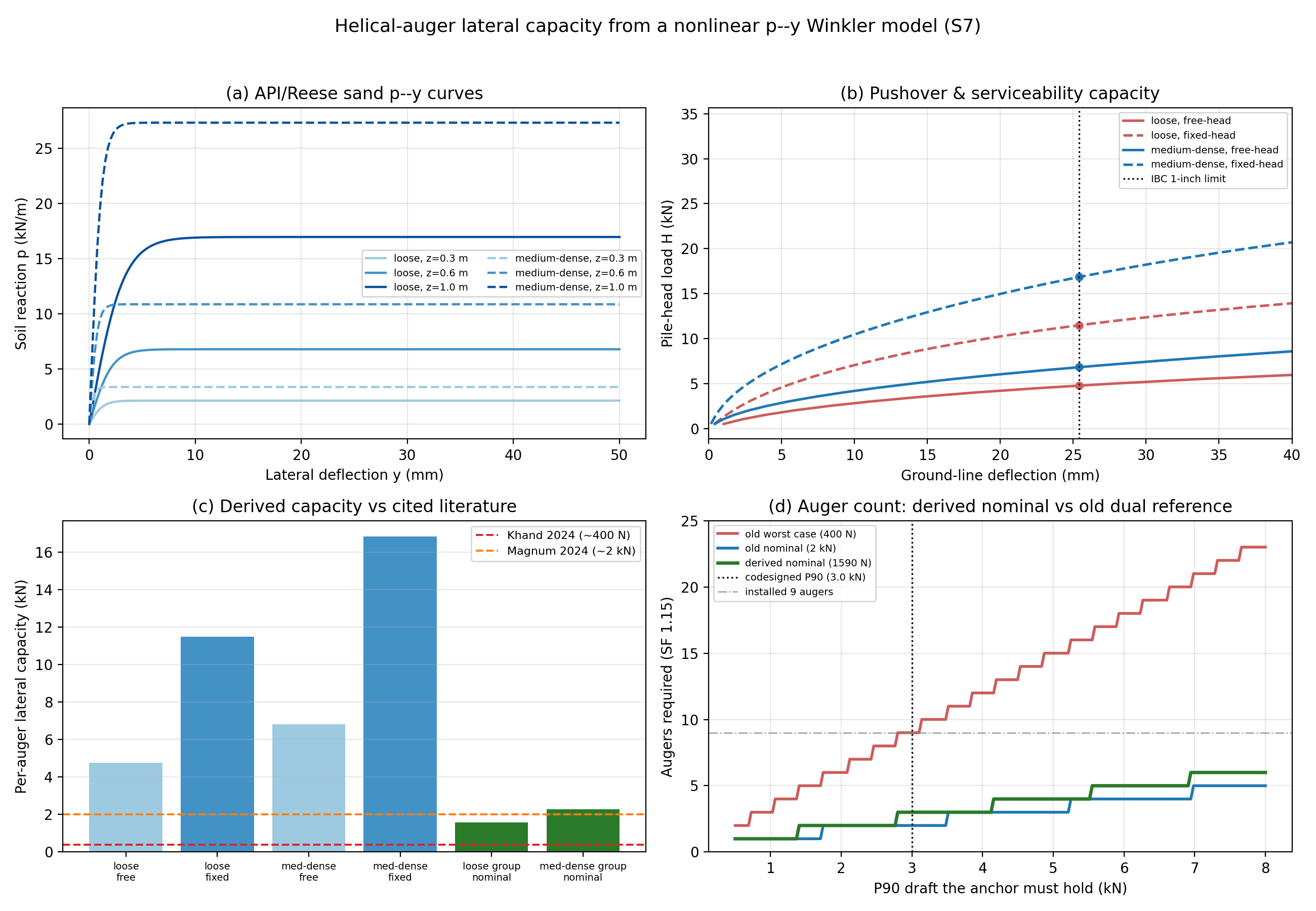}
\caption{Helical-auger lateral capacity from the nonlinear $p$--$y$ Winkler model (S7): (a) API/Reese sand $p$--$y$ curves; (b) pile-head pushover with the IBC \SI{1}{inch} serviceability capacity marked; (c) derived per-auger capacity versus the cited Khand/Magnum references; (d) auger count versus the draft the anchor must hold, for the derived nominal against the old dual reference.}
\label{fig:f28}
\end{figure}

\Cref{sec:anchor} deliberately bracketed the per-auger lateral capacity between two literature references (\SI{400}{\newton} Khand~\cite{khand2024helical}, \SI{2}{\kilo\newton} Magnum~\cite{magnum2024fixed}) rather than picking one. Here we replace that ``have-it-both-ways'' gap with a single, internally derived nominal. We solve the Euler--Bernoulli beam-column on a nonlinear elastic (Winkler) foundation, $EI\,y'''' = -p(y,z)$, with the API RP 2A~\cite{api2014rp2a} / Reese~\cite{reese1974analysis} sand $p$--$y$ curves $p = A\,p_u\tanh(kzy/A p_u)$ (coefficients after Murchison \& O'Neill~\cite{murchison1984evaluation}), as a nonlinear boundary-value problem (finite differences $+$ Newton) for the lateral capacity at the IBC 2021 \SI{1}{inch} (\SI{25.4}{\milli\meter}) head-deflection serviceability limit, free- and fixed-head, with a \(3\times3\) cluster handled by row $p$-multipliers (Mokwa~\cite{mokwa1999resistance}; Reese \& Van Impe~\cite{reese2011single}). For a \SI{73}{\milli\meter} steel shaft embedded \SI{2}{\meter}, the single-pile serviceability capacity is \SI{4.8}{\kilo\newton} (loose, free-head) to \SI{16.8}{\kilo\newton} (medium-dense, fixed-head); the group efficiency is \(0.50\), giving a derived per-auger \emph{working} nominal of \SIrange{1.6}{2.3}{\kilo\newton} at a safety factor of \(1.5\). This brackets the Magnum \SI{2}{\kilo\newton} datasheet value and sits above the conservative Khand \SI{400}{\newton} floor (which the model now reads as a deliberately pessimistic lower bound, not a co-equal reference). The linear small-deflection limit reproduces the Het\'enyi~\cite{hetenyi1946beams} closed form $y_0 = H/(2\beta^3 EI)$ to within \SI{0.03}{\percent} (\cref{fig:f28}).

\textbf{Design decision: the \(9\)-auger Anchor is retained.} At the codesigned P90 of \SI{3.0}{\kilo\newton}, the derived nominal implies only ${\approx}3$ augers. We nonetheless keep the installed count at \(9\), which the $p$--$y$ result reframes as a deliberate ${\approx}3\times$ capacity margin --- and as redundancy against a single auger landing in a weak pocket or loosening over a season (\cref{sec:open-risks}) --- rather than as a minimum-spec count. The extra augers carry \emph{no throughput penalty}: they belong to the stationary Anchor and are driven in once, concurrently, at strip setup (\cref{sec:open-risks}, parallel-drive nominal), never during the per-pass traverse. Their only cost is the additional \gls{bom} and embodied steel already carried in \cref{sec:economics}, plus a little setup labour. One caveat bounds the claim conservatively: the model computes the \emph{lateral} capacity of near-vertical piles resisting the horizontal cable tension; augers raked toward the horizontal would instead resist through axial helix pull-out, which for a helical pile in sand is typically higher than its lateral capacity, so the \(9\)-auger design is, if anything, more conservative under that geometry.

\subsection{Closed-loop tilling-depth control (S1)}\label{sec:v-s1}

\Cref{sec:open-risks} names the depth-control loop the single largest unresolved risk. We give it a closed-loop co-simulation. A planar lumped-mass cable (axial Hookean springs $+$ gravity) reproduces the \cref{sec:catenary} catenary sag to \SI{0.0}{\percent}, the taut-string first natural frequency $f_1 = \tfrac{1}{2L}\sqrt{T/\mu}$ to \SI{0.03}{\percent}, and conserves energy in undamped free vibration to \SI{0.008}{\percent}. It yields the cable's vertical point stiffness at midspan, ${\approx}4T/L \approx \SI{144}{\newton\per\meter}$ --- some three orders of magnitude softer than a gauge wheel on firm soil (${\sim}\SI{3e5}{\newton\per\meter}$), confirming quantitatively the \cref{sec:tension-balance} claim that \emph{the gauge wheel, not the cable, sets depth}. We then drive a reduced vertical plant (carriage mass, a unilateral gauge-wheel ground contact, and a bandwidth- and rate-limited down-pressure actuator) with three programmed disturbances --- a buried-stone impulse, a hardpan step, and a moisture ramp --- under a tuned PID and an offset-free model-predictive controller (a self-contained box-constrained QP), against a heavy-tractor benchmark.

\begin{figure}[H]
\centering
\includegraphics[width=0.95\linewidth]{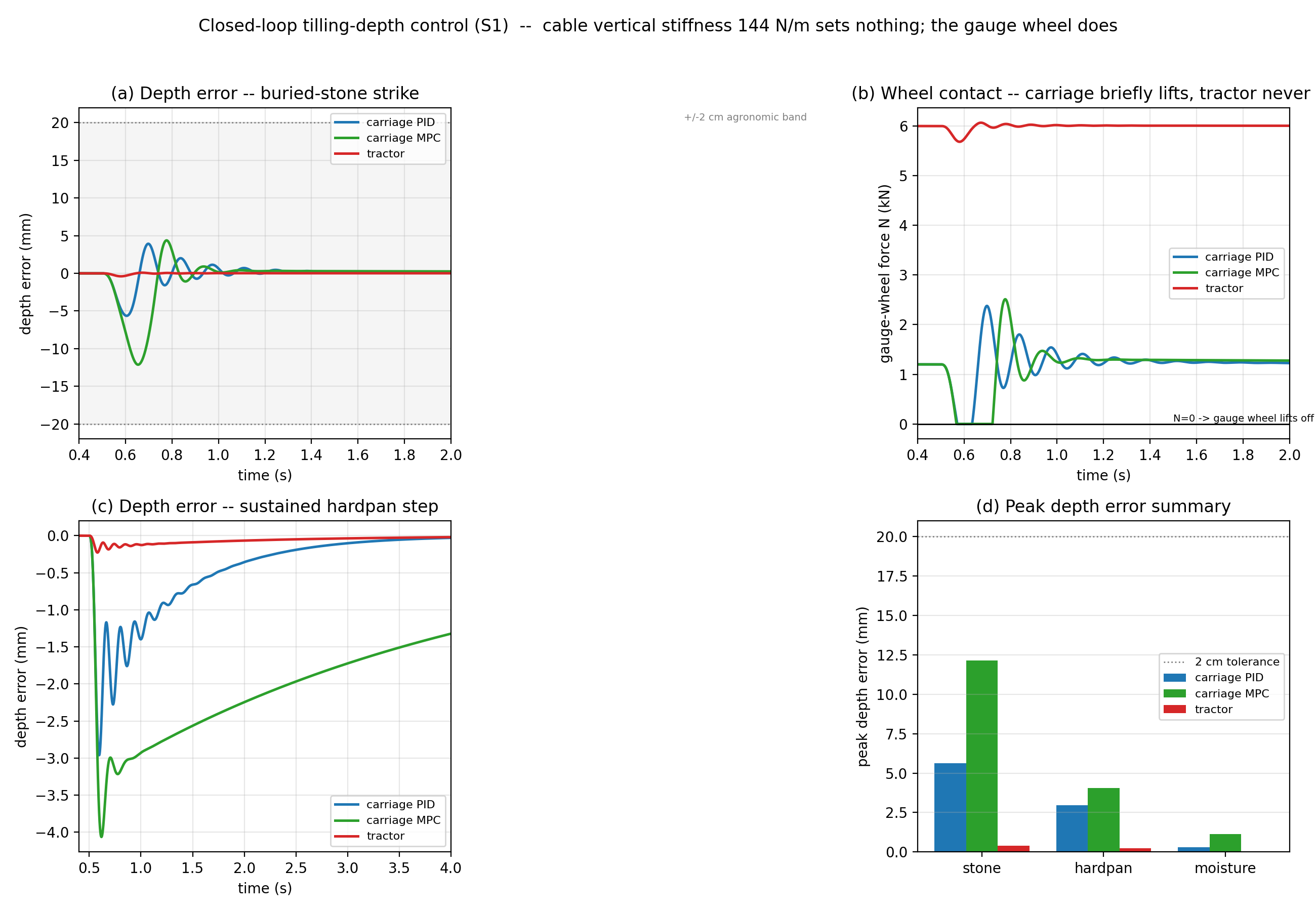}
\caption{Closed-loop depth control (S1). Depth error and gauge-wheel contact force for the light cable carriage (PID and MPC) versus a heavy tractor under a buried-stone strike and a hardpan step; all controllers hold the \(\pm\SI{2}{\centi\meter}\) agronomic tolerance, but the light carriage's gauge wheel briefly unloads (\(N\!\to\!0\)) where the heavy tractor's never does.}
\label{fig:f23}
\end{figure}

\begin{figure}[H]
\centering
\includegraphics[width=0.95\linewidth]{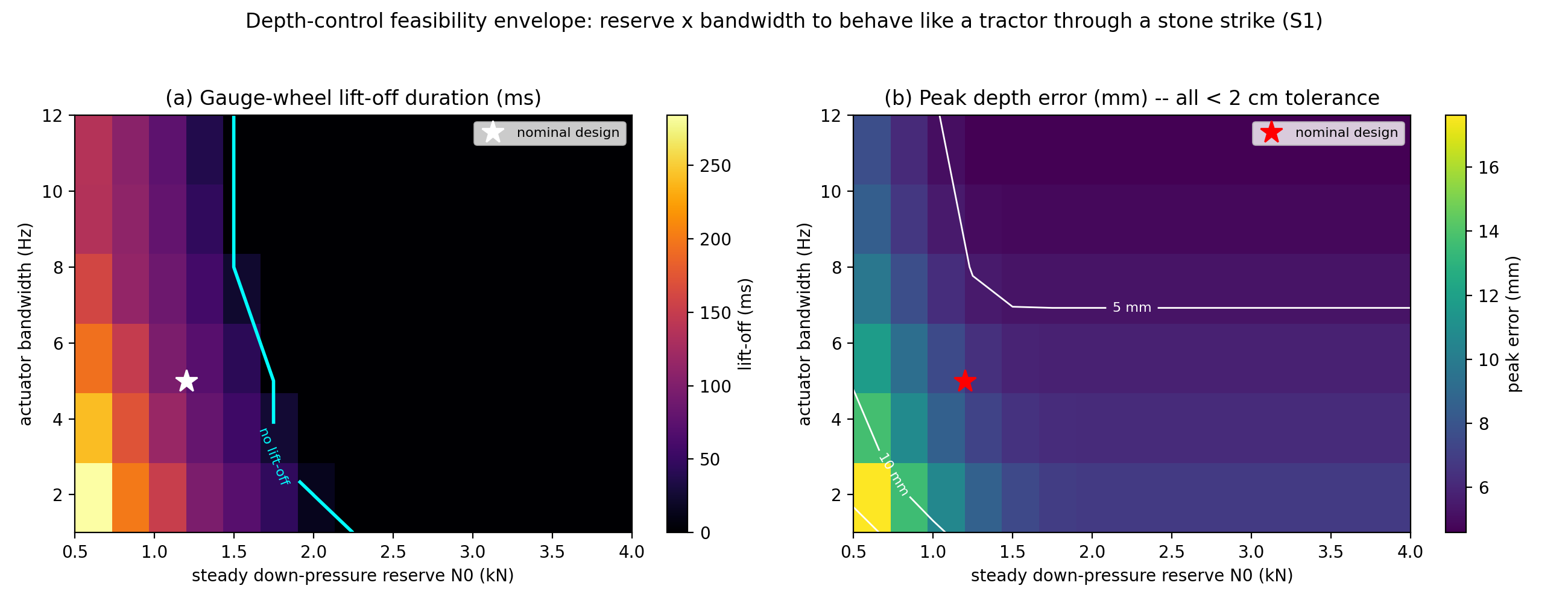}
\caption{Depth-control feasibility envelope (S1): gauge-wheel lift-off duration (a) and peak depth error (b) over the steady down-pressure reserve $N_0$ and the actuator bandwidth. A reserve of ${\approx}\SI{1.75}{\kilo\newton}$ at ${\geq}\SI{5}{\hertz}$ removes lift-off in a \SI{1.5}{\kilo\newton} stone strike; peak depth error stays within the \(\pm\SI{2}{\centi\meter}\) tolerance throughout.}
\label{fig:f24}
\end{figure}

All three controllers hold the depth error inside the \(\pm\SI{2}{\centi\meter}\) agronomic tolerance for every disturbance. The heavy tractor never lifts off (peak error \SI{0.4}{\milli\meter}; contact force stays ${\sim}\SI{5.7}{\kilo\newton}$). The light carriage holds depth too (PID peak \SI{5.6}{\milli\meter} on the stone strike) but its gauge wheel \emph{briefly lifts off} (${\leq}\SI{0.15}{\second}$) when the transient lift exceeds the steady wheel reserve faster than the actuator can respond. The feasibility envelope over the down-pressure reserve and actuator bandwidth (\cref{fig:f24}) gives the design specification: a steady down-pressure reserve of ${\approx}\SI{1.75}{\kilo\newton}$ at an actuator bandwidth ${\geq}\SI{5}{\hertz}$ eliminates gauge-wheel lift-off through a \SI{1.5}{\kilo\newton} stone strike. The honest reading is that depth control is \emph{feasible but must be engineered}: unlike a tractor, which holds depth for free with several tonnes of inertia, the cable carriage must carry an active down-pressure reserve, which trades against compaction (\cref{sec:compaction}). This narrows the open risk but does not close it --- the quarter-scale soil-bin bench rig of \cref{sec:open-risks} remains the right next step.

\subsection{Cable safety and durability (S2)}\label{sec:v-s2}

\begin{figure}[H]
\centering
\includegraphics[width=0.95\linewidth]{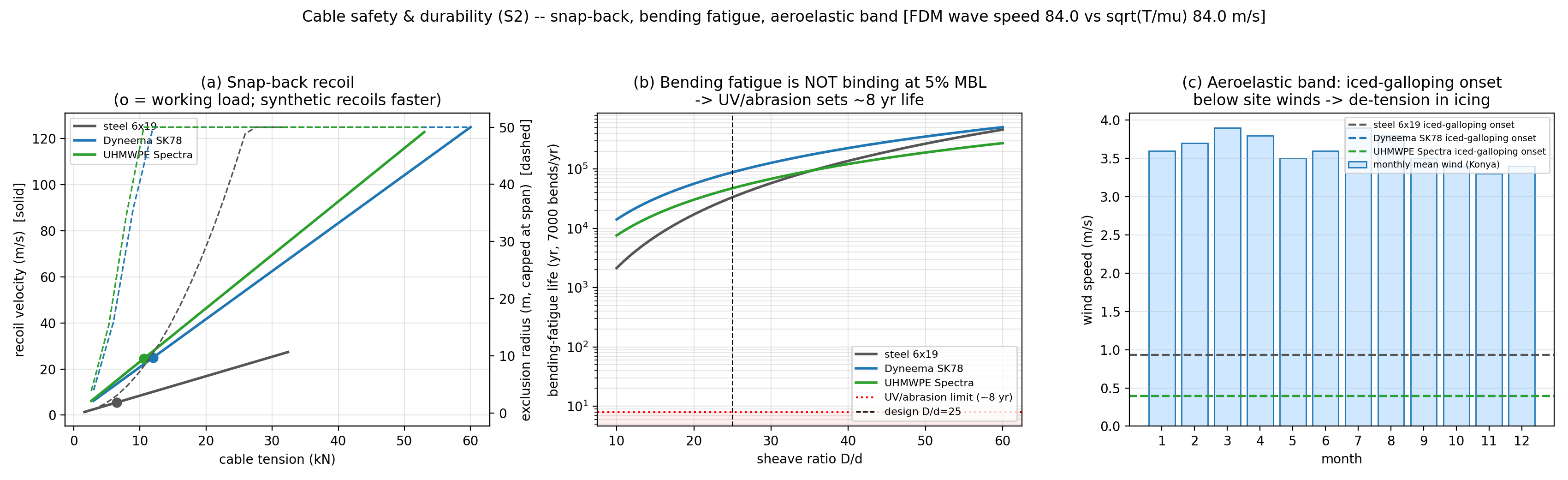}
\caption{Cable safety and durability (S2): (a) snap-back recoil velocity and exclusion radius versus tension for steel, Dyneema, and UHMWPE; (b) bending-fatigue life versus sheave ratio $D/d$, showing the working load is UV/abrasion-limited, not fatigue-limited; (c) aeroelastic band --- iced-galloping onset against the bundled monthly mean winds.}
\label{fig:f25}
\end{figure}

\Cref{sec:open-risks} flags the cable as an unquantified safety surface with no replacement schedule in the \gls{bom}. We address both with three sub-models (\cref{fig:f25}). \textbf{Snap-back recoil:} the axial release velocity of a suddenly-freed tensioned end is $v = T/\sqrt{EA\,m'}$; at working load the \emph{lighter synthetic recoils faster} (\SI{8}{\milli\meter} Dyneema ${\approx}\SI{25}{\meter\per\second}$ versus steel ${\approx}\SI{5.5}{\meter\per\second}$, because of its much lower mass per length), so synthetic rope is \emph{not} inherently safer for snap-back despite its higher break load --- the recoil kinetic energy exceeds projectile-injury thresholds, so a full-span exclusion zone during tensioned operation is a hard requirement. A finite-difference transverse-wave solver validates the wave speed $c = \sqrt{T/\mu}$ to \SI{0.03}{\percent}. \textbf{Bending-over-sheave fatigue:} a Feyrer-type model~\cite{feyrer2015wireropes} shows that at the working load of only ${\approx}5\%$ of break load the bending-fatigue life is effectively unbounded (${\sim}10^4$ yr at the design $D/d=25$); replacement is therefore governed by UV and abrasion at ${\approx}\SI{8}{\year}$, which we add as a new operating-cost and life-cycle line (${\approx}\SI{60}{\EUR\per\year}$ steel, ${\approx}\SI{180}{\EUR\per\year}$ Dyneema; a few \si{\kilogram} CO\textsubscript{2}/yr) in \cref{sec:economics}. \textbf{Aeroelastic stability:} the long taut cable has a dense modal spectrum, so vortex-induced vibration~\cite{blevins1990flow} is possible at the bundled site winds and a deployed cable would need dampers; the Den Hartog~\cite{denhartog1956mechanical} galloping onset for an iced section (\SIrange{0.4}{0.9}{\meter\per\second}) sits below the site winds, so the cable should be de-tensioned in icing conditions --- a constraint largely mitigated by growing-season-only operation.

\subsection{Soil--tool DEM of the co-designed implements (S3)}\label{sec:v-s3}

\begin{figure}[H]
\centering
\includegraphics[width=0.95\linewidth]{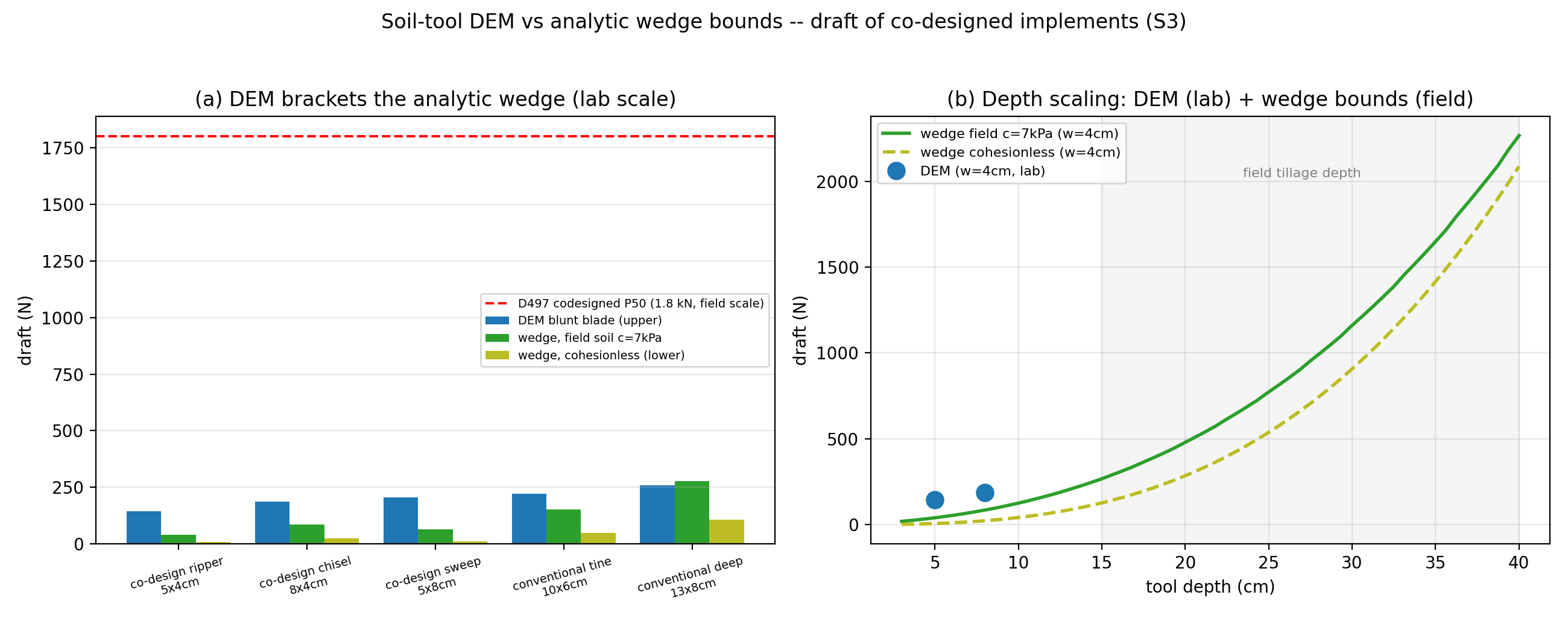}
\caption{Soil--tool DEM versus analytic wedge bounds (S3): (a) DEM draft brackets the slender-tine wedge (cohesionless lower bound and field-cohesion estimate) at lab scale; (b) depth scaling --- DEM lab points with the wedge bounds extrapolated to field depth, where the field-cohesion wedge meets the D497 codesigned reference.}
\label{fig:f26}
\end{figure}

\begin{figure}[H]
\centering
\includegraphics[width=0.95\linewidth]{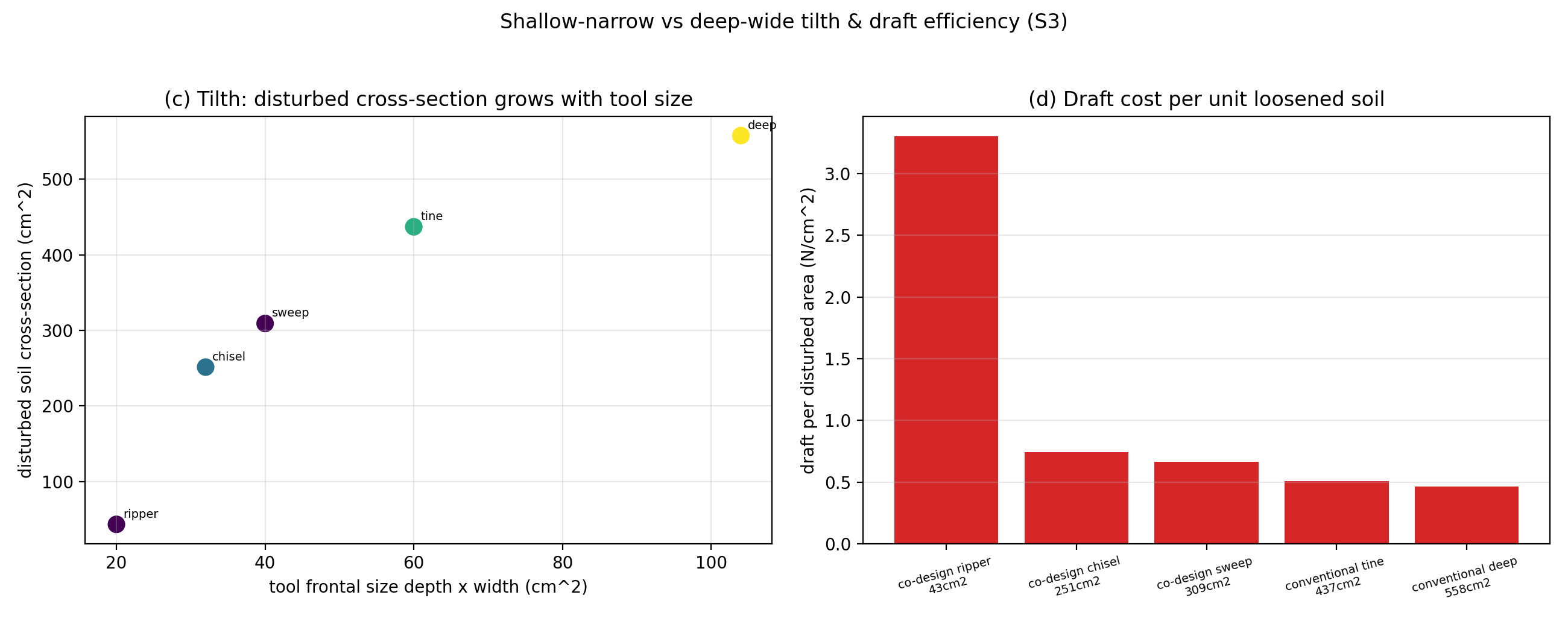}
\caption{Shallow-narrow versus deep-wide tilth (S3): (c) disturbed soil cross-section grows with tool size --- a shallow-narrow co-design pass disturbs far less soil than a deep-wide conventional one; (d) draft per unit disturbed area, showing the narrow tool's higher cost per unit loosened soil.}
\label{fig:f27}
\end{figure}

Finally we test the co-design draft prediction and the shallow-narrow-versus-deep-wide tilth question with a three-dimensional soft-sphere DEM (Hertz--Mindlin contacts, ${\approx}8000$ particles), cross-checked against an analytic McKyes--Godwin~\cite{mckyes1985soil} / Reece~\cite{reece1965fundamental} soil-cutting wedge (Godwin \& Spoor~\cite{godwin1977soil}). The bed packs to a realistic tilled-soil bulk density of \SI{1625}{\kilogram\per\meter\cubed}, and the drag force rises with tool depth, width, and inter-particle friction as expected; the field-cohesion wedge, extrapolated to field tillage depth, reaches ${\approx}\SI{2.3}{\kilo\newton}$ at \SI{40}{\centi\meter} --- consistent with the codesigned D497 P50 of \SI{1.8}{\kilo\newton} (\cref{fig:f26}). On tilth (\cref{fig:f27}), a shallow-narrow co-designed pass disturbs ${\approx}13\times$ less soil cross-section at ${\approx}$ half the draft of a deep-wide conventional pass, which supports the co-design less-energy/less-disturbance argument --- with the honest nuance that the narrow tool costs \emph{more} draft per unit of soil loosened, so its advantage is in needing only a small loosened zone (a seed slot or a controlled-traffic strip), not in loosening a large area more cheaply. We flag the modelling limit plainly: the blunt full-depth DEM blade over-predicts and the slender-tine cohesionless wedge under-predicts the absolute draft, so the two idealisations \emph{bracket} the true value (with the D497 reference between them) --- the robust DEM outputs are the trends and the tilth comparison, not the absolute draft.

\medskip
Taken together, these five studies tighten four of the open risks of \cref{sec:open-risks} (depth control, anchor capacity, cable safety, and co-design draft) and revise two artefacts of the screening analysis: the per-auger anchor nominal (\cref{sec:anchor}) and a previously missing cable-replacement cost line (\cref{sec:economics}). They do not close the risks --- a prototype season remains the arbiter --- but they replace ``not simulated'' with ``simulated at higher fidelity, with the following bounds.''

\section{Discussion}\label{sec:discussion}

\subsection{Where CableTract wins}\label{sec:wins}

The combined evidence of \cref{sec:r-c1,sec:r-c2,sec:r-c3,sec:r-c4,sec:r-envelope} points to a clear operating envelope:

\begin{itemize}[leftmargin=1.6em,itemsep=0.15em]
    \item \textbf{Operation type:} medium-load --- light primary tillage (narrow ripper / chisel up to ${\approx}\SI{3}{\kilo\newton}$ P50), all secondary tillage, all seeding, all spraying, all weeding, all mowing. The codesigned implement library is built for this envelope.
    \item \textbf{Farm size:} $\geq\!\SI{10}{\hectare}$ annual operating area. \gls{npv} is positive everywhere we test, but the \emph{fastest} paybacks ($<1$ year) require $\geq\!\SI{25}{\hectare}$. Below \SI{10}{\hectare}, the 1.2--1.9 yr payback is still competitive but the absolute \gls{npv} is small.
    \item \textbf{Climate:} anywhere with $>\!\SI{1500}{\kWh\per\meter\squared\per\year}$ annual \gls{ghi} for fully off-grid operation. Mediterranean climates (1300--1500) work with summer-only off-grid; northern temperate climates ($<\!1200$) work as grid-tied.
    \item \textbf{Field shape:} rectangular and L-shaped fields with $\geq\!\SI{1}{\hectare}$ individual size. Small fragmented fields hit the polygon shape efficiency wall (median $\eta \approx 0.68$ on the irregular-concave class).
\end{itemize}

\subsection{Where it doesn't (binding constraints)}\label{sec:loses}

The constraints to flag:

\begin{itemize}[leftmargin=1.6em,itemsep=0.15em]
    \item \textbf{Heavy codesigned primary tillage in loose-sand soils} sits outside the 9-auger Anchor envelope on the conservative Khand (2024)~\cite{khand2024helical} reference (\cref{sec:anchor}). Specifically, the narrow sweep (\SI{5.2}{\kilo\newton} P50), narrow chisel (\SI{4.2}{\kilo\newton}), narrow disk (\SI{3.2}{\kilo\newton}), and codesign drill (\SI{4.6}{\kilo\newton}) all require 12--18 augers in loose sand (\cref{fig:f4}). On these sites the operating envelope shrinks to the lighter half of the codesigned library --- cultivator, planter, rotary hoe, sprayer, mower, and the narrow ripper at its P50. In medium-dense soils with fixed-head pile fixity (Magnum 2024~\cite{magnum2024fixed} reference) the same 9-auger Anchor carries the entire codesigned library with $2\times$--$10\times$ margin, so the binding case is specifically loose-sand sites attempting heavy primary tillage --- not heavy primary tillage in general.
    \item \textbf{Off-grid in continental winter climates} is not achievable at any practical hardware scale (Beauce: \SI{911}{\hour\per\year} grid hours minimum). CableTract still pays back as a grid-tied diesel-replacement in these climates, but the marketing language must drop ``off-grid''.
    \item \textbf{Small fragmented fields} below ${\approx}\SI{0.5}{\hectare}$ individual size hit the polygon shape efficiency wall and lose 30--50\,\% of nominal throughput. CableTract is not the right architecture for vegetable garden plots.
    \item \textbf{Operation diversity vs throughput} is a real trade-off. A system specialised to a single operation (sprayer-only, planter-only) can be sized for that operation's draft and width and would dominate the codesigned reference on throughput per day for that one task --- but it cannot do the rest of the agricultural calendar. The codesigned reference deliberately picks the design point that supports all 10 implements in \cref{tab:codesigned-library} rather than optimising any single one.
\end{itemize}

\subsection{Sensitivity to the central design choice}\label{sec:codesign-sensitivity}

The single most consequential choice in the paper is the \textbf{co-designed implement library}. Without the co-design framing --- that is, with conventional tractor implements substituted in place of the codesigned ones --- the analysis weakens in three concrete places:

\begin{enumerate}[leftmargin=1.6em,itemsep=0.1em]
    \item The \SI{2}{\kilo\watt} \gls{pmsm} is undersized (peak draft $\times$ tractor speed requires ${\approx}\SI{5}{\kilo\watt}$), and the inverter, gearbox, and cooling stack scale up with it.
    \item The energy-per-decare claim weakens by ${\approx}2\times$ (because the draft drops by ${\approx}2\times$ under co-design via the strip-width and $v^2$ levers), which directly weakens the off-grid PV+battery sizing.
    \item The 9-auger Anchor loses its worst-case-soil margin: a conventional chisel/sweep plow at \SIrange{14}{15}{\kilo\newton} P50 would need 41--45 augers on the worst-case Khand (2024)~\cite{khand2024helical} loose-sand bound, although the same load fits in 9 augers on the Magnum (2024)~\cite{magnum2024fixed} medium-dense nominal.
\end{enumerate}

The first two bullets are the load-bearing ones --- they hold for \emph{any} soil regime --- and the economics still work even in this degraded scenario (\gls{npv} stays positive at \SI{25}{\hectare\per\year}), but the off-grid story and the compaction story are weaker. The co-designed framing is therefore not a marketing convenience; it is the load-bearing assumption of the paper for energy and motor cost reasons that are independent of soil conditions, with the worst-case-soil anchor envelope as a bonus third leg, and we make this explicit. \Cref{sec:v-s4} quantifies this by propagating a partial co-design shortfall through the full pipeline: the energy and off-grid claims degrade smoothly while the off-grid replacement-frame \gls{npv} stays essentially invariant, confirming which legs are sensitive.

\subsection{What the paper does not claim}\label{sec:not-claim}

We do not claim:

\begin{itemize}[leftmargin=1.6em,itemsep=0.1em]
    \item That CableTract is a finished product or has been prototyped.
    \item That cable-driven agriculture is a new idea (US 8,763,714 B2~\cite{orlando2014patent} predates this work).
    \item That CableTract beats every competitor on every metric (EcoRobotix is more efficient per hectare on spraying-only; Naio is cheaper).
    \item That off-grid operation is universal (it is climate-conditional).
    \item That the $73\times$ contact-energy reduction translates linearly into a yield increase (the soil-yield link depends on crop, climate, rotation, and irrigation).
\end{itemize}

We do claim, with every number reproducible from the open analytical pipeline:

\begin{itemize}[leftmargin=1.6em,itemsep=0.1em]
    \item A ${\sim}4\times$ reduction in useful (drawbar) energy per hectare versus a diesel tractor (${\sim}13\times$ on a primary-fuel basis, most of the extra factor being generic ICE-to-electric conversion efficiency).
    \item A 97--98\,\% reduction in compacted area and a ${\approx}\,73\times$ reduction in contact-energy index.
    \item Off-grid feasibility in solar-rich climates with \SI{15}{\meter\squared} \gls{pv} + \SI{9}{\kWh} battery.
    \item \gls{npv} positive across \SIrange{1}{100}{\hectare} at 5/8/12\,\% discount rates, with discounted payback $<\!\SI{1}{\year}$ at $\geq\!\SI{25}{\hectare}$.
    \item Lifecycle CO\textsubscript{2} $2.2\times$ lower than diesel, $1.6\times$ lower than a Monarch-class electric tractor.
    \item An NPV-positive operating envelope in 100\,\% of 3\,600 (annual \gls{ghi} $\times$ farm size) cells, with discounted payback never exceeding \SI{3.06}{\year}.
\end{itemize}

\section{Limitations}\label{sec:limitations}

This section separates two categories of limitation. \Cref{sec:analytical-limitations} lists the \emph{analytical} limitations of the present model --- what the framework does not include. \Cref{sec:open-risks} lists the \emph{engineering} risks that a prototype phase would have to resolve regardless of how comprehensive the analytical framework becomes; in the authors' candid view these are the dominant near-term feasibility questions for the architecture and they are not closed by additional simulation.

\subsection{Analytical limitations of the present model}\label{sec:analytical-limitations}

\begin{enumerate}[leftmargin=1.8em,itemsep=0.2em]
    \item \textbf{No prototype.} The strongest objection. Every number in the paper is from first-principles calculation; none has been validated against a working machine. We have tried to mitigate this by using \gls{asabe} D497.7~\cite{asabe-d497} (an experimentally calibrated standard), the conservative end of the helical-pile lateral capacity literature (Khand et al.\ 2024~\cite{khand2024helical} for loose-sand free-head bounds, Magnum Piering 2024~\cite{magnum2024fixed} design tables and the CHANCE Technical Design Manual~\cite{chance2024manual} for the medium-dense fixed-head reference), Keller and Lamand\'e (2010)~\cite{keller2010compaction} (experimentally measured tractor contact pressures), open published \gls{lca} inventories~\cite{hammond2019ice,emilsson2019battery,fraunhoferise2022pv,hawkins2013ev,defra2023factors,eea2023electricity}, and a Sobol global sensitivity analysis~\cite{sobol2001indices,saltelli2008primer,herman2017salib} to bound the uncertainty.
    \item \textbf{Reduced-order, not full multiphysics, control simulation.} \Cref{sec:v-s1} now adds a closed-loop depth-control co-simulation (a lumped-mass cable plus a unilateral gauge-wheel contact and a bandwidth-limited down-pressure actuator under PID and MPC), and \cref{sec:v-s7,sec:v-s2} add a nonlinear $p$--$y$ anchor model and a transient cable-recoil model. These are reduced-order, single-axis models validated against analytic limits, not a coupled full-multiphysics simulation of cable tension, three-dimensional anchor stability, and depth control together; that coupling, and its validation against hardware, remains for a prototype phase. \Cref{sec:open-risks} continues to treat the depth-control loop as the dominant unresolved risk, now bounded rather than unexamined.
    \item \textbf{Static contact-pressure compaction model.} No dynamic load transfer, no tyre deflection, no multi-pass cumulative deformation, no soil-yield link, and the carriage load excludes the vertical reaction of cable tension and tool downforce. The model is intentionally simple; a richer model would add compaction to the tractor side (multi-pass deformation) but also to the carriage side (repeated narrow-strip trafficking, added carriage downforce), so we do not assume it can only favour CableTract.
    \item \textbf{Polygon corpus is small and curated.} 50 fields, hand-tuned, biased toward small awkward shapes. A real deployment would need a corpus pulled from EuroCrops or \gls{usda} at scale.
    \item \textbf{PV harvest model in the operating-envelope sweep is a one-parameter linear fit.} The full hourly TMY pipeline lives in \cref{sec:energy} and was used for \cref{fig:f6,fig:f7,fig:f8}, but the (annual \gls{ghi} $\times$ farm size) sweep in \cref{fig:f21} uses a calibrated linear approximation for tractability. The $\alpha = 0.169$ value was fitted as the median across the 6 bundled sites.
    \item \textbf{The codesigned implement library is a paper exercise.} The 10 implements are sized using the \gls{asabe} D497.7 equation but have not been physically built. The draft reductions are predictions, not measurements.
    \item \textbf{The competitor comparison uses public-domain manufacturer claims.} Some of these may not reproduce in the field. The source of every number is flagged in the bundled competitor table.
\end{enumerate}

\subsection{Open mechanical risks}\label{sec:open-risks}

The seven items above are the analytical limitations of the present model: they describe what the framework does not include. This subsection separates out a different category --- \textbf{the engineering risks that a prototype phase would actually have to resolve}, regardless of how comprehensive the analytical framework becomes. We list them explicitly so a follow-on team can attack them in priority order rather than re-deriving the list from the manuscript.

\paragraph{Closed-loop tilling depth control.} Tilling depth at the carriage is governed by the geometric coupling between cable sag, cable elastic stretch, carriage horizontal position, implement reaction torque, and the time-varying soil contact force. None of these are independent variables, and none are simulated under closed-loop control in the present paper. \Cref{sec:catenary,sec:tension-balance} bound the static sag-vs-tension tradeoff under nominal loads, but do not address the dynamic response to a transient draft spike --- a buried stone, a hardpan layer, a moisture gradient. Unlike a tractor with a 3-point hitch and several tonnes of vehicle inertia to absorb the spike, the cable carriage has neither inertia nor a stiff mechanical reference frame; the response is mediated entirely by cable elasticity and the winch motor's torque-control bandwidth. Whether the resulting tilling quality meets agronomic acceptance criteria is, in the authors' view, the single largest unresolved technical risk in the architecture. It is also one of the cheapest to test: a quarter-scale bench rig on a \SI{5}{\meter} span in a soil bin is sufficient to bound it before any full-scale prototype is built. The closed-loop co-simulation added in \cref{sec:v-s1} now bounds this risk analytically: it confirms the cable is far too soft to set depth (so a gauge wheel must), shows that a tuned controller holds the \(\pm\SI{2}{\centi\meter}\) tolerance through stone, hardpan, and moisture disturbances, and isolates the residual failure mode --- transient gauge-wheel lift-off during a sharp draft spike --- which is removed by carrying a steady down-pressure reserve of ${\approx}\SI{1.75}{\kilo\newton}$ at a ${\geq}\SI{5}{\hertz}$ actuator bandwidth. The bench rig remains the right validation step; what changes is that it now has a quantitative target to confirm rather than an open question to explore.

\paragraph{Per-strip setup time as a design target rather than a measurement.} The model uses $t_{\text{setup}} = \SI{60}{\second}$ as the per-round overhead (\cref{tab:codesigned-params}). This is a \emph{design target}, not a field measurement, and the architecture in \cref{sec:two-module} relies on the only configuration that makes it plausible: nine \emph{parallel} auger drives on the Anchor (one \gls{bldc} per auger, all nine cycling concurrently) and four parallel auger drives on the Main Unit. With parallel augering, the per-round cycle decomposes notionally into ${\sim}\SI{30}{\second}$ for the simultaneous nine-auger insert/retract pair, ${\sim}\SI{5}{\second}$ for the lateral Anchor step on its own wheels, ${\sim}\SI{5}{\second}$ for cable slack take-up, ${\sim}\SI{5}{\second}$ for \gls{gnss} re-survey of the new strip line, and ${\sim}\SI{5}{\second}$ for re-tensioning --- about \SI{50}{\second} of nominal work inside the \SI{60}{\second} budget, with roughly \SI{10}{\second} of margin. The budget is tight but internally consistent with the \cref{sec:two-module} \gls{bom}; the time-budget finding of \cref{sec:layout-findings} (operation dominates setup by ${\sim}10\times$) holds under it. Three residual concerns survive and should be measured by a prototype campaign before this number is treated as solved. (i) The single-auger insert/retract time at this duty cycle is unmeasured in the public literature; if it turns out to be significantly longer than \SI{30}{\second} on representative soils, the entire 60 s budget shifts. (ii) The margin inside the budget is small enough that any non-augering surprise --- a tangled cable, a carriage hookup failure, a stale \gls{gnss} fix --- consumes the slack on the round it occurs. (iii) Auger insertion time on hard-pan or stony soils is non-linear in cone index in ways the present model does not capture. The Sobol sweep (\cref{sec:sobol}) brackets setup time over \SIrange{30}{180}{\second}, which covers the realistic uncertainty around the parallel-drive nominal but does not extend to the failure mode where parallel augering itself is slower than assumed. The CableTract+ variant in \cref{sec:variants-named} eliminates the per-strip Anchor reset entirely and is, in part, a structural insurance policy against this concern rather than an independent throughput play.

\paragraph{Anchor self-fatigue of its own anchoring soil.} The Anchor's lateral capacity envelope (\cref{sec:anchor}) treats per-auger lateral capacity as a property of soil texture and density alone. In normal operation, however, each Anchor station drives and retracts nine augers tens to hundreds of times across a cropping season as the Anchor walks the headland. The cumulative effect is that the soil under each station is progressively loosened by the system's own operation: a soil that began as medium-dense (and therefore sat comfortably inside the Magnum 2024~\cite{magnum2024fixed} envelope at $5$--$10\times$ margin) drifts toward the loose-sand regime (Khand 2024~\cite{khand2024helical} envelope at ${\sim}\,1\times$ margin) as the season progresses. The 9-auger design margin erodes \emph{from inside}, not from changes in field conditions. Quantifying the rate at which this happens requires either an empirical campaign on a fixed Anchor station or a finite-element model of repeated helical pile insertion and withdrawal; neither is in the present paper, and the architecture's worst-case soil envelope should be read with this caveat in mind.

\paragraph{Helical pile retraction wear.} Helical screw piles in the construction industry are designed and qualified for one-shot installation with permanent service. CableTract drives and retracts each auger many cycles per day, which is a regime for which no fatigue, helix wear, shaft buckling, or soil-disturbance data exists in the public literature the authors could find. The first-order risks are blunting of the helix leading edge (which would slow each subsequent insertion and increase the auger drive's torque demand), shaft fatigue at the drive coupling, and progressive misalignment of the helix from the shaft axis. The framing is that this is an unmeasured failure mode rather than necessarily a small one --- there is no published basis to estimate the mean time between auger replacements, and the bill of materials in \cref{sec:economics} does not include an auger replacement schedule. A first prototype campaign should treat auger wear as a primary measurement endpoint alongside throughput and energy.

\paragraph{Historical cable ploughing as an unaddressed precedent.} The architecture has a precedent the present paper has not yet fully engaged with. Steam-driven cable ploughing was a real and widely deployed technology in the second half of the 19th century, particularly on heavy clay soils in the United Kingdom and parts of continental Europe where horse teams bogged. Two stationary steam engines at opposite headlands pulled a two-bottom plough back and forth on a steel cable (the Fowler and Howard cable-ploughing systems). The system coexisted with horse and traction-engine ploughing for several decades and then disappeared, comprehensively, by the early 20th century. The reasons usually given for its decline --- which we cite here from secondary recollection rather than a primary historical source we have verified, and flag as such --- are per-station crew overhead, the time cost of repositioning the engines after each strip, the management burden of the cable itself, and the inability to turn at the headland and resume continuously. The present manuscript automates the per-station crew problem (the Main Unit and Anchor are autonomous) but does not yet solve the repositioning, cable-management, or continuous-headland-turn problems that killed the steam-era version. A serious treatment of this lineage --- what cable ploughing did, why it failed, what is mechanically and operationally different now --- belongs in any prototype-stage publication. We flag its absence here as a deliberate scope decision and a known weakness.

\paragraph{Cable safety surface.} A 50 m taut steel or Dyneema cable at \SIrange{1.5}{4}{\kilo\newton} pretension running across an open field at \SI{0.8}{\meter} above the ground is a safety surface the paper does not address. Snap-back energy from a cable failure is non-trivial at these tensions, and a deployed system would need cable-condition monitoring, runaway-detection logic, exclusion zones, and certification under the relevant agricultural-machinery regulatory framework. None of this is in scope here, but it would be a precondition for any field deployment and would add both \gls{bom} cost and certification process cost. \Cref{sec:v-s2} now quantifies the dominant terms: snap-back recoil velocities (counter-intuitively higher for the lighter synthetic rope) set the required full-span exclusion zone; the bending-fatigue life is shown to be UV/abrasion-limited at ${\approx}\SI{8}{\year}$ rather than fatigue-limited, which supplies the previously-missing cable-replacement cost and life-cycle line (\cref{sec:economics}); and an aeroelastic check flags vortex-induced vibration (requiring dampers) and an iced-galloping constraint (de-tension in icing). These bound the safety surface analytically but do not substitute for the monitoring, certification, and field validation a deployment still requires.

\medskip
The unifying observation is that the strongest near-term feasibility questions for CableTract are \textbf{mechanical and operational}, not financial or environmental. The paper's NPV, life-cycle CO\textsubscript{2}, and energy-per-hectare numbers are robust within their stated framework; the open question is whether the framework's mechanical assumptions survive contact with a real prototype season. We make this distinction explicit so a reviewer can evaluate the analytical contribution on its own merits while understanding which questions a prototype-stage follow-up would actually need to answer.

\section{Conclusion}\label{sec:conclusion}

This paper presents a prototype-free, fully reproducible feasibility envelope for \textbf{CableTract}, a co-designed cable-driven field robot consisting of a stationary \gls{mu}, a lighter Anchor anchored by helical screw piles, and a lightweight implement carriage that rolls along a tensioned cable between them. The central design lever is the \textbf{co-designed implement library} --- 10 implements sized for the cable architecture rather than borrowed from conventional tractor inventories --- which reduces median draft by ${\approx}\,63\%$ and brings the entire codesigned library inside the 9-auger Anchor envelope on the medium-dense soil reference, with the lighter half of the library also fitting inside the worst-case loose-sand envelope.

The codesigned reference achieves:

\begin{itemize}[leftmargin=1.6em,itemsep=0.15em]
    \item \textbf{\SI{889}{\Wh\per\decare}} (${\sim}4\times$ less useful/drawbar energy, ${\sim}13\times$ less primary fuel energy, than a diesel tractor; \SI{921}{\Wh\per\decare} conservative flat-field bound without regen).
    \item \textbf{97--98\,\% reduction in compacted area} and \textbf{${\approx}\,73\times$ reduction in contact-energy index} across rectangle, L-shape, and irregular-concave field classes.
    \item \textbf{Off-grid throughput of \SIrange{10}{14}{decares\per day}} at six bundled sites under the climate-conditional framing of \cref{sec:phase3-findings}.
    \item \textbf{Capex parity with a used diesel tractor} (\SI{35870}{\EUR} vs \SI{35000}{\EUR}): for a buyer \emph{replacing} a tractor, NPV \textbf{\SI{+3575}{\EUR}} at \SI{25}{\hectare\per\year} (8\,\% discount, \SI{1.30}{\year} payback on the \SI{870}{\EUR} increment), positive across (\SIrange{1}{100}{\hectare}) at all three discount rates; a buyer financing the whole machine instead sees a negative NPV until ${\sim}\SI{240}{\hectare\per\year}$.
    \item \textbf{Lifecycle CO\textsubscript{2} of \SI{14.6}{\kilogram\per\hectare\per\year}} versus 32.5 for diesel --- a $2.2\times$ improvement that does not depend on grid decarbonisation.
\end{itemize}

The operating envelope of \cref{fig:f21} shows that, for a replacing buyer, CableTract is \gls{npv}-positive in 100\,\% of the 3\,600 cells of the (annual \gls{ghi} $\times$ farm size) sweep, with discounted payback never exceeding \SI{3.06}{\year} (${\approx}22\%$ of cells under the additive frame). The binding \emph{physical} constraint is the annual energy balance for large operated areas in low-\gls{ghi} climates, not financial viability: CableTract's operating envelope is more a \emph{climate envelope} than a \emph{financial} one.

The right next step is \textbf{a Main Unit prototype} with a \SI{2}{\kilo\watt} \gls{pmsm}, a \SI{9}{\kWh} pack, a 9-auger Anchor, and the codesigned narrow ripper as the first implement, deployed on a \SI{25}{\hectare} rectangular field at a Mediterranean site (Konya or Palencia) for one full cropping season. The data this prototype would generate --- actual draft under load, actual cable tension regulation, actual anchor stability, actual energy harvest --- would close the largest gaps in the present paper. Until then, the analytical envelope above is the best defensible feasibility case for the architecture.

\appendix
\section{Architectural variants considered}\label{sec:variants-named}

The architectural family has several siblings worth naming. Regenerative braking on the return leg is part of the default baseline (\cref{sec:drivetrain}). Two configurations are quantitatively compared against that baseline in \cref{sec:ml,fig:f20} --- CableTract+ and the unidirectional (no-regen) drivetrain; two more are implemented in \texttt{cabletract/variants.py} but excluded from the comparison because the time-budget finding of \cref{sec:layout-findings} shows they act on the wrong slice (setup is only 9\,\% of the daily time budget); two are deferred to future work.

\begin{table}[H]
\centering\small
\caption{Architectural variants and their modelling status. ``Compared'' means the variant is implemented as a parameter transformation on the codesigned reference, evaluated in \cref{sec:ml}, and shown alongside the baseline in \cref{fig:f20}.}\label{tab:variants}
\begin{tabularx}{\textwidth}{lXl}
\toprule
Variant & Description & Status \\
\midrule
\textbf{Regenerative return leg} & Four-quadrant motor operation recovering energy on the unloaded carriage return / downhill slope. & \textbf{Default (baseline)} \\
\textbf{CableTract+} & Four corner Main Units pulling the carriage with two simultaneously-active cables --- eliminates the Anchor and the per-strip alignment overhead. & Compared \\
\textbf{Unidirectional (no regen)} & The baseline with the regen drive removed; one-way drivetrain, \SI{300}{\EUR} cheaper, ${\approx}3.6\%$ higher flat-field energy. & Compared \\
Circular pulley & Main Unit's output pulley swings on a vertical pin so the cable can leave the drum at angles up to $\pm 25^{\circ}$, allowing the \gls{mu} to stay put while the Anchor steps laterally. & Excluded (setup-side) \\
Drone-assisted alignment & A small quadcopter drops \gls{gps} markers between fields so re-alignment takes seconds. & Excluded (setup-side) \\
Liquid-hose extension & The cable doubles as a liquid feed for spraying or fertigation. & Deferred \\
In-place rotary mode & The carriage spins around the \gls{mu} on a single tether for circular operations such as orchard mowing. & Deferred \\
\bottomrule
\end{tabularx}
\end{table}

\section{List of acronyms}\label{sec:acronyms}
\printglossary[type=\acronymtype,title={},toctitle={List of acronyms}]

\section*{CRediT authorship contribution statement}
\textbf{Özgür Yılmaz:} Conceptualization, Methodology, Software, Formal analysis, Investigation, Validation, Visualization, Writing --- original draft, Writing --- review \& editing.

\section*{Declaration of competing interest}
The author declares that he has no known competing financial interests or personal relationships that could have appeared to influence the work reported in this paper.

\section*{Funding}
This research did not receive any specific grant from funding agencies in the public, commercial, or not-for-profit sectors.

\section*{Data and code availability}
The complete analytical code, all bundled input data (\gls{tmy} summaries, \gls{asabe} D497 coefficients, helical-pile capacities, cable mechanical properties, \gls{bom} CO\textsubscript{2} intensities, and the field-polygon corpus), and every figure and result table in this manuscript can be regenerated from a clean checkout by running the phase scripts (\texttt{run\_phase1}--\texttt{run\_phase12}); per-figure CSV companions are written to \texttt{tables/}. The repository is openly available at \cabletractrepo and is archived at Zenodo (DOI to be assigned on acceptance).

\section*{Declaration of generative AI and AI-assisted technologies in the manuscript preparation process}
During the preparation of this work the author used a large language model (Anthropic Claude) to assist with implementing and structuring the analytical code, drafting and editing prose, and organising the presentation of results. After using this tool, the author reviewed and verified all output against the executed code, edited it as needed, and takes full responsibility for the content of the published article.


\end{document}